\documentclass[pdflatex,sn-aps,iicol]{sn-jnl}

\usepackage{graphicx}%
\usepackage{multirow}%
\usepackage{amsmath,amssymb,amsfonts}%
\usepackage{amsthm}%
\usepackage{mathrsfs}%
\usepackage[title]{appendix}%
\usepackage{xcolor}%
\usepackage{textcomp}%
\usepackage{manyfoot}%
\usepackage{booktabs}%
\usepackage[utf8]{inputenc}
\usepackage{algorithm}%
\usepackage{algorithmicx}%
\usepackage{algpseudocode}%
\usepackage{listings}%
\usepackage{caption}%
\usepackage{graphicx}%
\usepackage{lipsum}%
\usepackage{array}%
\usepackage{makecell}%
\usepackage{caption} 
\usepackage{adjustbox}
\usepackage{csquotes}
\usepackage{hyperref}
\usepackage[normalem]{ulem}
\usepackage[switch,pagewise]{lineno}
\hypersetup{hidelinks,
	colorlinks=true,
	allcolors=blue,
	pdfstartview=Fit,
	breaklinks=true}
\usepackage[table]{xcolor}
\usepackage{booktabs}       
\usepackage{longtable}      
\usepackage{array}          
\newcolumntype{L}[1]{>{\raggedright\let\newline\\\arraybackslash\hspace{0pt}}m{#1}}
\newcolumntype{C}[1]{>{\centering\let\newline\\\arraybackslash\hspace{0pt}}m{#1}}
\newcolumntype{R}[1]{>{\raggedleft\let\newline\\\arraybackslash\hspace{0pt}}m{#1}}

\theoremstyle{thmstyleone}%
\theoremstyle{thmstyletwo}%

\theoremstyle{thmstylethree}%
\begin{document}

\title[Article Title]{Instruction-based Image Editing: A Survey on Data, Models, Evaluation, and Applications}


\author*[]{\fnm{Xianghao} \sur{Zang*}}\email{zangxh@pku.edu.cn}
\author[]{\fnm{Zijian} \sur{Jiang}}\email{jiangzj@tongji.edu.cn}
\author[]{\fnm{Jiarong} \sur{Cheng}}\email{lf\_cyan27@outlook.com}
\author[]{\fnm{Qianrui} \sur{Teng}}\email{qrteng@bupt.edu.cn}
\author[]{\fnm{Ying} \sur{He}}\email{heying719@seu.edu.cn}
\author[]{\fnm{Yuxuan} \sur{Mu}}\email{muyuxuan@bupt.edu.cn}
\author[]{\fnm{Chao} \sur{Ban}}\email{ban\_0330@163.com}
\author[]{\fnm{Huayu} \sur{Zhang}}\email{marvelzhy@gmail.com}
\author[]{\fnm{Lanxiang} \sur{Zhou}}\email{lansonzhou@gmail.com}
\author[]{\fnm{Zerun} \sur{Feng}}\email{zerunfeng@gmail.com}
\author[]{\fnm{Chi} \sur{Zhang}}\email{zhangc120@chinatelecom.cn}


\affil[]{\orgdiv{Institute of Artificial Intelligence (TeleAI)}, \orgname{China Telecom}, \orgaddress{\city{Beijing}, \country{China}}}


\abstract{	
	Instruction-based Image Editing (IIE) aims to transform a given image into a new one based on textual instructions. Advances in Large Language Models (LLMs) and Vision-Language Models (VLMs) have accelerated progress toward practical ``one-sentence image editing" systems. This survey presents a systematic taxonomy and comprehensive review of IIE research, structured around five core dimensions: (1) task definition and hierarchical categorization of editing operations, (2) methodologies for training data construction, (3) architectural evolution from GAN-based to diffusion and autoregressive paradigms, (4) standardized evaluation metrics and benchmark development, and (5) introduction of commercial solutions. Our analysis shows critical technological milestones across model generations.  We further propose a Comprehensive, in-Depth, and Diagnostic benchmark for IIE task (CDD-IIE Bench), which can rigorously assess the multiple aspects of model performance. Through empirical comparisons of open-source solutions, we highlight their respective capabilities and limitations. Finally, we discuss future research directions to advance the field. 
}


\maketitle

\section{Introduction}\label{sec1}
With the ubiquitous presence of visual capture devices in modern society, vast quantities of digital images are being generated continuously. However, these raw images often fail to meet human aesthetic or functional requirements. While professional editing tools (e.g., Adobe Photoshop) and consumer applications (e.g., Meitu) have addressed this gap, their reliance on manual operations presents significant usability barriers. The emergence of Large Language Models (LLMs) and Vision-Language Models (VLMs) has catalyzed a paradigm shift toward instruction-based image editing (IIE), enabling ``one-sentence photo manipulation" through natural language instructions. This technological breakthrough demonstrates significant practical value across numerous applications.

Early image editing models predominantly relied on Generative Adversarial Networks (GANs)~\cite{goodfellow2014generative} for image synthesis. These methods demonstrated notable advantages over manual editing techniques, such as style transfer~\cite{baykal2023clipguided}, face editing~\cite{tero2019stylegan}, image restoration~\cite{pathak2016context}, super-resolution~\cite{ledig2017photo}, and semantic manipulation~\cite{patashnik2021styleclip}. However, GAN-based approaches were often constrained to narrow editing domains and primarily facilitated coarse, large-scale modifications to input images. The emergence of diffusion models marked a significant advancement in image editing capabilities. Compared to GANs, diffusion models exhibit superior performance, offering enhanced fine-grained controllability. A key strength of these models lies in their capacity to effectively incorporate textual guidance (e.g., through CLIP embeddings), enabling sophisticated applications such as text-conditioned image manipulation (e.g., Prompt-to-Prompt~\cite{hertz2022prompttoprompt}, DiffEdit~\cite{couairon2022diffedit}), image inpainting (e.g., RePaint~\cite{lugmayr2022repaint}, Blended Diffusion~\cite{avrahami2022blended}). Later, autoregressive models gained prominence, leveraging their strong language comprehension abilities to achieve finer-grained image editing~\cite{yu2023cm3leon}. To further enhance editing performance, unified models integrating diffusion and autoregressive architectures were developed~\cite{wang2025ovisu, wu2025omnigen, lu2024unifiedio}. These hybrid models combined the strengths of both paradigms—enhanced semantic understanding and high-quality image generation—thereby advancing the field of image editing.

Although significant progress has been made in IIE models, several critical challenges remain unresolved: (1) The diversity of image editing tasks lacks a comprehensive and unified definition across different editing tasks. (2) There is a notable absence of systematic reviews summarizing recent advancements, particularly in emerging approaches like autoregressive models and unified architectures. (3) This field suffers from insufficient evaluation metrics to comprehensively assess various editing models. These gaps hinder the development of more robust and generalizable IIE systems.

To address these challenges, this paper presents a comprehensive review of the field. Specifically, we organize our work as follows: Section 2 formally defines the task of image editing and categorizes it into a hierarchical taxonomy of sub-tasks. Additionally, it introduces the fundamental principles of Generative Adversarial Networks (GANs), Diffusion Models (DMs), and Autoregressive (AR) Models, along with key terminology used for IIE task. Section 3 details the data preparation pipeline for different image editing sub-tasks, providing the foundation for training data-driven editing models. Section 4 presents a chronological review of state-of-the-art models in image editing, tracing their evolution and key innovations. Section 5 evaluates existing performance metrics for image editing tasks and introduces CDD-IIE Bench, a comprehensive, in-depth, and diagnostic benchmark designed to address current evaluation limitations. This section also includes comprehensive testing of leading open-source models. Section 6 introduces mainstream commercial instruction-based editing products. Section 7 explores future research directions for both open-source development and commercial product innovation. Finally, Section 8 concludes the paper with a summary of key findings and insights.

The key contributions of this work are threefold: (1) We establish a systematic taxonomy of diverse image editing tasks, providing clear definitions and boundaries for the IIE scope. (2) We conduct a comprehensive survey and rigorous analysis of instruction-based image editing (IIE), encompassing four critical dimensions: training data construction methodologies, evolutionary trends in model architectures, evaluation frameworks, and comparative assessment of both open-source and proprietary solutions. (3) We introduce a novel hierarchical evaluation metric system consisting of 5 principal categories with 21 fine-grained sub-criteria, accompanied by extensive benchmarking of cutting-edge models.

\section{Preliminaries} \label{secB2}

\begin{table*}
	\centering
	\caption{Task description of basic atomic editing suite.}\label{tab:taskDef_basic}
	\begin{tabular}{@{}p{3.5cm}p{11.5cm}@{}}
		\toprule
		\textbf{Tasks} & \textbf{Definition} \\
		\midrule
		
		\multicolumn{2}{@{}l}{\textbf{Object-Level Operations}} \\
		Addition & Inserting a new object into the image. \\
		Removal & Removing a specific object from the image and filling the space with a consistent background. \\
		Attribute Modification & Altering specific attributes of objects in the image, such as color, material, texture, or overall appearance. \\
		Replacement & Replacing a specific object in the image with a new object while maintaining the consistency of the scene. \\
		Text Modification & Detecting and editing textual elements that are present within an image. \\
		Portrait Enhancement & Applying cosmetic or stylistic enhancements to a person's portrait, such as skin refinement or makeup adjustments. \\
		Motion Change & Modifying the posture, action, or dynamic state of a person or object to depict a different movement or pose. \\
		Extraction & Isolating a specified subject onto a clean background while preserving their identity. \\
		\midrule
		
		\multicolumn{2}{@{}l}{\textbf{Image-Level Operations}} \\
		Tone Transformation & Modifying the overall tone of the image, including aspects like color, lighting, weather conditions, and seasons. \\
		Style Transfer & Modifying the overall style of the entire image according to a given style description. \\
		Background Change & Modifying the background of an image while precisely preserving the subjects in the foreground. \\
		\midrule
		
		\multicolumn{2}{@{}l}{\textbf{Utility Tasks}} \\
		Image Repair & Correcting degradations in an image to improve its overall quality. This includes tasks like deblurring, denoising, and watermark removal. \\
		Visual Effect Removal & Eliminating specific environmental effects from an image, such as haze, rain, snow, and shadows. \\
		\bottomrule
	\end{tabular}
\end{table*}

\subsection{Task Definition}
This paper formally defines Instruction-based Image Editing (IIE) as a transformation process that takes a source image and textual instructions as inputs, and produces an edited image as output. We categorize IIE tasks into two primary evaluation suites: (1) a basic atomic editing suite designed to assess fundamental, singular editing operations, and (2) an advanced compositional editing suite for evaluating complex, multi-faceted editing capabilities. Together, these suites comprehensively cover 5 major evaluation dimensions encompassing 21 specific editing tasks. 

\subsubsection{Basic Atomic Editing}
The basic atomic editing suite focuses on testing core, isolated editing operations that serve as building blocks for more complex transformations. For specific task definitions, see Table~\ref{tab:taskDef_basic}.

\subsubsection{Advanced Compositional Editing}

The advanced compositional editing suite targets a model's ability to comprehend complex instructions, perform multi-step reasoning, and understand intricate spatial environments and object relationships. These tasks are designed to effectively probe the upper limits and capability boundaries of existing models. For specific task definitions, see Table~\ref{tab:taskDef_advanced}.

\begin{table*}
	\centering
	\caption{Task description of advanced compositional editing suite.}\label{tab:taskDef_advanced}
	\begin{tabular}{@{}p{3.5cm}p{11.5cm}@{}}
		\toprule
		\textbf{Tasks} & \textbf{Definition} \\
		\midrule
		
		\multicolumn{2}{@{}l}{\textbf{Complex Instruction \& Reasoning}} \\
		Parallel Instructions & Executing multiple, independent editing instructions on a single image. \\
		Sequential Instructions & Executing a series of edits where each subsequent instruction depends on the result of the previous one. \\
		Implicit Reasoning & Performing an edit that requires reasoning to infer the target and the desired change, as they are not explicitly stated in the prompt. \\
		\midrule
		
		\multicolumn{2}{@{}l}{\textbf{Spatial Understanding \& Reasoning}} \\
		Viewpoint Transformation & Altering the image to simulate camera movement, including sub-tasks like translation, zooming in/out, and rotating. \\
		Referring-based Editing & Accurately identifying the object or area of interest from a complex scene based on referring expressions and performing precise edits. \\
		Position Adjustment & Modifying the spatial location of a single object or adjusting the relative positioning between multiple objects within the image. \\
		Size Adjustment & Adjusting the size of a specified object in relation to other elements within the scene. \\
		Count Change & Modifying the number of a particular object within the image to match a specified quantity. \\
		\bottomrule
	\end{tabular}
\end{table*}

\subsection{Model Preliminaries}
This section presents the fundamental model architectures and their associated loss functions for Instruction-based Image Editing (IIE) tasks, including Generative Adversarial Networks (GANs), Diffusion Models, and Autoregressive (AR) approaches.

\subsubsection{Generative Adversarial Networks}
Generative Adversarial Networks (GANs)~\cite{goodfellow2014generative} consist of two competing neural networks: a generator ($G$) and a discriminator ($D$). The generator synthesizes fake samples (e.g., images) from random noise, while the discriminator distinguishes between real and generated samples. This adversarial process is formalized as a Minimax game, where the generator aims to fool the discriminator, and the discriminator improves its detection capability. The training objective is given by:
\begin{equation}
	\begin{split}
		\underset{G}{\min}~\underset{D}{\max}~ \mathcal{L}(D,G)= \mathbb{E}_{x \sim p_\mathrm{r}(x)}\log D(x) + \\
		\mathbb{E}_{z \sim p_z(z)}\log [1-D(G(x))]
	\end{split}
\end{equation}
where $p_\mathrm{r}(x)$ is the real data distribution, $p_z(z)$ is the noise prior (e.g., Gaussian), and $D(x)$ outputs the probability that $x$ is real.

GAN training involves optimizing $G$ and $D$ alternately. The discriminator maximizes its classification accuracy, while the generator minimizes $\log (1 - D(G(z)))$, equivalent to maximizing $\log D(G(z))$.

\subsubsection{Diffusion Models}
Stable diffusion~\cite{rombach2021highresolution} is a Latent Diffusion Model (LDM) that synthesizes high-quality images by iteratively denoising a latent representation. Stable diffusion employs a Variational AutoEncoder (VAE) to compress images into a lower-dimensional latent space, significantly improving computational efficiency. The model consists of three key components: (1) VAE Encoder/Decoder: Maps images $x$ to latent codes $z=\mathcal{E}(x)$ and reconstructs them as $\tilde{x} = \mathcal{D}(z)$. (2) Denoiser: Predicts noise $\epsilon_\theta$ in the latent space conditioned on text prompts or other inputs. (3) Diffusion Process: Gradually adds and removes Gaussian noise over $T$ timesteps. The forward diffusion process corrupts $z_0$ (encoded input) into $z_t$ via:
\begin{equation}
	z_t = \sqrt{\alpha_t}z_0 + \sqrt{1-\alpha_t}\epsilon, \epsilon \sim \mathcal{N}(0, I),
\end{equation}
where $\alpha_t$ is a noise schedule.

Stable diffusion is trained to reverse the diffusion process by minimizing the denoising score matching loss:
\begin{equation}
	\mathcal{L}_{\mathrm{DM}} = \mathbb{E}_{z_0, \epsilon, t}[\| \epsilon - \epsilon_\theta(z_t, t, c) \|_2^2],
\end{equation}
where $c$ is a conditioning vector (e.g., text embeddings from CLIP), and $\epsilon_\theta$ is the noise prediction. 

Stable diffusion latent-space diffusion framework enables efficient and controllable image synthesis. By integrating domain-specific losses (e.g., CLIP, LPIPS), it achieves state-of-the-art performance in text-to-image generation and editing. 

\subsubsection{Autoregressive Models}
Autoregressive (AR) models generate data sequentially by modeling the conditional probability distribution of each element given previous elements. For images, pixels are typically ordered in a raster scan (row-wise), and the likelihood of an image $x=(x_1, \cdots, x_N)$ is factorized as:
\begin{equation}
	p(x) = \prod_{i=1}^{N}p(x_i \| x_1, \cdots, x_{i-1}),
\end{equation}
where $x_i$ represents a pixel or a patch. 
AR models are trained via Maximum Likelihood Estimation (MLE), minimizing the Negative Log-Likelihood (NLL) of the training data:
\begin{equation}
	\mathcal{L}_\mathrm{MLE} = - \mathbb{E}_{x \sim p_\mathrm{data}} [\sum_{i=1}^{N} \log p_\theta(x_i \| x_{1:i-1})],
\end{equation}
where $p_\theta$ is the model learned distribution. For discrete pixel values (e.g., 8-bit images), $p_\theta$ is often a categorical distribution, while continuous values (e.g., normalized pixels) use a Gaussian Mixture Model (GMM) or logistic distribution.

Autoregressive models offer a principled framework for image generation and editing, with flexibility in incorporating task-specific losses. 

\subsubsection{Loss Functions}
To address diverse editing tasks, the aforementioned fundamental model architectures can be effectively combined with specialized loss functions. The key loss functions include:
\begin{itemize}
	\item Reconstruction Loss: Preserves content fidelity between edited $\hat{x}$ and target $x$:
	\begin{equation}
		\mathcal{L}_\mathrm{rec} = \| x - \hat{x} \|_1 + \lambda_\mathrm{LPIPS} \cdot \mathcal{L}_\mathrm{LPIPS}(x, \hat{x}),
	\end{equation}
	where $\lambda_\mathrm{LPIPS}$~\cite{zhang2018unreasonable} measures perceptual similarity using a pre-trained VGG network.
	\item Perceptual Loss~\cite{johnson2016perceptual}: Uses a pre-trained network (e.g., VGG) to match high-level features:
	\begin{equation}
		\mathcal{L}_\mathrm{prec} = \| \phi(x) - \phi(\hat{x}) \|_2^2
	\end{equation}
	where $\phi(x)$ denotes deep features
	\item Adversarial Loss: A discriminator $D$ can refine outputs by encouraging realism:
	\begin{equation}
		\mathcal{L}_\mathrm{adv} = \mathrm{E}[\log D(x) + log(1-D(\hat{x}))].
	\end{equation}
	\item Latent Alignment Loss: Ensures edited latent codes $\hat{z}$  align with the target distribution:
	\begin{equation}
		\mathcal{L}_\mathrm{latent} = \| \mathcal{E}(x)-\hat{z} \|_2^2,
	\end{equation}
	\item Text-Guided Loss: Aligns edits with a text prompt $c$ via CLIP:
	\begin{equation}
		\mathcal{L}_\mathrm{CLIP} = - \langle \phi_\mathrm{CLIP}(\hat{x}), \phi_\mathrm{CLIP}(c) \rangle,
	\end{equation}
	where $\phi_\mathrm{CLIP}$ denotes CLIP embedding space.
	\item Conditional Likelihood Loss: For tasks like inpainting, where observed pixels $\mathrm{x}_{\mathrm{obs}}$ constrain the generation of missing pixels $\mathrm{x}_{\mathrm{mask}}$
	\begin{equation}
		\mathcal{L}_\mathrm{edit} = - \mathbb{E} [\sum_{i\in \mathrm{mask}} \log p_\theta(x_i \| \mathrm{x}_\mathrm{obs} , x_{1:i-1})],
	\end{equation}
\end{itemize}

\section{Data Construction} \label{secB3}

The impressive progress in instruction-based image editing owes much to the high-quality, well-structured training datasets. This section provides a systematic overview of mainstream data construction methodologies for image editing across various editing task categories.

\subsection{Dataset Construction}
\begin{figure}[htbp]
	\centering
	\includegraphics[width=\columnwidth]{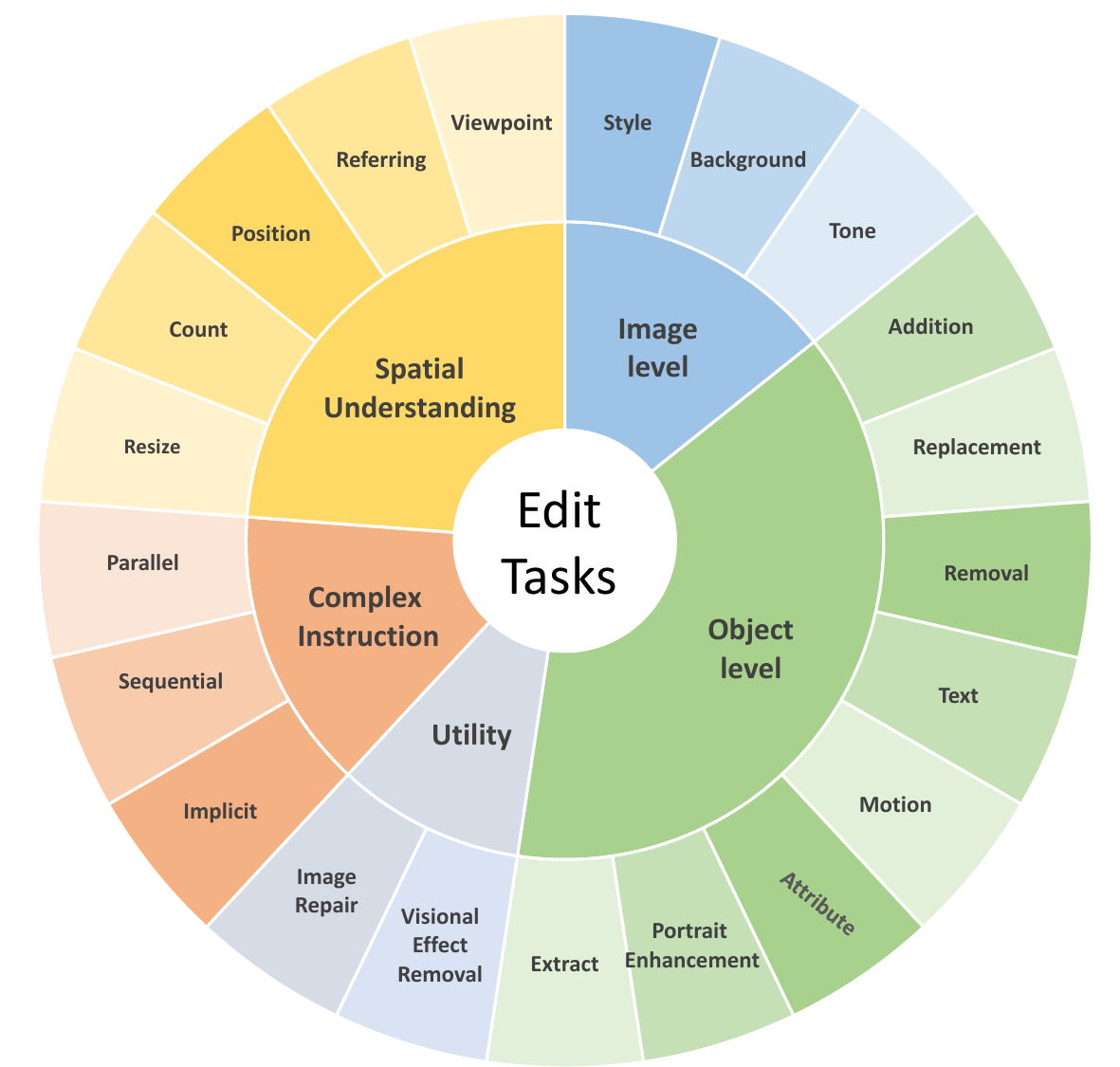}
	\caption{Different edit tasks.}
	\label{fig:edit_tasks}
\end{figure}

\begin{figure*}[htbp]
	\centering
	\includegraphics[width=1.0\textwidth]{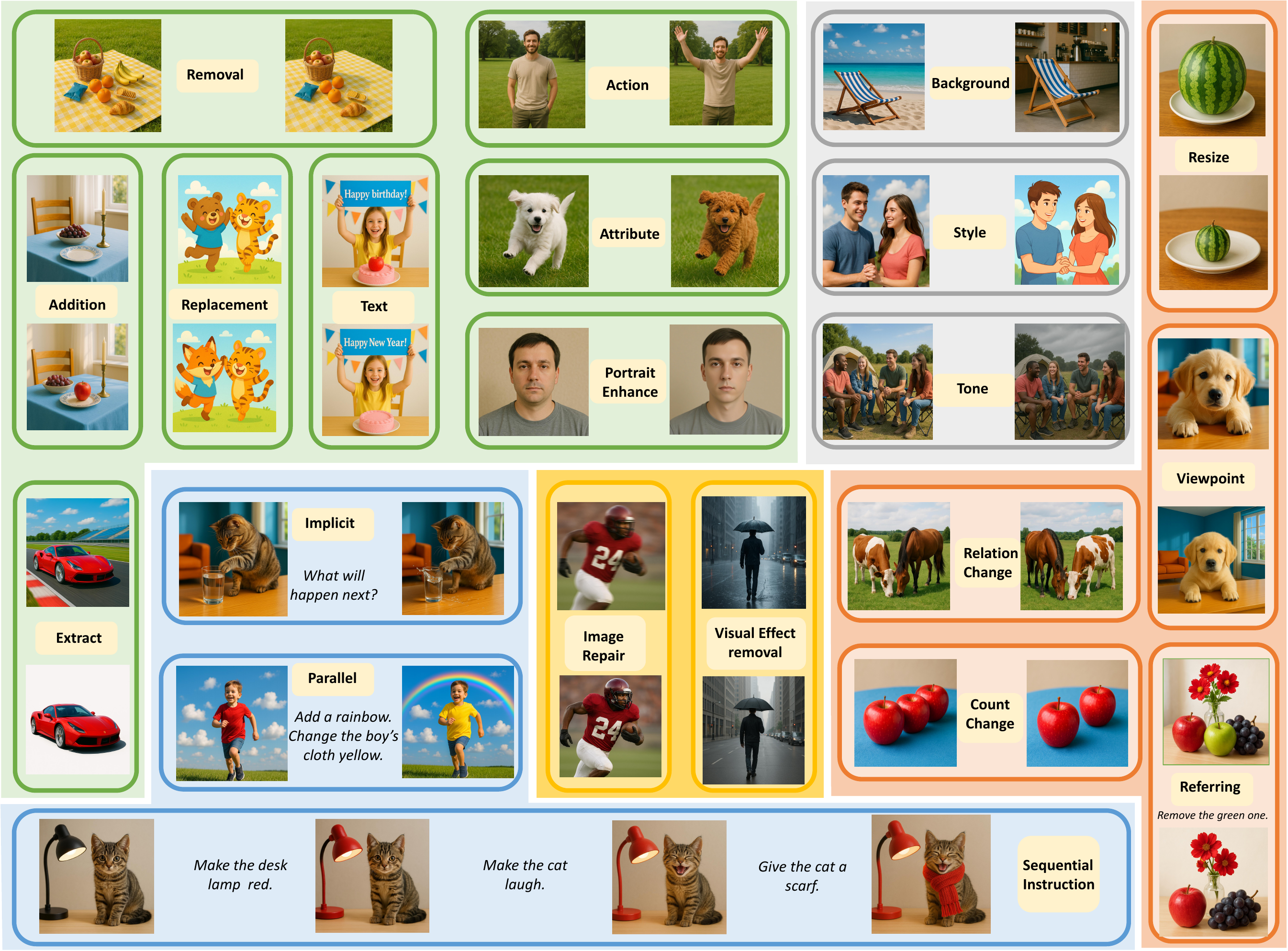}
	\caption{Presentation of different edit tasks. In this image, the colors green, gray, orange, blue, and red as background colors represent Object-Level Operations, Image-Level Operations, Utility Tasks, Complex Instruction and Reasoning, and Spatial Understanding and Reasoning tasks respectively.}
	\label{fig:different_edit_task}
\end{figure*}

To systematically construct instruction-based training datasets, we adopt a refined taxonomy that categorizes editing tasks into five representative types based on their operation granularity and reasoning complexity. As illustrated in the Fig.~\ref{fig:edit_tasks}, these categories include: 1) Object-Level Operations: focuses on localized modifications. 2) Image-Level Operations: encompasses global transformations affecting the entire image. 3) Utility Tasks: involves editing operations aimed at improving image quality. 4) Complex Instruction and Reasoning: includes tasks that require high-level linguistic comprehension and multi-step reasoning. 5) Spatial Understanding and Reasoning: involving positional or relational changes between multiple objects. Fig.~\ref{fig:different_edit_task} shows comparison examples of pre- and post-editing for various tasks.

\subsection{Object-Level Operations}

\subsubsection{Removal}
The removal task aims to eliminate specific objects or regions from an image according to textual instructions, while keeping the visual plausibility and structural consistency of the surrounding context.

Early approaches first generated the edited images, then the corresponding original version was inferred by removing object. For example, EmuEdit~\cite{sheynin2024emuedit} synthesizes the edited image from output captions, then applies attention-controlled masking to erase the object. This strategy ensures semantic alignment between the image and instruction but may suffer from inpainting artifacts due to the generation process.

Recent methods adopt a more explicit modular pipeline, typically involving: 1) object identification guided by text prompts, 2) precise segmentation, and 3) content-aware inpainting for background restoration. For instance, AnyEdit~\cite{jiang2025anyedit} utilizes GroundingDINO~\cite{liu2024grounding} and SAM~\cite{kirillov2023segment} to locate and mask target objects, followed by Stable Diffusion-based inpainting to complete the background. Step1X~\cite{liu2025stepxedit} further refines this pipeline by combining large vision-language models (e.g., Qwen2.5-VL~\cite{bai2025qwen2}) with structure-aware inpainting tools like Flux Fill model, enhancing semantic grounding and visual fidelity. This evolution from implicit generation to modular construction reflects a shift toward controllable, interpretable pipelines for removal data creation.

To filter the unqualified data, several metrics are employed to ensure that the edited images align with the given instructions while preserving the visual integrity of unedited regions. First, CLIP-based text-image similarity is used to verify semantic alignment between the instruction and the generated image. Second, CLIP-based image-image similarity and L1 distance between the original and edited images are computed to ensure that unedited areas remain consistent and that the global structure is preserved. Additionally, CLIP text-image similarity is applied to confirm that the visual changes correspond to the intended modification direction implied by the instruction. These metrics together enable robust filtering of low-quality or misaligned samples, ensuring both semantic fidelity and visual plausibility in the resulting dataset.

\subsubsection{Addition}
The addition task involves inserting a new object into the image according to the instruction, such as ``add a cup on the table".

Given the difficulty of obtaining accurate masks for newly added objects, most existing approaches treat the add task as the inverse of removal task. EmuEdit~\cite{sheynin2024emuedit} first generates the edited image based on the output caption, then extracts the added object mask and regenerates the original image by masking and attention-controlled inpainting. Similarly, AnyEdit~\cite{jiang2025anyedit} constructs edited images from target captions, then produces the source image by applying a ``Removal" instruction. Step1X~\cite{liu2025stepxedit} and ImgEdit~\cite{ye2025imgedit} also adopt this reverse-removal paradigm to simplify data construction.

This strategy offers a practical advantage by circumventing the need to annotate new object masks manually. However, it also inherits and amplifies the dependency on the accuracy and quality of the removal pipeline. The filtering strategy for add task generally follows that of remove tasks, focusing on instruction-image alignment, visual consistency, and semantic plausibility.

\subsubsection{Replacement}
The replacement task aims to substitute an object or region in the image with a new one described in the instruction, while maintaining scene coherence and spatial alignment. Compared to the removal task, replacement requires not only removal of the original object but also insertion of a semantically relevant and contextually appropriate new object at the same location.

A typical pipeline for the replacement task mirrors that of the removal task: it involves detecting and segmenting the target object, removing it via inpainting, and then inserting the new object. The key difference lies in how the substitute content is determined and integrated. AnyEdit~\cite{jiang2025anyedit} address replacement by modifying the instruction prompt after object removal. Specifically, once the original object is segmented and masked, the prompt is updated to include the new target category. This guides the generative model to synthesize the replacement within the masked region, effectively transforming removal into replacement through prompt conditioning.

More recent approaches like Step1X~\cite{liu2025stepxedit} enhance this pipeline with deeper semantic understanding. It employs large vision-language models (e.g., Qwen2.5-VL~\cite{bai2025qwen2} and Recognize-Anything~\cite{zhang2024recognize}) to explicitly identify both the object to be replaced and its substitute. SAM-2~\cite{ravi2024sam} and Flux Fill model are used for segmentation and structure-aware inpainting, ensuring that the inserted object aligns with scene geometry and maintains visual realism.

To further improve perceptual quality, ImgEdit~\cite{ye2025imgedit} introduces edge softening techniques around the mask boundaries, minimizing visual artifacts and ensuring smoother transitions between the edited region and its surroundings.

Since the replacement task builds upon removal, its evaluation inherits most criteria such as instruction alignment and structural consistency. However, it further requires verifying whether the inserted object matches the intended substitute. This can be achieved through CLIP-based similarity to confirm that the visual change corresponds accurately to the specified replacement instruction.

\subsubsection{Motion Change}
The motion change task focuses on modifying the pose, motion direction, or dynamic attributes of objects in an image, as directed by textual instructions such as ``make the person sit down" or ``make the dog jump". Unlike appearance edits, this task involves structural or temporal transformations, requiring the model to reason about fine-grained spatial changes or motion semantics.

Some approaches explored low-level appearance cues to approximate motion effects. For instance, ImgEdit~\cite{ye2025imgedit} combines canny edge maps with ControlNet~\cite{lv2023controlnet} and Canny LoRA~\cite{hu2022lora} to guide localized edits. While primarily used for attribute shifts, this setup supports coarse motion-related modifications such as pose changes or exaggerated body gestures. Edge smoothing further improves visual continuity around edited regions.

To achieve more complex and semantically accurate motion transformations, AnyEdit~\cite{jiang2025anyedit} addresses the architectural limitations of traditional models in handling action change. It proposes a joint mechanism composed of mutual self-attention and masked cross-attention modules, allowing the model to better align motion-specific semantics with visual edits. Step1X~\cite{liu2025stepxedit} introduces a data-centric pipeline based on real motion supervision. It leverages Koala-36M~\cite{wang2025koala}, a video-based dataset, and selects frame pairs with meaningful foreground movement using motion estimation techniques like BiRefNet~\cite{zheng2024bilateral} and RAFT~\cite{teed2020raft}. Clean supervision is achieved by foreground-background separation. The differences between frames are described using GPT-4o~\cite{openai2024gpt4o}, generating fine-grained motion instructions for training.

Evaluation of action editing involves checking whether the modified pose matches the instruction and remains visually realistic. This is done using CLIP text-image similarity for instruction alignment, pose estimation metrics such as Percentage of Correct Keypoints (PCK) or Object Keypoint Similarity (OKS), and image similarity metrics (e.g., CLIP image-image similarity, L1 distance) to ensure background and identity preservation.

\subsubsection{Attribute Modification}
The attribute modification task typically includes color and material change task. It focuses on modifying an object appearance such as its color, texture, or surface quality without altering its structure or semantic identity. This task is typically triggered by instructions like ``make the car red" or ``change the sofa to leather", and emphasizes fine-grained visual control.

Early efforts focused on text-driven generation using prompt engineering. For example, InstructPix2Pix~\cite{brooks2023instructpixpix} leverages prompt-to-prompt transformation, where a base prompt is modified into a target prompt that reflects the desired attribute change. Paired images are generated using attention-controlled diffusion models.

Then, EmuEdit~\cite{sheynin2024emuedit} adopts a simulation-based strategy by applying low-level transformations to synthesize attribute variations. It generates the base image from text, localizes the target region via semantic grounding and attention maps, and performs iterative edits using sampled parameters (e.g., blending ratio, guidance strength). 

To enhance semantic precision, AnyEdit~\cite{jiang2025anyedit} introduces a Normalized Attention Difference (NAD) mechanism that computes CLIP-based differences between input and output captions to produce a soft attention mask. This highlights regions requiring appearance change while preserving structure. The masked region is then edited via InstructPix2Pix~\cite{brooks2023instructpixpix} and blended with the original image to ensure global coherence.

Step1X~\cite{liu2025stepxedit} takes a geometry-aware approach by incorporating depth priors. It first detects the target object and applies ZeoDepth to estimate its 3D geometry. ControlNet~\cite{lv2023controlnet} then guides the attribute change, ensuring that edits like color or material transformation follow spatial layout and object boundaries, resulting in more physically plausible outcomes.

\subsubsection{Text Modification}
The text modification task focuses on modifying scene text within an image, including operations such as text addition, removal, and replacement, while preserving visual realism and layout consistency. It requires accurate localization of text regions and the ability to generate visually coherent edits guided by user instructions.

Two primary strategies are used to construct such datasets: (1) extracting image-text pairs from existing OCR-rich datasets, and (2) leveraging OCR models to detect in-image text and then applying editing operations (e.g., add, remove, replace) to generate new samples.

AnyEdit~\cite{jiang2025anyedit} adopts the first approach by collecting samples from the large-scale AnyWord-3M dataset, which aggregates multiple OCR benchmarks. Instructions are carefully designed to constrain edit types and improve instruction diversity. In contrast, both EmuEdit~\cite{sheynin2024emuedit} and Step1X~\cite{liu2025stepxedit} rely on OCR-driven pipelines: EmuEdit extracts text masks and overlays replacement text with customizable fonts and colors;  Step1X uses PPOCR~\cite{du2009pp} for text recognition, filters out invalid regions, and generates instructions via Step-1o, with final results verified by human annotators for accuracy.

Evaluation of text editing typically uses OCR-based recognition to check if the edited text matches the instruction. CLIP-based text-image alignment ensures semantic consistency, while image similarity metrics help verify that background and layout remain unchanged.

\subsubsection{Portrait Enhancement}
Portrait enhancement focuses on improving or modifying facial features in images to enhance perceived attractiveness while preserving identity and realism. Common edits include smoothing skin, adjusting facial contours, whitening teeth, or altering makeup. This task requires fine-grained control and semantic understanding of facial attributes, making it particularly sensitive to over-editing or identity distortion.

Moreover, portrait enhancement is inherently subjective, with aesthetic standards varying significantly across individuals, cultures, and contexts. As a result, data construction for this task tends to rely more heavily on real-world datasets and manual annotations to ensure visual plausibility and cultural relevance.

For instance, Step1X~\cite{liu2025stepxedit} addresses portrait enhancement by leveraging both public and human-curated data sources. It collects beautification pairs from publicly available datasets and applies Step-1o to ensure consistency in layout and background. In addition, human editors are invited to manually perform enhancement on selected portrait images, and all edited results undergo human validation to ensure visual quality and editing accuracy.

\subsubsection{Extraction}
Object extraction also relies on segmentation. It focuses on isolating the target object and re-rendering it as a stand-alone entity against a clean (typically white) backdrop. It not only removes irrelevant visual context but also presents the object in a structured and display-oriented format, making it valuable for applications such as product cataloging, visual asset reuse, and modular image editing.

One effective strategy for constructing the pipeline is to begin in reverse: first generating the edited image that contains the target object on a clean background, and then retrieving the corresponding original image through segmentation and alignment. For example, ImgEdit~\cite{ye2025imgedit} starts by generating an isolated object image using a prompt such as ``a teddy bear on a white background". It then locates the same object within a real-world image via segmentation, extracts it, and aligns it with the generated version using Flux-Redux. This process yields both the standalone object image and a naturally edited version where the object is composited into the real scene, forming a coherent instruction-image pair.

\subsection{Image-Level Operations}

\subsubsection{Background Change}
The background change task focuses on semantically modifying the background of an image according to a given textual instruction, while preserving the integrity of foreground object and ensuring overall image realism. A common data construction pipeline includes three steps: instruction parsing, background mask extraction, and inpainting-based image generation.

Early approaches adopted reverse construction strategies, where the edited image with the new background was first synthesized, and the original version was inferred through masking or subtraction. For example, EmuEdit~\cite{sheynin2024emuedit} begins with an edited background image, extracts the background mask, refines it with minimum filtering and Gaussian smoothing to reduce contour artifacts, and blends the inpainted region with the preserved foreground. Multiple variants are generated with different guidance scales and noise levels, and the best result is selected using CLIP-based similarity ranking.

Recent methods emphasize more explicit and modular pipelines, typically comprising: (1) instruction parsing to identify background semantics, (2) foreground segmentation and mask inversion to isolate the background, and (3) inpainting-based synthesis to generate the new scene. AnyEdit~\cite{jiang2025anyedit} automatically extracts foreground masks from captions and generates refined background masks through dilation and smoothing, which are used to guide Stable Diffusion for realistic background replacement. Step1X~\cite{liu2025stepxedit} follows a more structured pipeline by using large vision-language models (e.g., Qwen2.5-VL~\cite{bai2025qwen2}) to identify background-related concepts, segmenting them via SAM-2~\cite{ravi2024sam}, and finally applying Flux-Fill for content-aware completion, ensuring semantic alignment and geometric plausibility.

\subsubsection{Style Transfer}
The style transfer task aims to modify the overall visual style of an image according to a textual instruction (e.g., ``turn this into a painting" or ``make it look like a Ghibli frame"), while preserving the image structural and semantic content. 

Early attempt like EmuEdit~\cite{sheynin2024emuedit} adopts a Plug-and-Play (PnP) guidance strategy for more nuanced control. Given an input image and style instruction, DDIM inversion is used to guide the generation of ten stylized variants with varying guidance scales (6.5-10.0), structural blending ratios (set to 0.8), and denoising strengths. 

Some methods leveraged direct stylization APIs and rule-based filtering to construct paired training data. For instance, AnyEdit~\cite{jiang2025anyedit} utilizes the Prisma API to apply various artistic styles to MSCOCO images, followed by keyword-based filtering to ensure alignment between the transformed images and their textual instructions.

Recent approaches have introduced more controllable and bidirectional pipelines to enhance generation quality and structural preservation. Step1X~\cite{liu2025stepxedit} supports both stylization and de-stylization by extracting edge maps from stylized or photorealistic inputs. These maps are fed into ControlNet-guided diffusion models to reconstruct either stylized or realistic outputs, enabling consistent structure retention across styles.

\subsubsection{Tone Transformation}
This category aimed at altering the overall appearance of an image by simulating environmental changes such as seasonal shifts, time-of-day transitions or weather variations. Methods typically apply algorithmic filters or learned transformations to adjust the image tone while preserving its structural content.

Early methods introduces diversity through low-level transformations across multiple samples. EmuEdit~\cite{sheynin2024emuedit} generates several candidates with varied transformation parameters (e.g., guidance strength, region intensity) and selects the most instruction-consistent result based on CLIP similarity.

Today, methods primarily relied on algorithmic filters or hand-crafted transformations to simulate such changes. For example, Step1X~\cite{liu2025stepxedit} implements tone modifications like dehazing, deraining, color grading, and seasonal shifts through automated tools and rule-based adjustments. The other approaches adopt learning-based pipelines that offer more semantic control and visual diversity. AnyEdit~\cite{jiang2025anyedit} categorizes tone edits into three scene types ``season, time, and weather" and generates corresponding instructions. It leverages InstructPix2Pix~\cite{brooks2023instructpixpix} to perform instruction-guided, full-image modifications, enabling context-aware and stylized tone changes.

\subsection{Utility Tasks}

\subsubsection{Image Repair}
Image repair focuses on correcting degradations in images to improve visual quality, encompassing tasks such as deblurring, denoising, and watermark removal. Unlike semantic editing, image repair emphasizes low-level fidelity while preserving structural and content integrity.

Early dataset construction approaches synthetically degrade high-quality images through operations such as Gaussian noise addition, motion blur simulation, or watermark overlay to obtain paired data such as BSD500~\cite{martin2001bsd500}, GoPro~\cite{nah2017gopro}, and Watermark-Removal datasets~\cite{qin2021watermarkremoval}. However, purely synthetic degradation often fails to capture the complexity of real-world distortions.

To address this limitation, later efforts incorporate real degradations or more realistic simulation processes. For instance, RealBlur~\cite{rim2020realblur} collects naturally blurred and corresponding sharp image pairs via burst photography to reflect real-world motion and defocus blur. For denoising, SIDD~\cite{abdelhamed2018sidd} captures real noisy images from smartphones under diverse lighting conditions to support realistic noise reduction. For watermark removal, CLWD~\cite{zhang2021clwd} combines synthetic and real-world watermarked images, improving generalization across varied watermark styles.

\subsubsection{Visual Effect Removal}
Similar to the tone transformation task, visual effect removal aims to eliminate environmental artifacts such as rain, fog, and snow from images, thereby enhancing clarity and restoring visual realism. This task is essential for improving image quality in adverse weather conditions and supporting downstream vision applications under challenging environments.

Due to its relatively tractable nature, many existing dataset construction pipelines employ either algorithmic filters or data-driven models to generate de-weathered counterparts from weather-affected images. These methods enable the creation of large-scale paired data, facilitating supervised training while ensuring the preservation of structural details and scene semantics.

\subsection{Complex Instruction Parsing and Reasoning}

\subsubsection{Parallel Instructions}
Parallel instruction editing refers to scenarios where multiple independent editing instructions are applied simultaneously to a single image (e.g., ``make the sky blue and add a hat on the person"). This requires models to execute each instruction accurately without semantic or visual interference across edits.

To date, no existing dataset or model explicitly focuses on parallel instruction execution. Most methods handle complex edits by decomposing them into modular steps or supporting multi-task editing, and operate in a single-instruction-per-edit paradigm. Even where the editing model supports multiple tasks, the edits are typically executed sequentially or greedily, rather than as a unified parallel instruction.

This gap highlights an opportunity for future research: constructing dedicated parallel-instruction datasets and models capable of compositional editing in a single forward pass.

\subsubsection{Sequential Instructions}
Sequential Instructions editing refers to the sequential modification of same object in image through a series of step-wise instructions. Unlike single-step editing, this task requires the model to track historical changes, interpret current instructions in context, and produce edits that are consistent with the accumulated editing intent. Each step must not only satisfy the current prompt but also preserve coherence with prior modifications in terms of semantics, appearance, and spatial structure. 

ImgEdit~\cite{ye2025imgedit} implements multi-turn editing by dividing the editing process into three sequential iterations, where each step depends on the previous one. Most existing datasets support only single-turn edits, lacking structured and context-aware instruction chains. Additionally, there is often a lack of semantic coherence between consecutive instructions, which can lead to conflicting edits or semantic drift.

\subsubsection{Implicit Editing}
Implicit Editing represents a class of instruction-driven edits where the required visual modification is not explicitly described, but instead inferred through commonsense or world knowledge. This inherently challenges current models due to the need for high-level reasoning and the ambiguity of visual targets.

A promising direction to address these challenges is to introduce explicit intermediate reasoning. For example, AnyEdit~\cite{jiang2025anyedit} employs a reverse construction pipeline: large language models (LLMs) first convert implicit instructions into explicit, executable commands, which are then processed by standard instruction-based editing models.

\subsection{Spatial Understanding and Reasoning}

\subsubsection{Size Adjustment}
The size adjustment task aims to adjust the scale of a specified object within an image according to textual instructions such as ``make the dog smaller". While conceptually straightforward, this task presents challenges in preserving visual realism, particularly in terms of lighting, shadow alignment, and seamless blending with the background.

A practical construction paradigm is a modular, object-centric pipeline. Methods such as Step1X~\cite{liu2025stepxedit} first localize and segment the target, perform inpainting to recover a clean background, and then reinsert a rescaled foreground cutout.

\subsubsection{Referring-based Editing}
Referring-based Editing focuses on accurately locating and editing objects or regions within complex scenes based on natural language expressions that describe them (e.g., ``the man on the left wearing a blue shirt"). 

Early works such as RefCOCO and RefCOCO+ were primarily designed for referring expression segmentation, laying the foundation for understanding language-guided localization. However, these datasets did not support actual image editing.


More recently, RefEdit~\cite{Pathiraja2025RefEditAB} proposes a scalable synthetic data pipeline for this task. Based on RefCOCO, it combines GPT-4o for instruction and expression generation, FLUX for image synthesis, Grounded Segment Anything for masks, and FlowChef or Inpaint Anything for editing, generating about 20k triplets covering color change, object add/remove, and texture modification.

\subsubsection{Viewpoint Transformation}
Viewpoint transformation refers to modifying the image to simulate changes in camera perspective, such as translation, zooming, or rotation. Unlike object-level motion editing, this task emphasizes global structural change that reflects how a scene appears from a different viewpoint.

Considering the difficulty of synthesizing rotated images, existing methods, like Step1X~\cite{liu2025stepxedit}, select multi-view image pairs from the real-world dataset, MVImgNet, to obtain original and edited examples accompanied by instructions such as ``rotate the object clockwise".

\subsubsection{Count Change}
Counting-based editing introduces a unique challenge wherein the model must modify the number of specific objects in an image based on explicit numerical instructions (e.g., ``change the number of apples to four"). While current approaches such as AnyEdit~\cite{jiang2025anyedit} address this by iteratively adding or removing objects conditioned on the instruction prompt, they often fail to ensure appearance consistency between newly generated objects and existing ones. 

The other common strategy kind like movement task, which involves detecting and segmenting existing object instances using tools such as Recognize-Anything and SAM-based variants. These instances are then duplicated and lightly transformed (e.g., through scaling, flipping, or color jittering) to introduce variation while preserving key attributes. The transformed objects are reinserted into the scene at specified locations, and image inpainting techniques are applied to smooth transitions and ensure visual coherence.

\subsubsection{Position Adjustment}
Position adjustment task focuses on modifying the relative spatial positions of objects within an image while preserving all other visual content, such as object appearance and background. A common approach is to first generate a layout from the original caption, then alter this layout by swapping the positions of two target objects to construct the edited layout. This is followed by using a layout-to-image generation model to synthesize the original and edited images, ensuring that only the spatial relationship is changed without affecting other components.

The core challenge of this task lies in precise spatial control and semantic consistency. While layout-based methods provide effective structure-level manipulation, current layout-to-image models still struggle with high-fidelity generation, particularly in maintaining object details consistent with the original image. Moreover, spatial terms in textual instructions (e.g., ``on the left", ``near") often introduce ambiguity depending on context, making it harder for models to interpret intent accurately. 

\begin{figure*}[htbp]
	\centering
	\includegraphics[width=2.2\columnwidth]{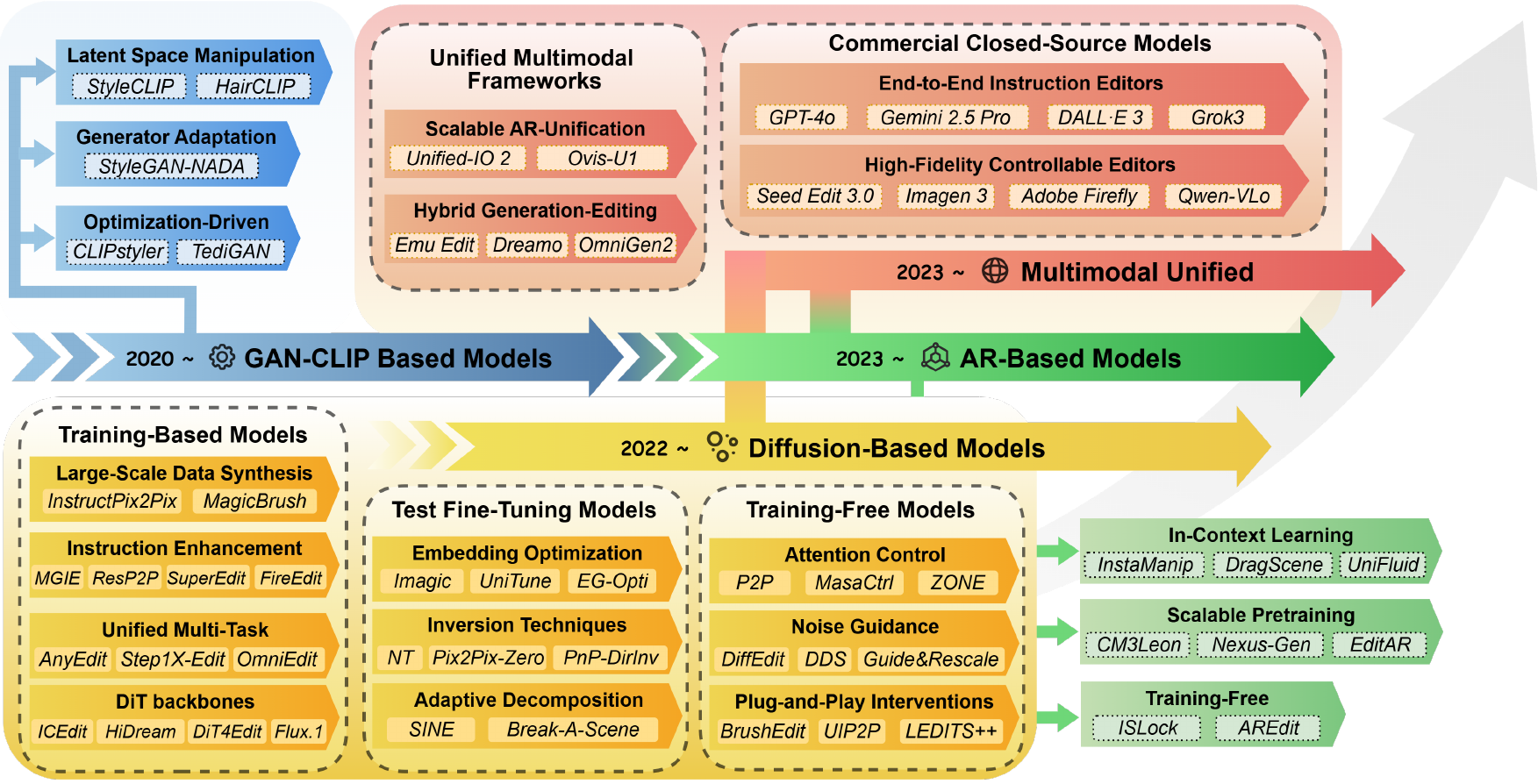}
	\caption{Summarization of representative works}
	\label{fig:framework}
\end{figure*}

\section{Models} \label{secB4}	
As illustrated in Fig.~\ref{fig:framework}, the evolution of Instruction-based Image Editing (IIE) task reflects progressive advancements in generative architectures: early works (2019-2023) predominantly relied on GAN-based latent space manipulations and CLIP-guided optimizations to achieve domain-specific zero-shot editing~\cite{tero2019stylegan,tero2020stylegan2,patashnik2021styleclip,wei2022hairclip,gihyun2022clipstyler,baykal2023clipguided,xia2021tedigan,jiang2021talktoedit}; from 2022 onward, diffusion models have dominated the paradigm, incorporating supervised and unsupervised frameworks to enhance fidelity and generalization~\cite{lv2023controlnet,brooks2023instructpixpix,zhang2023magicbrush,huang2024smartedit,yang2024diffusionmaster,ju2024brushnet,yu2024anyedit,zhao2024ultraedit,geng2024instructdiffusion,jin2024reasonpixpix,artur2024grounded-instructpix2pix,li2025superedit,liu2025stepxedit,ye2025imgedit,kawar2023imagic,mokady2023nulltext,valevski2023unitune,avrahami2023breakascene,jiang2025energyguided,dong2023ptinversion,li2023stylediffusion,wang2023instructedit,parmar2023pix2pixzero,qiao2024baret,meng2022sdedit,hertz2022prompttoprompt,couairon2022diffedit,tumanyan2023plugandplay,titov2024guideandrescale,li2024zone,yaowei2024brushedit,simsar2024uip2p,brack2024ledits,cao2023masactrl,hertz2023delta,ju2024pnpinversion,arar2025negative,wallace2023edict,han2024proxedit,huberman2024ddpminv,song2021ddim,lin2024scheduledit,nie2024}; more recent developments (2023-) have incorporated Diffusion Transformers (DiTs), autoregressive (AR) models, and unified multimodal frameworks, advancing toward scalable, multimodal, and instruction-agnostic systems~\cite{kunyu2025dit4edit,zhou2025fireedit,wei2024omniedit,wang2025unicombine,cai2025hidreami,fu2024guiding,guo2024focus,mao2025ace,yu2023cm3leon,mu2025editar,zhang2025nexusgen,lai2025instamanip,fan2025unified,wang2025trainingfree,tai2025islock,lu2024autoregressive,cheng2024dragsence,lu2024unifiedio,wang2025ovisu,sheynin2024emuedit,labs2025flux,wu2025omnigen,tu2025dreamo,xiao2025omnige,bachmann20244m2i,hu2024instructimagen,phung2024grounded}.  In parallel, commercial closed-source systems exhibit end-to-end engineering advantages in high-fidelity controllable editing and multi-turn instruction following ~\cite{openai2024gpt4o,openai2025gpt4-1,google2025gemini2-5,openai2023dalle3,artificial2025grok3,bytedance2025seededit,google2025imagen3,qwen2025qwen25vlo,adobe2024firefly,google2025gemini2flash,midjourney2024v6,anthropic2024claude35}.

\subsection{Introduction to Models}
\begin{figure*}[htbp]
	\centering
	\includegraphics[width=2.0\columnwidth]{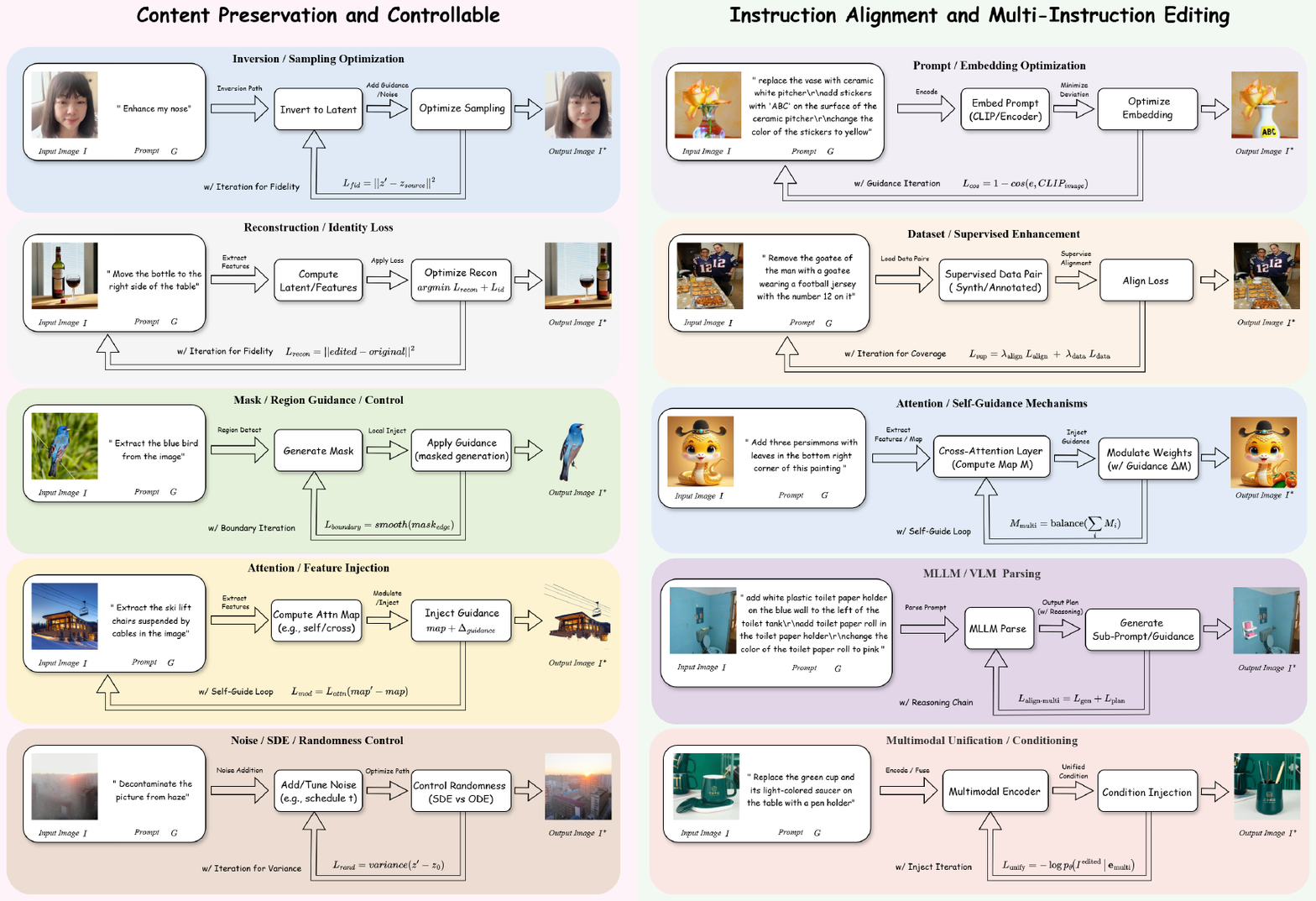}
	\caption{Mechanism Diagram of Solutions for Content Preservation and Instruction Alignment}
	\label{fig:flow1}
\end{figure*}

To move beyond a purely chronological summary, we complement Fig.~\ref{fig:framework} with a mechanism-centric synopsis in Fig.~\ref{fig:flow1}. The diagram organizes representative IIE methods around four core challenges: content preservation, content controllability, instruction alignment, and multi-instruction handling, and aligns them with reusable model families within a unified flow from input, through core operators with optional guidance, to output. Concretely, the left side of Fig.~\ref{fig:flow1} addresses content preservation and controllability through five mechanisms: inversion and sampling optimization; reconstruction and identity losses; mask- or region-based guidance; attention and feature injection; and noise and randomness control using stochastic differential equation paths. The right side emphasizes instruction alignment and multi-instruction editing via prompt and embedding optimization; dataset-driven and supervised enhancement; attention modulation and self-guidance mechanisms; parsing and planning with multimodal large language models and vision and language models; and multimodal unification and conditioning. This alignment of challenges, mechanisms, and architectures clarifies how ideas transfer across model families. For example, combining inversion with masking can be carried over from diffusion samplers to token-level locking in autoregressive models; likewise, instruction parsing (recaptioning and planning) together with unified conditioning enables a closed loop for complex multi-instruction edits within a single backbone.

Guided by this framing, we review more than \textbf{70} studies that adhere strictly to the IIE setting (input: a source image and textual instructions; output: an edited image). The survey spans hybrids of GAN and CLIP, three diffusion paradigms (training based, test time fine-tuning, and training free), and more recent diffusion transformers, autoregressive models, and unified multimodal systems. Commercial systems serve as an end to end reference point. Our taxonomy balances historical progression with strategies for signal integration, such as cross modal attention and token level fusion. Together, Fig.~\ref{fig:framework} and Fig.~\ref{fig:flow1} offer complementary perspectives: the former traces architectural evolution and representative milestones, whereas the latter provides a mechanism level map of how models address the core challenges. 
	
\subsection{GAN-CLIP Based Models}
Early approaches predominantly leveraged generative adversarial networks (GANs), such as StyleGAN~\cite{tero2019stylegan}, StyleGAN2~\cite{tero2020stylegan2}, for latent space manipulation, often integrated with CLIP for cross-modal guidance. These models invert the source image into a GAN's latent space (e.g.,$w$-space), fuse textual instructions via embeddings or gradients, and generate edits through affine transformations or optimization. Signal integration typically occurs in latent vectors using adaptive instance normalization (AdaIN) layers or CLIP similarity losses, such as the directional CLIP loss:
\begin{equation}
	\mathcal{L}_{dir}=1-\frac{\langle\Delta E_{I},\Delta E_{T}\rangle}{\|\Delta E_{I}\|\cdot\|\Delta E_{T}\|},
\end{equation}
where $\Delta E_I$ and $\Delta E_T$ denote differences in CLIP image and text embeddings, enabling zero-shot or domain-specific edits like facial attributes. While efficient for high-resolution outputs, they are limited by domain specificity (e.g., faces, hairs) and optimization instability.

\subsubsection{Latent Space Manipulation}
GAN-centric methods pioneered latent manipulation by embedding text into GAN architectures for controlled edits, emphasizing disentangled shifts to improve content preservation and localized controllability.
	
StyleCLIP~\cite{patashnik2021styleclip} optimizes StyleGAN $w$-space using CLIP similarity gradients for zero-shot text-driven manipulations, ensuring instruction adherence via directional alignment in multimodal space while preserving content through global direction mapping. However, its optimization slowness limits multi-instruction scalability.

HairCLIP~\cite{wei2022hairclip} disentangles hair attributes using StyleGAN2, e4e inversion and sub-mappers, injecting CLIP-encoded text into $w+$ space for decoupled edits. This fine-grained control addresses mask-free localized precision and adherence in domain-specific instructions.

\subsubsection{Optimization-Driven}
CLIP-guided optimization methods extend flexibility by leveraging CLIP multimodal space without full GAN retraining, using iterative gradients or inversions to boost instruction adherence and content fidelity.
	
CLIP-guided StyleGAN Inversion~\cite{baykal2023clipguided} inverts real images into StyleGAN space with CLIP supervision, optimizing $w$-space for text-matched edits. It bridges to real-world applications by balancing reconstruction and CLIP losses for enhanced preservation and controllability, despite domain constraints.

CLIPstyler~\cite{gihyun2022clipstyler} optimizes a lightweight patch-wise style network with CLIP content and style losses for fast text-driven style transfers. Multiview augmentations ensure robust source structure preservation, addressing adherence in artistic edits. However, artifacts may occur in complex scenes.

Talk-to-Edit~\cite{jiang2021talktoedit} enables dialog-style multi-round editing via iterative attention in StyleGAN latent space. Progressive instruction injection models sequential semantics for improved multi-instruction adherence, offering controllability for attribute adjustments, albeit confined to faces.

TediGAN~\cite{xia2021tedigan} unifies StyleGAN2 with cross-modal embeddings, mapping text to $w$-space offsets for multimodal edits. Joint embedding optimization preserves identity and supports diverse instructions. However, computational demands limit it to facial domains.

\subsubsection{Adaptive Fine-Tuning}
Adaptive methods fine-tune GANs guided by CLIP, enabling zero-shot adaptations to generalize instruction adherence across different domains. StyleGAN-NADA~\cite{rinon2022stylegan-nada} adapts StyleGAN domains without training images using CLIP-guided shifts. It expands applicability to open-vocabulary instructions with minimal content distortion. However, prompt specificity affects alignment in disparate domains.

\subsubsection{Comments}
In summary, GAN-CLIP methods laid the foundational groundwork through latent fusion and cross-modal optimization, responding to instruction adherence via similarity losses and content preservation through disentangled latents. This kind of methods catalyzed evolution toward more generalizable paradigms in subsequent diffusion-based models.

\subsection{Diffusion-Based Models}
Diffusion-based models have emerged as the dominant paradigm for IIE since 2022, leveraging the iterative denoising process of U-Net or Diffusion Transformer (DiT) architectures to fuse source images (via encoders or conditioning mechanisms) and textual instructions (through cross-attention or adapters). By progressively refining noisy latents conditioned on multimodal inputs, these methods excel in generating high-fidelity edits while addressing core challenges such as instruction adherence, content preservation, localized precision, and multi-instruction handling. We categorize them based on training dimensions: training-based approaches, testing-time fine-tuning methods and training- and fine-tuning-free techniques.

\subsubsection{Training-Based Models}
Training-based methods involve full supervision or adaptation of pre-trained backbones like Stable Diffusion (SD) or DiTs. These approaches cluster into data synthesis-driven techniques, instruction enhancement via MLLMs, unified multi-task frameworks, modular adaptations, and DiT-centric innovations.

\textbf{Large-scale Data Synthesis.} 
A series of methods focuses on large-scale data synthesis to train generalist models. InstructPix2Pix~\cite{brooks2023instructpixpix}  pioneered this by fine-tuning Stable Diffusion with GPT-3-generated triplets, introducing dual classifier-free guidance (CFG) to balance image fidelity and instruction influence. Signal integration via cross-attention in the U-Net yields versatile edits, though limited by synthetic data noise. Building on this, MagicBrush~\cite{zhang2023magicbrush} curates a manually annotated 10K+ dataset for multi-turn editing, extending InstructPix2Pix with mask support for localized precision. Its innovation lies in high-quality human-verified triplets, improving adherence in sequential instructions, albeit at high annotation cost. InstructDiffusion~\cite{geng2024instructdiffusion} scales to 0.7M triplets across non-face domains, training a generalist interface for universal edits. Cross-attention fusion ensures content preservation, but data noise impacts open-domain generalization.

\textbf{Instruction Enhancement.} 
MLLM-assisted methods enhance reasoning for nuanced instructions. ReasonPix2Pix~\cite{jin2024reasonpixpix} employs LLaVA to infer complex prompts, synthesizing data with GLIGEN for Stable Diffusion fine-tuning. This boosts adherence in semantic edits, though reliant on MLLM accuracy. Grounded-InstructPix2Pix~\cite{artur2024grounded-instructpix2pix} integrates entity grounding (e.g., via DINO detection models) into InstructPix2Pix, automatically generating masks for precise localization without user input. Adapters fuse grounded features, addressing mask-free challenges, but adds computational overhead. SuperEdit~\cite{li2025superedit} rectifies noisy supervision with triplet losses and contrastive learning, enhancing instruction following. Its signal integration via rectified noise prediction preserves structures, limited by dataset scale demands.

\textbf{Unified Multi-Tasking Framework.} Unified multi-task frameworks aim for versatile, high-quality editing. AnyEdit~\cite{yu2024anyedit} unifies 25+ edit types with a 2.5M dataset, introducing task-aware routing and learnable embeddings in Stable Diffusion. This enables ``any idea" edits with cross-attention fusion, excelling in multi-instruction handling, though training-intensive. Step1X-Edit~\cite{liu2025stepxedit} aggregates 11 task datasets with MLLM and DiT, approaching GPT-4o performance in general edits. Token concatenation integrates multi-conditions, promoting scalability, but requires vast dataset resources. ImgEdit~\cite{ye2025imgedit} proposed a 1.2M dataset, fine-tuning VLMs with diffusion backbones for multi-turn edits. Unified pipelines via MLLM parsing enhance adherence, constrained by data curation.

\textbf{Modular Components.} Modular adaptations further refine precision. BrushNet~\cite{ju2024brushnet} decomposes dual-branch diffusion for plug-and-play inpainting, trained on aesthetics datasets. Its feature fusion branches separate localization and generation, aiding mask-free edits. FreeEdit~\cite{he2024freeedit} injects multimodal instructions via adapters, staged training on curated triplets ensures reference-based fusion. This supports diverse modalities, though alignment challenges persist. In-Context Edit~\cite{zhang2025incontext} leverages DiT for in-context generation with AnyInsertion data, enabling zero-shot generalization.

\textbf{DiT Backbone.} DiT-centric innovations emphasize global context. IEAP~\cite{hu2025ieap} decomposes instructions into atomic programs with lightweight DiT adapters for parallel edits. Its program-guided fusion via adapters enhances multi-instruction processing. DiT4Edit~\cite{kunyu2025dit4edit} unifies rigid/non-rigid edits in DiT, injecting text and image guidance. This preserves identities in structural changes. FireEdit~\cite{zhou2025fireedit} parses fine-grained instructions with region-aware VLMs before DiT editing. Its attention modulation refines localization. OmniEdit~\cite{wei2024omniedit} routes tasks to experts in multi-task DiT, supporting multi-domain instructions. The specialist supervision via routing boosts precision. UniCombine~\cite{wang2025unicombine} combines conditions in DiT for identity/subject/style edits. The expert modules fuse signals, though conflicts arise in multi-conditions. HiDream-I1~\cite{cai2025hidreami} integrates T2I and editing in sparse DiT, trained for efficiency. The multimodal fusion elevates generalization. MGIE~\cite{fu2024guiding} via MLLMs decomposes prompts with MLLMs for DiT-like fusion. This refines adherence in multimodal edits. FYI~\cite{guo2024focus} modulates attention for fine-grained/multi-instruction edits. Its transformer blocks inject signals, mitigating conflicts. MagicEdit~\cite{liew2023magicedit} supports temporally coherent video edits, with image variants fitting the IIE task, which sequential consistency preserves frames. Flux-Kontext~\cite{labs2025flux} employs flow matching in dual-stream encoders for semantic fusion. Its attention mechanisms unify text/image streams, though flow computation is intensive. ACE++~\cite{mao2025ace} uses context-aware filling in diffusion frameworks, two-stage training on zero-reference and ACE data. This supports ID/IP preservation in edits.

These training-based methods dominate IIE by harnessing data-driven adaptation to overcome challenges like instruction adherence (via MLLM synthesis) and locating precision (through routing/adapters). Their evolution from SD fine-tuning to DiT unification underscores trends toward scalability and multimodality, though future work must address data efficiency and ethical biases in synthetic triplets.

\subsubsection{Testing-Time Fine-Tuning Models}

Testing-time fine-tuning adapts pre-trained diffusion models per input instance, optimizing embeddings or parameters for personalized edits while preserving source identity. We group these methods into embedding optimization, inversion techniques, and adaptive decomposition for scene-aware fidelity.

\textbf{Embedding Optimization.}
Imagic~\cite{kawar2023imagic} optimizes target embeddings to reconstruct the source, then fine-tunes the diffusion model for high-fidelity edits. This enables text-based real-image modifications without inversion, addressing content preservation, though per-image tuning is computationally heavy.

\textbf{Inversion Techniques.}
Null-Text Inversion~\cite{mokady2023nulltext} uses empty-text inversion to boost fidelity in guided diffusion edits. DDIM inversion followed by optimization preserves details, enhancing adherence in zero-shot settings, limited by iteration speed. UniTune~\cite{valevski2023unitune} fine-tunes on a single image for text-driven variants. The latent direction optimization personalizes edits, supporting style adjustments while maintaining identity.

\textbf{Adaptive Decomposition.}
Zero-Shot Image-to-Image Translation~\cite{parmar2023pix2pixzero} optimizes latent directions via reconstruction losses for domain conversions. This achieves zero-shot edits akin to IIE, with signal integration through denoising guidance. SINE~\cite{bao2023sine} employs semantic-driven NeRF editing with prior-guided fields. The test-time prior optimization encodes 3D edits, using proxy meshes for geometry and feature clustering for preservation, though extended to diffusion for IIE semantics. Break-A-Scene~\cite{avrahami2023breakascene} decomposes single images into concepts, fine-tuning embeddings per concept for multi-object edits. This facilitates instruction-guided manipulations, with attention preserving unrelated areas. Energy-guided optimization~\cite{jiang2025energyguided} uses energy-based tuning of latent features in pre-trained models. This personalizes edits without full retraining, optimizing consistency via guided objectives.

In summary, these methods bridge pre-trained generality with instance-specific adaptation, trending toward efficient few-step tuning.

\subsubsection{Training and Fine-Tuning Free Models}
Training and fine-tuning free models intervene at inference time without parameter updates, relying on modular injections or guidance for zero-shot edits. We cluster them into attention control, noise/delta guidance, and plug-and-play fusions.

\textbf{Attention Control.}
Prompt-to-Prompt (P2P)~\cite{hertz2022prompttoprompt} injects cross-attention maps post-DDIM inversion for partial modifications. Its self-attention preserves structures, addressing multi-instruction via map replacements, though leakage risks unintended changes. MasaCtrl~\cite{cao2023masactrl} tunes mutual self-attention for consistent synthesis. The attention swapping at inference ensures fidelity in sequential edits.

\textbf{Noise Guidance.}
DiffEdit~\cite{couairon2022diffedit} generates masks from noise differences, conditioning denoising on text for regional edits. This implicit masking boosts precision without user input. Delta Denoising Score~\cite{hertz2023delta} guides semantics via delta scores. Pure sampling adjustments enable fine-grained changes. Guide-and-Rescale~\cite{titov2024guideandrescale} rescales noise predictions with self-guidance. This enhances consistency in tuning-free real-image edits.

\textbf{Plug-and-Play Interventions.}
Plug-and-Play Diffusion Features~\cite{tumanyan2023plugandplay} injects source features into text-guided denoising. Its decoder-layer fusion supports image-to-image translations without optimization. ZONE~\cite{li2024zone} derives masks from attention for zero-shot local edits. Its map fusion refines targeting. BrushEdit ~\cite{yaowei2024brushedit} pipelines LLM parsing, SAM masking, and diffusion inpainting. Its end-to-end inference aids complex instructions, dependent on external tools. UIP2P~\cite{simsar2024uip2p} enforces cycle consistency in unsupervised DDIM inversion. This realizes zero-shot edits with loss-guided loops. LEDITS++~\cite{brack2024ledits} extends DDPM inversion with semantic guidance. Its fast solvers reduce steps for real-image edits. Instruct-Imagen~\cite{hu2024instructimagen} decomposes prompts for zero-shot generation. The guided sampling fits IIE variants. Grounded Text-to-Image Synthesis~\cite{phung2024grounded} with Attention Refocusing refocuses maps for localized synthesis. This supports instruction-driven refinements. CycleDiffusion~\cite{wu2023a} applies cycle consistency in latent space for zero-shot editing. Its stochastic guidance preserves identities.

In summary, these methods dominate efficient IIE, trending toward advanced attention/guidance.

\subsection{AR-Based Models}
Autoregressive (AR) models have increasingly complemented diffusion paradigms for IIE since 2023, leveraging sequential token prediction within Transformer architectures to process multimodal inputs as unified sequences. By tokenizing images and instructions into a shared space, these models enable progressive generation conditioned on prior tokens, such as through conditional probability modeling:
\begin{equation}
	p(I_e|I_s,T)=\prod_{t=1}^Np(x_t|x_{t-1},c),
\end{equation}
where $I_e$ is the edited image, $x_t$ are tokenized elements, and $c$ fuses source image $I_s$ and instruction $T$ via token concatenation or attention. 

This facilitates fine-grained control, with advantages in contextual reasoning and compositionality, though limitations include slower inference due to sequential decoding and potential error accumulation in long sequences. In addressing core challenges, AR methods enhance instruction adherence and multi-instruction handling through in-context conditioning, while improving content preservation and controllability via token-level fusion and adaptive prompting.

\subsubsection{Scalable Pretraining-Based Models}
Early AR efforts focused on large-scale pretraining to unify multimodal understanding and generation, emphasizing scalable token prediction for robust instruction adherence. 

CM3Leon~\cite{yu2023cm3leon} scales autoregressive multimodal models through instruction tuning on vast datasets, supporting text-guided edits like caption-based modifications. Its bootstrapped alignments preserve source identity for content fidelity, though pretraining biases limit open-domain multi-instruction generalization. 

EditAR~\cite{mu2025editar} proposes a unified AR model for conditional generation, tokenizing images and instructions for autoregressive prediction of localized/global modifications. Its dynamic routing enhances adherence by fusing conditions, enabling controllability in diverse edits, albeit with sequential slowdowns. 

Nexus-Gen~\cite{zhang2025nexusgen} extends shared embeddings for understanding, generation, and editing, supporting parallel instructions via pre-filled AR. Its token-level integration preserves fine details and addresses multi-instruction conflicts, though computationally intensive for complex scenes.

\subsubsection{In-Context Learning Models}
In-context mechanisms leverage few-shot adaptation within AR flows, prioritizing rapid learning from instructions to boost adherence and localized precision.

InstaManip~\cite{lai2025instamanip} utilizes group self-attention for few-shot manipulation, learning instructions in-context from source images. This supports rapid multi-object edits with attention modulation refining controllability, limited by sample efficiency in diverse domains.

UniFluid~\cite{fan2025unified} introduces continuous tokens in AR frameworks for visual tasks, handling text-conditioned edits autoregressively. Its shared space fusion elevates multimodal synergy and content preservation, though token continuity may introduce artifacts in high-resolution outputs.

DragScene~\cite{cheng2024dragsence} enables interactive 3D scene editing from single-view drag instructions, extending AR to spatial manipulations. Its proxy representations preserve source geometry for fidelity, addressing multi-instruction handling in 3D contexts, but reliant on view consistency.

\subsubsection{Training-Free Efficient Editing Models}
Training-free variants prioritize efficiency in visual AR, intervening at inference for zero-shot edits while maintaining content controllability.

Training-Free Text-Guided Editing~\cite{wang2025trainingfree} conditions VAR sequences on instructions without fine-tuning, enabling zero-shot edits like style transfers. Its direct AR decoding boosts preservation and adherence, though iterative prediction lags behind parallel methods.

Anchor Token Matching~\cite{tai2025islock} employs implicit structure locking via anchor tokens for training-free AR editing. This refines localized precision and multi-instruction adherence through matched conditioning, mitigating error accumulation in open-vocabulary scenarios.

Additional AR innovations explore specialized extensions, such as panoramic generation, to enhance generalization for the IIE task. AOGO~\cite{lu2024autoregressive} employs AR sequences for 360-degree image generation conditioned on instructions, enabling source image outpainting. Its omni-aware tokens ensure structural coherence and content preservation, though limited by resolution in high-fidelity outputs.

\subsubsection{Comments}
In summary, AR-based models improve the edit ability through sequential compositionality and contextual fusion, effectively responding to multi-instruction challenges via in-context mechanisms and content preservation through token-level conditioning. The trends indicate hybrid integrations with diffusion can accelerate the inference process.

\subsection{Multimodal Unified and Commercial Large Models}
Multimodal unified and commercial large models mark a pivotal advancement, consolidating diverse sensory inputs (visual, textual, positional) into cohesive architectures for seamless reasoning and generation. Leveraging shared representational spaces and attention mechanisms, they enable robust cross-modal integration, supporting nuanced edits that preserve source image fidelity while accommodating complex instructions. Key strengths include enhanced generalization and user-centric scalability, with challenges in computational overhead and interpretability, particularly pronounced in closed-source variants. We classify these into unified multimodal frameworks (academic/open-source efforts prioritizing architectural coherence) and commercial closed-source models (proprietary systems optimized for deployment efficiency).

\subsubsection{Unified Multimodal Frameworks}
Unified multimodal frameworks integrate perceptual analysis, semantic interpretation, and generative capabilities into cohesive architectures, frequently drawing on autoregressive or hybrid Transformer structures to handle tokenized inputs across modalities. These systems rely on integrated representational domains and attention layers for effective signal merging, exemplified by autoregressive conditioning over unified tokens:
\begin{equation}
	p(O\mid I_s,T,M)=\prod_{t=1}^Np(o_t\mid o_{t-1},c),
\end{equation}
where $O$ denotes the output sequence (such as an edited image), $o_t$ are cross-modal tokens, and $c$ encapsulates the source image $I_s$, textual instruction $T$, and auxiliary modalities $M$ through fused embeddings. With respect to central challenges, these frameworks refine compliance with directives and sequential command integration via expansive tuning regimens.

\textbf{Scalable AR-Unification}
Efforts in this category emphasize broad pretraining to achieve multimodal cohesion, leveraging autoregressive expansion to sharpen directive fidelity in intricate scenarios. Unified-IO 2~\cite{lu2024unifiedio} advances autoregressive synthesis across visual, linguistic, auditory, and behavioral modalities, encompassing more than 100 functions including directive-based image alteration. Its modality-neutral token scheme supports coherent sequencing to maintain input fidelity amid compound directives, tempered by constraints on sequence depth for detailed renderings. Ovis-U1~\cite{wang2025ovisu} harnesses 3B-parameter Transformers for amalgamated synthesis and modification, employing anticipatory auto-regression within joint manifolds. This facilitates targeted adjustments like scene refinements with preserved elements, bolstered by scale for operational agility in compliance, yet bound by data demands.

\textbf{Hybrid Generation-Editing}
This strand pursues integrated pipelines for task-aligned operations, merging detection and creation to yield accurate directive execution.
Emu Edit~\cite{sheynin2024emuedit} capitalizes on vision-language foundations from over 10 million pairs to enable multifaceted IIE, including entity adjustments. By partitioning into perceptual and productive phases via common encoders, it upholds compositional accuracy and region-specific targeting without masks.
OmniGen2~\cite{wu2025omnigen} investigates sophisticated cross-modal synthesis via Transformer-diffusion amalgamations, directing modifications with combined textual and visual cues. Its comprehensive autoregressive pathways manage layered directives for superior integration and precision, though vulnerable to inherent predispositions in broad training setups.
DreamO~\cite{tu2025dreamo} consolidates personalization through dynamic 3D assembly from video inputs, broadening to IIE with incremental tactics for entity retention. Its perpendicular refinements blend directives to safeguard coherence in application contexts like virtual fitting, despite occasional discrepancies in dynamic-to-static conversions.

In essence, unified multimodal frameworks cultivate synergies between sensing and synthesis via collective manifolds and calibrated datasets, thereby advancing directive fidelity through cross-modal orchestration and input integrity via blended decompositions. As these open architectures mature, they pave the way for proprietary commercial systems that prioritize deployment readiness and user accessibility.

\subsubsection{Commercial Closed-Source Models}
The analysis in this subsection draws primarily from publicly available technical reports~\cite{openai2024gpt4o, google2025gemini2-5, openai2023dalle3, qwen2025qwen25vlo}, official documentation provided on the respective model websites~\cite{google2025imagen3, adobe2024firefly} and third-party user evaluations~\cite{artificial2025grok3}. The objectives are that examining their demonstrated capabilities and underlying design principles rather than speculating on undisclosed internal mechanisms.

Proprietary closed-source models signify a maturation in IIE applications, providing accessible platforms equipped with intuitive controls and compliance features. These implementations generally embed advanced vision-language processors alongside generative engines, enabling directive-driven modifications through optimized, albeit opaque, integration pathways. Notable strengths include practical scalability and built-in safeguards.

\textbf{End-to-End Instruction Editors}
Models in this vein function as adaptable processors for intent-focused operations, incorporating broad sensory inputs to enable fluid, directive-centric refinements.
GPT-4o~\cite{openai2024gpt4o} assimilates visual, textual, and auditory elements to perform IIE functions, such as compositional tweaks via generative linkages. Its handling of extended sequences aids in maintaining intent across revisions, with embedded protocols ensuring balanced outcomes.
Gemini 2.5 Pro~\cite{google2025gemini2-5} streamlines evaluations for intent-based alterations. Its condensed architectures deliver swift execution, augmented by internal validations for focused reliability.
Grok 3~\cite{artificial2025grok3} accommodates cross-modal interactions pertinent to IIE, supporting intent-oriented shifts. The external benchmarks underscore its interpretive strengths, contributing to consistent handling of layered intents through implicit harmonizations, tempered by subscription frameworks.

\textbf{High-Fidelity Controllable Editors}
This segment concentrates on precision-oriented synthesizers, geared toward intent-guided creation.
SeedEdit 3.0~\cite{bytedance2025seededit} enables rapid component substitutions in creative workflows. Embedded tools optimize user interactions, reinforcing intent realization in multifaceted settings.
Imagen 3~\cite{google2025imagen3} operates within interconnected suites for intent-amplified detail. Its advanced curation preserves unity, fostering integrated modalities.
Qwen2.5-Vlo~\cite{qwen2025qwen25vlo} processes vision-language tasks for descriptive transformations. The mitigation strategies uphold progression stability, promoting versatile implementation.
Adobe Firefly~\cite{adobe2024firefly} embeds within design ecosystems for intent-led completions. Its format-aware elements guarantee exactness in tailored applications.

Commercial closed-source models ultimately broaden IIE accessibility via refined ecosystems and protective measures.

\section{Evaluation} \label{secB5}
In the domain of instruction-based image editing, the evolution of evaluation methods has progressed from objective, computable metrics to subjective, human-aligned judgments, culminating in complex hybrid evaluation systems. This section systematically analyzes the metrics used for assessment, the benchmarks that provide the testing grounds. At last, we propose a benchmark for Comprehensive, in-Depth, and Diagnostic evaluation (CDD-IIE Benchmark).

\subsection{Traditional Metrics}
The ultimate objective of instruction-based image editing is to precisely and faithfully modify an image according to a given text prompt, while maximally preserving the content and structure of the original, unedited regions. Consequently, the evaluation of such systems is typically bifurcated into two principal axes of assessment: 1) instruction adherence, which quantifies the fidelity to the instruction; 2) consistency preservation, which measures the preservation of source content.

\subsubsection{Instruction Adherence}
This category of metrics evaluates the model's fidelity to the instructional prompt, $T_{edit}$. A successful edit must not only preserve the original content but also accurately reflect the changes specified in the text.

\textbf{CLIP Directional Similarity.}
This is arguably the most critical metric for instruction-based editing. It assesses whether the semantic change in the image space aligns with the semantic change described in the text space. It requires a source text prompt ($T_{src}$, e.g., ``a photo of a horse") and a target text prompt ($T_{gt}$, e.g., ``a photo of a zebra"). The metric computes the \emph{cosine similarity} between the image-embedding delta $\Delta E_I$ and the text-embedding delta $\Delta E_T$:
\begin{equation}
	S_{\text{CLIP-Dir}} = \frac{\Delta E_I \cdot \Delta E_T}{\|\Delta E_I\|_2 \cdot \|\Delta E_T\|_2}.
\end{equation}

\textbf{CLIP Target Similarity.}
This metric directly measures the semantic similarity between the final edited image $I_{edit}$ and the target text description $T_{tgt}$. Let $E_I$ and $E_T$ be the CLIP image and text encoders, respectively:
\begin{equation}
	S_{\text{CLIP-out}} = \frac{E_I(I_{edit}) \cdot E_T(T_{tgt})}{\|E_I(I_{edit})\|_2 \cdot \|E_T(T_{tgt})\|_2}.
\end{equation}

\textbf{CLIP Ground Truth Similarity.}
In benchmarks where a ground-truth (GT) edited image $I_{gt}$ is available, this metric evaluates the similarity between the model output $I_{edit}$ and the desired output $I_{gt}$: 
\begin{equation}
	S_{\text{CLIP-GT}} = \frac{E_I(I_{edit} ) \cdot E_I(I_{gt} )}{\|E_I(I_{edit} )\|_2 \cdot \|E_I(I_{gt} )\|_2}.
\end{equation}

The comparison is often restricted to the foreground edited region, to determine if the edited content visually aligns with the ground truth.

\subsubsection{Consistency Preservation}
Image faithfulness metrics quantify the degree to which an edited image, $I_{edit}$, retains the structural and semantic integrity of the source image, $I_{src}$. These metrics are crucial for penalizing unwanted or hallucinatory modifications outside the scope of the edit instruction. They can be broadly categorized into structural and semantic indicators.

\textbf{Structural Metrics.}
The L1 Pixel Distance metric provides a direct measure of the absolute difference in pixel values between the source and edited images. A lower L1 distance signifies a smaller low-level change, indicating that the overall image structure is well-preserved. Given a source image $I_{src}$ and an edited image $I_{edit}$, the L1 distance is formulated as:
\begin{equation}
	\mathcal{L}_{L1} = |I_{src}- I_{edit}|.
\end{equation}
Structural Similarity Index Measure (SSIM) is a perception-based model that evaluates image similarity by considering three key components: luminance, contrast, and structure. Unlike traditional metrics that rely on pixel-by-pixel comparisons, SSIM models how the human visual system perceives images. It computes a similarity score by first evaluating these components within local image windows, then averaging the results. This approach makes SSIM a more nuanced measure of overall image structure, which aligns better with human visual perception than simpler metrics like L1 distance. In the context of instruction-based editing, SSIM is used to assess the structural consistency between the original and edited images.

\textbf{Semantic Metrics.}
Learned Perceptual Image Patch Similarity (LPIPS) is a metric that measures the perceptual similarity between two images. Unlike traditional methods that compare pixel values directly, LPIPS computes the distance between images in a learned deep feature space. This is achieved by extracting feature activations from a pre-trained deep neural network (e.g., AlexNet or VGG), a process that has been shown to correlate strongly with human perceptual judgment. In the context of instruction-based editing, LPIPS is used to assess the perceptual consistency between the original and edited images.

CLIP Image Similarity (CLIP-img) metric leverages the joint text-image embedding space of CLIP (Contrastive Language-Image Pre-training) to measure high-level semantic similarity. The \emph{cosine similarity} between the CLIP image embeddings of the source and edited images is computed. A high score indicates that the edited image remains semantically close to the original. Let $E_I$ be the CLIP image encoder:
\begin{equation}
	S_{\text{CLIP-img}} = \frac{E_I(I_{src}) \cdot E_I(I_{edit})}{\|E_I(I_{src})\|_2 \cdot \|E_I(I_{edit})\|_2}
\end{equation}

Similar in principle to CLIP-img, DINO similarity uses image features extracted from a DINO-pretrained Vision Transformer (ViT). DINO features are known to capture fine-grained semantic and structural information. The similarity is the \emph{cosine distance} between the DINO embeddings of the two images:
\begin{equation}
	S_{\text{DINO}}= \frac{E_{\phi}(I_{src}) \cdot E_{\phi}(I_{edit})}{\|E_{\phi}(I_{src})\|_2 \cdot \|E_{\phi}(I_{edit})\|_2}
\end{equation}
where $E_{\phi}$ is the DINO image encoder.

\subsubsection{Comments}
While these metrics provide a foundational framework for quantitative evaluation, they suffer from three inherent limitations:
\begin{itemize}
	\item [1)] There is an inherent trade-off among conflicting metrics. A fundamental tension exists between metrics of image faithfulness and those of text alignment. For instance, a successful edit that significantly alters an image (e.g., object replacement) will necessarily perform poorly on faithfulness metrics like L1 distance, even as it achieves a high score on alignment metrics like CLIP-Score.
	\item [2)] These metrics exhibit a poor correlation with human preferences. AnyEdit~\cite{yu2024anyedit} and ImgEdit~\cite{ye2025imgedit}  have demonstrated that conventional similarity metrics, whether pixel-based or feature-based, often fail to align with human perceptual judgments of editing quality and success.
	\item [3)] They are incapable of capturing the semantic nuances of complex edits. These metrics are often semantically impoverished, rendering them ill-equipped to evaluate high-level, sophisticated editing tasks~\cite{Ma2024I2EBenchAC}. For example, they cannot reliably verify the success of operations involving numeracy ``change the two balloons to three"), spatial awareness (``move object A to the left of object B"), or discerning edits like object removal and background replacement where significant, intentional pixel changes are the desired outcome.
\end{itemize}
These profound limitations underscore the urgent need for an evaluation paradigm endowed with a higher-level, human-like understanding of both visual content and instructional intent.

\subsection{VLM-Based  Metrics}\label{sec:vlm_metrics}
The recent ascendancy of Vision-Language Models (VLMs)~\cite{bai2025qwen2,li2024llava,liu2023visual}, has catalyzed a profound paradigm shift in the evaluation of instruction-based image editing. To overcome the inherent limitations of traditional metrics which primarily assess low-level pixel similarities, researchers have begun to leverage the advanced cognitive, reasoning, and visual understanding capabilities of contemporary VLMs. The core innovation lies in reframing the evaluation task itself as a sophisticated form of Visual Question Answering (VQA), where the model assesses the quality of an edit based on a holistic understanding of semantics and aesthetics.

A seminal contribution in this domain is VIEScore~\cite{Ku2023VIEScoreTE}, a pioneering framework that utilizes general-purpose Multimodal Large Language Models (MLLMs), such as GPT-4V~\cite{openai2024gpt4o}, as zero-shot evaluators. The evaluation process involves presenting the MLLM with a composite input, including the original image, the editing instruction, and the generated image. The model is then prompted to produce both a quantitative score and a natural language rationale that justifies its assessment in a manner consistent with human logic. The evaluation rubric for VIEScore is structured along two primary dimensions:
\begin{itemize}
	\item Perceptual Quality (PQ): This dimension assesses the overall visual appeal of the edited image, focusing on its naturalness and the absence of distracting artifacts.
	\item Semantic Consistency (SC): This dimension evaluates the degree to which the edited image faithfully adheres to the instruction, while also ensuring that there are no excessive or unintended modifications (i.e., over-editing).
\end{itemize}

The principal advantages of VIEScore are its capacity for high-level, explainable assessment, its strong alignment with human perceptual judgments, and its versatility across a wide spectrum of editing tasks, from complex semantic manipulations (e.g., object addition, style transfer) to low-level enhancements (e.g., deblurring, dehazing).

Inspired by the groundwork laid by VIEScore, subsequent research~\cite{liu2025stepxedit,Ma2024I2EBenchAC,Jia2025CompBenchBC,zhang2025incontext} has widely adopted more powerful proprietary models, such as GPT-4o~\cite{openai2024gpt4o} and Gemini~\cite{google2025gemini2flash}, to serve as evaluators, leading to the development of more refined and rigorous frameworks. ImgEdit~\cite{ye2025imgedit} is a prominent example of this evolution. Its accompanying benchmark ImgEdit-Bench, employs GPT-4o to conduct a meticulous, multi-faceted evaluation. For a given editing task defined by an input tuple: comprising the original image ($I_{\text{src}}$), the edited image ($I_{\text{edit}}$), and the textual instruction ($P_{\text{instr}}$)--- it assesses the output along three core axes, with scores ranging from 1 to 5 based on carefully designed rubrics:
\begin{itemize}
	\item Instruction Adherence ($S_{\text{adh}}$): Measures how precisely $I_{\text{edit}}$ fulfills the requirements specified in $P_{\text{instr}}$.
	\item Edit Quality ($S_{\text{qua}}$): Evaluates the visual fidelity, realism, and accuracy of the modified regions within $I_{\text{edit}}$.
	\item Detail Preservation ($S_{\text{pres}}$): Assesses whether the non-edited regions of $I_{\text{edit}}$ remain intact and free from unintentional artifacts or alterations.
\end{itemize}
The final score is computed as the average of these three components:
\begin{equation}
	S_{\text{final}} = \frac{S_{\text{adh}} + S_{\text{qua}} + S_{\text{pres}}}{3}
	\label{eq:imgedit_score}
\end{equation}

\subsection{Human Evaluation}
Despite significant advancements in automated evaluation for image editing, Human Evaluation remains the ``gold standard" for assessing the quality of edited images~\cite{sheynin2024emuedit,huang2025diffusion}. This direct assessment method provides nuanced judgments that automated approaches often struggle to capture. Currently, the following human evaluation methods are prevalent in the image editing field:
\begin{itemize}
	\item One-on-One Comparison: This method offers a detailed and subtle assessment by presenting an edited result alongside a strong baseline. Evaluators choose their preferred option based on consistency and image quality.
	\item Multiple-Choice Comparison: In this task, evaluators select the best image from several edited images based on consistency and image quality.
	\item Individual Evaluation: Individual evaluation typically uses a 5-point scale to gather subjective feedback on the quality of images generated by different models. 
\end{itemize}

While the focus of different studies may lead to variations in evaluation dimensions, human evaluation generally centers on the following three core aspects:
\begin{itemize}
	\item Instruction Following: Assesses the extent to which the edited result accurately adheres to the given editing instructions.
	\item Consistency Between Edited and Original: Examines the coherence of the edited image with the original in terms of content and style, ensuring that the edit does not introduce unnatural artifacts or inconsistent elements.
	\item Image Quality: Evaluates the overall visual quality of the edited image, including clarity, naturalness, color accuracy, and the absence of distortion, noise, or other visual defects.
\end{itemize}

\subsection{Benchmarks}\label{sec:benchmarks}
The rapid development of models has a critical need for comprehensive and standardized benchmarks to effectively assess and compare their performance. To address these challenges, a growing number of benchmarks have been proposed, specifically designed for evaluating IIE models~\cite{Ma2024I2EBenchAC,zhang2023magicbrush,sheynin2024emuedit,liu2025stepxedit,ye2025imgedit}. In this section, we review the prominent evaluation benchmarks in the field, which can be broadly classified into two primary categories: (1) general-purpose benchmarks targeting foundational, atomic editing tasks, and (2) specialized benchmarks designed to probe advanced capabilities such as compositional complexity, reasoning, and performance on specific hard tasks.

\subsubsection{General-Purpose Benchmarks for Foundational Editing Tasks}
This category comprises benchmarks that aim to provide a broad assessment of a model's core editing capabilities across a wide range of common tasks. Early efforts in this area, such as EditBench~\cite{Wang2022ImagenEA}, TEdBench~\cite{kawar2023imagic} and EditEval~\cite{Basu2023EditValBD}, were primarily task-oriented. They focused on canonical sub-tasks like inpainting, attribute manipulation, or layout adjustment, establishing a baseline for structured evaluation.

As the field matured, the focus shifted towards accommodating more flexible and descriptive free-form instructions. MagicBrush~\cite{zhang2023magicbrush}, for instance, was a significant step forward, encompassing a variety of scenarios, including single-turn and multi-turn conversational edits, as well as tasks both with and without user-provided masks. 

More recent endeavors have focused on creating comprehensive and large-scale evaluation suites. Benchmarks such as I2EBench~\cite{Ma2024I2EBenchAC}, AnyEdit~\cite{yu2024anyedit}, ImgEdit-Bench~\cite{ye2025imgedit}, and ICE-Bench~\cite{Pan2025ICEBenchAU} have significantly scaled up the number of samples, the variety of editing types, and the granularity of evaluation metrics. 
A notable trend is the increasing emphasis on realism, with benchmarks like EMU-Edit~\cite{sheynin2024emuedit}, GEdit-Bench~\cite{liu2025stepxedit}, RealEdit~\cite{Sushko2025REALEDITRE}, and SuperEdit~\cite{li2025superedit} curated diverse sets of real-world user instructions and real-world images, moving away from synthetic or template-based prompts.

\subsubsection{Specialized Benchmarks for Advanced Capabilities}
Alongside general-purpose evaluation, a second category of benchmarks has emerged to probe the more advanced and nuanced capabilities of IIE models. These are designed to test the limits of models in domains that require more than simple object or style manipulation. They can be further subdivided based on their specific focus:

\textbf{Compositional Complexity.} Benchmarks such as CompBench~\cite{Jia2025CompBenchBC}, Complex-Edit~\cite{Yang2025ComplexEditCI}, and ComplexBench-Edit~\cite{Wang2025ComplexBenchEditBC} have been developed to assess a model's proficiency in managing instructions that involve multiple objects and attributes, as well as compositional instructions with multiple sub-commands that can be either dependent or independent. These benchmarks present complex compositional instructions that test a model's ability to parse and execute multifaceted requests accurately.

\textbf{Visual Reasoning.} 
A frontier in IIE is the ability to perform edits that require deeper semantic understanding and reasoning. Benchmarks such as Reason-Edit~\cite{huang2024smartedit}, RISE-Bench~\cite{Zhao2025EnvisioningBT}, and KRIS-Bench~\cite{Wu2025KRISBenchBN} are designed for this purpose. They feature tasks that necessitate spatial, causal, or counterfactual reasoning, moving beyond simple text-image matching to assess a model's cognitive understanding.

\textbf{Specific Hard Tasks.} 
A third group of specialized benchmarks targets known failure modes or particularly challenging sub-domains. For instance, ByteMorph~\cite{Chang2025ByteMorphBI} focuses specifically on the challenging task of editing non-rigid movements in both human and objects. RefEdit-Bench~\cite{Pathiraja2025RefEditAB} evaluates reference-based editing, assessing the ability to accurately identify the entity of interest and perform accurate edits based on referring expressions. Lastly, Aurora-Bench~\cite{Krojer2024LearningAA} focuses on action-centric edits by leveraging curated triplets from videos and simulations.

\subsubsection{Comments}
In summary, the landscape of IIE evaluation has evolved from general-purpose datasets for foundational tasks to a diverse ecosystem of specialized benchmarks aimed at probing advanced capabilities. Both categories are indispensable for the holistic advancement of the field.

\subsection{CDD-IIE Bench}
\subsubsection{Motivation and Design Principles}
A systematic analysis of prominent existing benchmarks for instruction-based image editing reveals two critical deficiencies in the current evaluation landscape. First, existing benchmarks exhibit a \textbf{fragmented coverage} of the task space. Each benchmark, while addressing certain common editing tasks, only represents a subset of all possible operations~\cite{liu2025stepxedit,yu2024anyedit,ye2025imgedit}. This fragmentation precludes a truly holistic assessment of a model's capabilities. Second, there is a \textbf{lack of dimensional depth}. A unified framework capable of systematically evaluating both foundational atomic editing capabilities and advanced compositional skills is notably absent. This limitation hinders a deep diagnosis of a model's operational ceiling and specific failure modes.

To address these shortcomings, we introduce a benchmark for Comprehensive, in-Depth, and Diagnostic evaluation (CDD-IIE Bench). It is built upon two core design principles:

\begin{itemize}
	\item Comprehensiveness: We aim to construct a systematic and broad-coverage task collection that spans the full spectrum of common instruction editing operations. This principle directly addresses the coverage deficit of prior works.
	\item In-depth and Diagnosability: We focus on probing the upper echelons of model performance through complex tasks that demand multi-step reasoning and the composition of skills. This allows for a precise diagnosis of a model's strengths and weaknesses when faced with challenging instructions.
\end{itemize}

In line with these principles, our benchmark is structured into two primary suites: one for testing basic atomic editing capability and another for advanced compositional editing capability, which is consistent with Tab.~\ref{tab:CDD-Bench_results_basic} and Tab.~\ref{tab:CDD-Bench_results_advanced} in Sec.~\ref{secB2}. This dual-suite structure ensures that our evaluation is both broad in scope and deep in its diagnostic power, facilitating a comprehensive and insightful assessment of model performance.

\subsubsection{Benchmark Composition}
Our benchmark comprises 1,353 meticulously curated image-instruction pairs, categorized into two main suites that cover 5 major evaluation dimensions and 21 specific editing tasks. These pairs are drawn from a diverse array of sources, including established datasets such as Gedit-Bench~\cite{liu2025stepxedit}, I2EBench~\cite{Ma2024I2EBenchAC}, ImgEdit-Bench~\cite{ye2025imgedit}, ComplexBench-Edit~\cite{Wang2025ComplexBenchEditBC}, CompBench~\cite{Jia2025CompBenchBC}, KRIS-Bench~\cite{Wu2025KRISBenchBN},  AnyEdit~\cite{yu2024anyedit}, ByteMorph~\cite{Chang2025ByteMorphBI}, Reason-Edit~\cite{huang2024smartedit} and RefEdit-Bench~\cite{Pathiraja2025RefEditAB}, and also collected image-instruction pairs from internet by ourselves. 

Each image-instruction pair is carefully manually selected to ensure it is highly distinctive, challenging, and diverse, specifically to expose the common failure modes of current models. After the curation, we hire 12 professional annotators to further verify the quality of each image-instruction pair to ensure its correctness and clarity. Each category contains average 63 image-instruction pairs. The maximum and minimum number of these editing tasks are 150 and 25. We will open-source the benchmark to facilitate future research.

\begin{table*}[h]
	\centering
	\caption{Comparison of CDD-IIE Bench with Existing Image Editing Benchmarks.} \label{tab:benchmark_comparison}
	\begin{tabular}{lcccc}
		\toprule
		\textbf{Benchmark} & \textbf{Publication} & \textbf{Total Tasks}& \textbf{Dataset Size} & \textbf{\begin{tabular}[c]{@{}c@{}}Atomic and \\ Compositional\end{tabular}} \\
		\midrule
		MagicBrush~\cite{zhang2023magicbrush} & NeurIPS 2023 & 5 & 1,053 & $\times$ \\
		Emu-Edit Bench~\cite{sheynin2024emuedit} & CVPR 2024 & 7 & 3,055 & $\times$ \\
		AnyEdit-Test~\cite{yu2024anyedit} & CVPR 2025 & 25 & 1,250 & $\times$ \\
		GEdit-Bench~\cite{liu2025stepxedit} & arXiv 2025 & 11 & 606 & $\times$ \\
		ImgEdit-Bench~\cite{ye2025imgedit} & arXiv 2025 & 14 & 811 & \checkmark \\
		\rowcolor{gray!15}
		\textbf{CDD-IIE Bench} & \textbf{-} & \textbf{20} & \textbf{1,353} & \textbf{\checkmark} \\
		\bottomrule
	\end{tabular}
\end{table*}

In summary, unlike existing benchmarks, CDD-IIE Bench provides a structured evaluation of both atomic and compositional capabilities and features a compact, yet comprehensive, evaluation set. This offers three salient advantages, as illustrated in Table~\ref{tab:benchmark_comparison}: 
\begin{itemize}
	\item [1)] Comprehensiveness, ensuring a wide-ranging assessment across diverse editing tasks.
	\item [2)] Depth and Diagnosability, our dual-oriented evaluation suites enabling a holistic, structured analysis of state-of-the-art model capabilities.
	\item [3)] Efficiency, as its medium-sized, ``concentrated" nature facilitates rapid yet thorough evaluation of a model's core instruction-following abilities, reducing the computational overhead associated with large-scale benchmarking.
\end{itemize}

\begin{sidewaystable}
	\centering
	\caption{Evaluation results on basic atomic suite of CDD-IIE Bench. We evaluated the performance of 10 models on 13 atomic-level editing tasks, covering object-level operations, image-level operations, and utility tasks. The models are organized based on their publication dates. Best results are shown in bold.}\label{tab:CDD-Bench_results_basic}
	\begin{tabular*}{\textwidth}{@{\extracolsep{\fill}}llccccccccccccc}
		\toprule
		& & \multicolumn{8}{c}{\textbf{Object-Level Operations}} & \multicolumn{3}{c}{\textbf{Image-Level Operations}} & \multicolumn{2}{c}{\textbf{Utility Tasks}} \\
		\cmidrule(r){3-10} \cmidrule(r){11-13} \cmidrule(l){14-15}
		\textbf{Model} &
		\textbf{Venue} &
		Add.\textsuperscript{1} &
		Rm.\textsuperscript{2} &
		Repl.\textsuperscript{3} &
		Attr.\textsuperscript{4} &
		Text\textsuperscript{5} &
		Port.\textsuperscript{6} &
		Mot.\textsuperscript{7} &
		Extr.\textsuperscript{8} &
		Tone\textsuperscript{9} &
		Style\textsuperscript{10} &
		BG.\textsuperscript{11} &
		Rep.\textsuperscript{12} &
		VFX.\textsuperscript{13} \\
		\midrule
		Flowedit~\cite{kulikov2025flowedit} & ICCV 2025 & 2.61 & 2.39 & 2.61 & 2.77 & 2.77 & 3.04 & 2.52 & 1.66 & 2.64 & 2.52 & 2.74 & 2.47 & 2.58 \\
		IC-Edit~\cite{zhang2025incontext} & NeurIPS 2025 & 2.78 & 2.89 & 2.95 & 3.28 & 2.62 & 2.62 & 2.43 & 1.51 & 3.36 & 3.10 & 3.08 & 2.32 & 2.37 \\
		OmniGen2~\cite{wu2025omnigen} & arXiv 25.06 & 3.10 & 3.04 & 3.22 & 3.25 & 3.24 & 3.22 & 2.83 & 1.67 & 3.33 & 3.51 & 3.15 & 2.82 & 3.15 \\
		Ovis-U1~\cite{wang2025ovisu} & arXiv 25.06 & 2.53 & 2.78 & 2.64 & 3.22 & 2.82 & 2.26 & 2.11 & 2.05 & 3.60 & 3.51 & 2.47 & 2.39 & 3.00 \\
		X2Edit~\cite{ma2025x2edit}& arXiv 25.06 & 2.97 & 3.40 & 3.34 & 3.32 & 3.19 & 2.62 & 2.32 & 2.14 & 3.32 & 3.32 & 3.33 & 2.85 & 1.42 \\
		Qwen-Edit\textsuperscript{14}~\cite{wu2025qwenimagetechnicalreport} & arXiv 25.09 & \textbf{3.82} & \textbf{3.99} & \textbf{3.76} & \textbf{3.98} & \textbf{4.41} & 3.60 & \textbf{3.58} & 3.08 & 3.99 & 4.07 & \textbf{3.85} & 3.63 & 3.68 \\
		DreamOmni2~\cite{xia2025dreamomni2} & arXiv 25.10 & 3.35 & 3.06 & 3.02 & 3.50 & 3.56 & 3.37 & 2.97 & 1.80 & 3.38 & 3.37 & 3.26 & 3.07 & 3.53 \\
		Uniworld-V2~\cite{li2025uniworldv2} & arXiv 25.10 & 3.70 & 3.81 & 3.72 & 3.95 & 4.22 & 3.27 & 3.50 & \textbf{3.68} & \textbf{4.01} & \textbf{4.11} & 3.75 & \textbf{3.77} & \textbf{3.90} \\
		Step1X-Edit-v1.2~\cite{liu2025stepxedit} & arXiv 25.11 & 3.72 & 3.88 & 3.74 & 3.87 & 4.23 & \textbf{3.69} & 3.42 & 2.87 & 3.84 & 3.91 & 3.59 & 3.49 & 3.73 \\
		Flux.2-dev~\cite{flux-2-2025} & arXiv 25.11 & 3.43 & 3.68 & 3.63 & 3.77 & 3.91 & 3.54 & 3.31 & 2.65 & 3.91 & 3.71 & 3.56 & 3.51 & 3.60 \\
		\bottomrule
	\end{tabular*}
	\footnotesize
	The full names of the corresponding tasks as shown in the table are:
	\textsuperscript{1}Addition,
	\textsuperscript{2}Removal,
	\textsuperscript{3}Replacement,
	\textsuperscript{4}Attribute Modification,
	\textsuperscript{5}Text Modification,
	\textsuperscript{6}Portrait Enhancement,
	\textsuperscript{7}Motion Change,
	\textsuperscript{8}Extraction,
	\textsuperscript{9}Tone Transformation,
	\textsuperscript{10}Style Transfer,
	\textsuperscript{11}Background Change,
	\textsuperscript{12}Image Repair,
	\textsuperscript{13}Visual Effect Removal,
	\textsuperscript{14}Qwen-Image-Edit-2509.
\end{sidewaystable}

\begin{sidewaystable}
	\centering
	\caption{Evaluation results on advanced compositional suite of CDD-IIE Bench. We evaluated the performance of 10 models on 8 tasks, categorized into complex instruction following and spatial understanding. The models are organized based on their publication dates. Best results are shown in bold.}\label{tab:CDD-Bench_results_advanced}
	\begin{tabular*}{\textwidth}{@{\extracolsep{\fill}}llcccccccc}
		\toprule
		& & \multicolumn{3}{c}{\textbf{Complex Instruction \& Reasoning}} & \multicolumn{5}{c}{\textbf{Spatial Understanding \& Reasoning}} \\
		\cmidrule(r){3-5} \cmidrule(l){6-10}
		\textbf{Model} &
		\textbf{Venue} &
		Parallel\textsuperscript{14} &
		Seq.\textsuperscript{15} &
		Implicit\textsuperscript{16} &
		View\textsuperscript{17} &
		Refer\textsuperscript{18} &
		Position\textsuperscript{19} &
		Count\textsuperscript{20} &
		Size\textsuperscript{21} \\
		\midrule
		Flowedit~\cite{kulikov2025flowedit} & ICCV 2025 & 2.40 & 2.66 & 2.86 & 2.44 & 2.22 & 2.71 & 3.26 & 2.77 \\
		IC-Edit~\cite{zhang2025incontext} & NeurIPS 2025 & 2.55 & 2.86 & 3.54 & 2.44 & 2.70 & 2.52 & 3.00 & 2.37 \\
		OmniGen2~\cite{wu2025omnigen} & arXiv 2506 & 2.73 & 3.04 & 3.34 & 2.80 & 2.98 & 2.72 & 2.95 & 2.61 \\
		Ovis-U1~\cite{wang2025ovisu} & arXiv 2506 & 2.26 & 3.03 & 3.20 & 2.42 & 2.19 & 2.27 & 2.98 & 2.73 \\
		X2Edit~\cite{ma2025x2edit}& arXiv 25.06 & 1.99 & 3.05 & 2.82 & 2.41 & 2.39 & 2.14 & 2.25 & 2.37 \\
		Qwen-Edit\textsuperscript{22}~\cite{wu2025qwenimagetechnicalreport} & arXiv 2509 & \textbf{3.50} & \textbf{3.86} & 3.90 & \textbf{3.42} & 3.64 & \textbf{3.58} & \textbf{3.74} & \textbf{2.88} \\
		DreamOmni2~\cite{xia2025dreamomni2} & arXiv 2510 & 2.60 & 3.28 & 3.58 & 2.61 & 3.21 & 2.75 & 3.58 & 2.80 \\
		Uniworld-V2~\cite{li2025uniworldv2} & arXiv 2510 & 3.45 & 3.85 & 3.80 & 3.30 & 3.43 & 3.46 & 3.55 & 2.71 \\
		Step1X-Edit-v1.2~\cite{liu2025stepxedit} & arXiv 2511 & 3.31 & 3.66 & \textbf{4.05} & 3.05 & \textbf{3.72} & 3.27 & 3.68 & 2.87 \\
		Flux.2-dev~\cite{flux-2-2025} & arXiv 2511 & 3.26 & 3.72 & 4.04 & 3.00 & 3.41 & 3.11 & 3.62 & 2.80 \\
		\bottomrule
	\end{tabular*}
	\footnotesize
	The full names of the corresponding tasks as shown in the table are:
	\textsuperscript{14}Parallel Instructions,
	\textsuperscript{15}Sequential Instructions,
	\textsuperscript{16}Implicit Reasoning,
	\textsuperscript{17}Viewpoint Transformation,
	\textsuperscript{18}Referring-based Editing,
	\textsuperscript{19}Position Adjustment,
	\textsuperscript{20}Count Change,
	\textsuperscript{21}Size Adjustment,
	\textsuperscript{22}Qwen-Image-Edit-2509.
\end{sidewaystable}

\begin{table*}
	\centering
	\caption{Summary of Model Performance on CDD-IIE Bench. This table presents the average scores across five dimensions for two evaluation suites. The ``Avg." column shows the averaged score of each suite, while the final column presents the overall average score. Best results are shown in bold.}\label{tab:CDD-Bench_results_summary}
	\begin{tabular*}{\textwidth}{@{\extracolsep{\fill}}lcccccccc}
		\toprule
		& \multicolumn{4}{c}{\textbf{Basic Atomic Eval}} & \multicolumn{3}{c}{\textbf{Advanced Compositional Eval}} & \textbf{Total} \\
		\cmidrule(r){2-5} \cmidrule(l){6-8}
		\textbf{Model} & 
		Obj.\textsuperscript{1} & 
		Img.\textsuperscript{2} & 
		Utility\textsuperscript{3} & 
		Avg. &
		Complex\textsuperscript{4} & 
		Spatial\textsuperscript{5} & 
		Avg. &
		\textbf{Avg.} \\
		\midrule
		Flowedit~\cite{kulikov2025flowedit} & 2.55 & 2.63 & 2.53 & 2.57 & 2.64 & 2.68 & 2.66 & 2.60 \\
		IC-Edit~\cite{zhang2025incontext}   & 2.64 & 3.18 & 2.35 & 2.72 & 2.98 & 2.61 & 2.79 & 2.75 \\
		OmniGen2~\cite{wu2025omnigen}  & 2.95 & 3.33 & 2.99 & 3.09 & 3.04 & 2.81 & 2.92 & 3.02 \\
		Ovis-U1~\cite{wang2025ovisu}  & 2.55 & 3.19 & 2.70 & 2.81 & 2.83 & 2.52 & 2.67 & 2.76 \\
		X2Edit~\cite{ma2025x2edit} & 2.91 & 3.32 & 2.14 & 2.79 & 2.62 & 2.31 & 2.47 & 2.66 \\
		Qwen-Edit\textsuperscript{6}~\cite{wu2025qwenimagetechnicalreport}  & \textbf{3.78} & \textbf{3.97} & 3.66 & 3.80 & \textbf{3.75} & \textbf{3.45} & \textbf{3.60} & \textbf{3.72} \\
		DreamOmni2~\cite{xia2025dreamomni2}  & 3.08 & 3.34 & 3.30 & 3.24 & 3.15 & 2.99 & 3.07 & 3.17 \\
		Uniworld-V2~\cite{li2025uniworldv2}  & 3.73 & 3.96 & \textbf{3.84} & \textbf{3.84} & 3.70 & 3.29 & 3.50 & 3.70 \\
		Step1X-Edit-v1.2~\cite{liu2025stepxedit}  & 3.68 & 3.78 & 3.61 & 3.69 & 3.67 & 3.32 & 3.50 & 3.61 \\
		Flux.2-dev~\cite{flux-2-2025}  & 3.49 & 3.73 & 3.56 & 3.59 & 3.67 & 3.19 & 3.43 & 3.53 \\
		\bottomrule
	\end{tabular*}
	\footnotesize
	Full names for each evaluation dimension:
	\textsuperscript{1}Object-Level Operations, 
	\textsuperscript{2}Image-Level Operations, 
	\textsuperscript{3}Utility Tasks,
	\textsuperscript{4}Complex Instruction and Reasoning, 
	\textsuperscript{5}Spatial Understanding and Reasoning,
	\textsuperscript{6}Qwen-Image-Edit-2509.
\end{table*}

\subsection{User Study Protocol}
There are 12 visual experts, each of whom has practical experience in image editing and possesses a professional background in the field. They have participated in subjective evaluation experiments for text-to-image and image-to-image models before, gaining a thorough understanding of aspects such as prompt alignment, image quality, distortion in the spatial domain. For this work, they all underwent thorough training prior to the evaluation.

In the evaluation process, each expert was assigned a set of images generated by different models for the same editing instruction. 
All the generated images were presented in a randomized order to eliminate any potential bias. Therefore, the experts were not aware of which model produced which image during the evaluation.
For 21 distinct evaluation tasks, each task is assigned 3 specific evaluation dimensions to assess the performance of the models. And each dimension has 1-5 rating levels, where 1 indicates poor performance and 5 indicates excellent performance. Each level is defined by specific criteria tailored to the task at hand.

For example, for the ``Addition" task, the evaluation dimensions include ``Prompt Compliance", ``Visual Naturalness", and ``Physical and Detail Coherence"; for the ``Viewpoint Transformation" task, the evaluation dimensions include ``Viewpoint Transformation Accuracy", ``Content Consistency and Coherence", and ``Geometric Fidelity and Visual Quality". Specifically, for the first dimension ``Prompt Compliance" in ``Addition" task, the levels are defined as follows:
\begin{itemize}
	\item Level 1: Nothing added or the added content is corrupt.
	\item Level 2: Added object is a wrong class or unrelated to the prompt.
	\item Level 3: Correct class, but key attributes (position, colour, size, count, etc.) are wrong.
	\item Level 4: Main attributes correct; only minor details off or 1-2 small features missing.
	\item Level 5: Every stated attribute correct and scene logic reasonable; only microscopic flaws.
\end{itemize}
Similar detailed definitions were provided for each evaluation dimension across all tasks and will be open-sourced in the future.

Each evaluation dimension was evaluated by 4 different experts to ensure the reliability and robustness of the evaluation results. All dimensions were scored independently without considering other dimensions to avoid potential interference. The score of each dimension for an image was obtained by averaging the scores given by the 4 experts. The final score for each generated image was obtained by averaging the scores of all evaluation dimensions.

\subsection{Result Analysis}
This section presents an in-depth analysis of the empirical results, aiming to reveal the strengths and limitations of current technical paradigms and to provide diagnostic insights for future research.

For evaluating all models accurately, we use all dataset samples in CDD-IIE Bench to test each model, \textit{i.e.,} 1353 image-instruction pairs in total. And each result is evaluated by 3 dimensions with 4 experts per dimension, resulting in 12 scores per image-instruction pair, leading to a total of 162,360 expert ratings across all models and tasks. All ratings are averaged to obtain the final scores for each model on each task. The detailed results are shown in Tab.~\ref{tab:CDD-Bench_results_basic}, Tab.~\ref{tab:CDD-Bench_results_advanced}, and Tab.~\ref{tab:CDD-Bench_results_summary}.

\label{sec:result}
\subsubsection{Experimental Results and Analysis}

\textbf{Overall Performance Analysis.}
The empirical results clearly delineate a performance hierarchy among the evaluated models. As shown in Table~\ref{tab:CDD-Bench_results_summary}, Qwen-Image-Edit-2509 and Uniworld-V2 demonstrated outstanding performance, achieving the highest overall scores of 3.72 and 3.70, respectively, thereby establishing themselves as the top-tier performers. This indicates their superior comprehensive capability in executing diverse and complex human instructions for image editing.

Following them, Step1x-Edit-v1.2 and Flux.2-dev form a second tier, also showcasing robust and powerful performance.
In contrast, earlier models such as Flowedit achieved relatively lower scores in this comprehensive evaluation, a fact that reflects the rapid technological iteration in this field in recent years.
Similarly, several research-focused models, including IC-EDit and X2Edit, also performed poorly. These models exhibited notable limitations during our evaluation, such as failing to maintain consistent image dimensions between edits, indicating that these models are still nascent and not yet ready for practical applications.

\textbf{Basic Atomic versus Advanced Compositional Capabilities.}
Our analysis reveals a prevalent performance gap: models consistently score higher on basic capabilities (e.g., OmniGen2 at 3.09) compared to advanced capabilities (OmniGen2 at 2.92). This disparity highlights that while current models are proficient at atomic-level instructions, their performance degrades when faced with complex commands requiring logical composition, spatial planning, and deep semantic understanding. This bottleneck represents a obstacle in their evolution from useful tools to intelligent assistants.

\textbf{Analysis by Core Capability Dimension}
We delve further into the five core capability dimensions:
\begin{enumerate}
	\item Object-Level Operations: Qwen-Image-Edit-2509 and Uniworld-V2 excelled in this dimension, demonstrating precise control over core tasks such as adding, removing, replacing, and modifying object attributes.
	\item Image-Level Operations: Once again, Qwen-Image-Edit-2509 and Uniworld-V2 took the lead, showcasing superior abilities, particularly in tone and style transfer.
	\item Utility Tasks: For tasks such as image repair and visual effect removal, Uniworld-V2 and Qwen-Image-Edit-2509 delivered the best performance again.
	\item Complex Instruction and Reasoning: Qwen-Image-Edit-2509 was the top performer in handling parallel and sequential instructions. Step1X-Edit-v1.2 shows the best performance in Implicit Reasoning.
	\item Spatial Understanding and Reasoning: This was the weakest dimension for most models and best reflects their ``intelligence''. Qwen-Image-Edit-2509 performed best, showing its potential in handling difficult tasks. Step1X-Edit-v1.2 performs well in the ``Referring-based Editing'' tasks.
\end{enumerate}

\subsubsection{Insights and Conclusion}
Based on the foregoing analysis, we propose three diagnostic insights to advance the capabilities of next-generation instruction-based image editing models.

\textbf{First, current models exhibit a significant disconnect between high-level semantic editing and low-level utility tasks.}
A salient finding is the general poor performance of models on ``Utility Tasks''. For instance, Qwen-Image-Edit-2509, despite it getting 3.97 in image-level operations, scored only 3.66 on utility tasks, below its performance in other dimensions. This reveals a potential bias in the current research paradigm, where emphasis on advancing high-level semantic editing comes at the expense of low-level tasks that are equally crucial for a wide range of real-world applications. This trend of prioritizing semantics over fidelity may limit the reliability and application scope of these models in real-world scenarios. Future work must accord equal importance to high-level semantic understanding and low-level image processing. An ideal model should seamlessly perform low-level edits in conjunction with complex semantic manipulations, which may necessitate novel model architectures or the introduction of multi-task training paradigms.

\textbf{Second, all models demonstrate a phenomenon where they are strong in atomic editing capabilities but weak in compositional editing capabilities.}
This is rooted in a fundamental lack of reasoning ability. Specifically, ``Spatial Understanding \& Reasoning'' has emerged as a critical bottleneck for all models, with an average score of just 2.92, the lowest among all dimensions. Even the top-performing model, Qwen-Image-Edit-2509, scored lower in this area (3.45) compared to its other capabilities. This indicates that current models lack the ability to precisely interpret and execute commands involving orientation, spatial relations and viewpoint transformations, which critically constrains their upper performance limits. Future research should therefore prioritize enhancing the compositional reasoning abilities of models, with a particular focus on improving the modeling and manipulation of spatial relationships. Integrating the reasoning paradigms of Large Language Models (LLMs) into the domain of instruction-based editing presents a promising avenue for breaking through this bottleneck.

\textbf{Third, most models lack robust generality, excelling in specific domains but failing to achieve high performance across all editing tasks.}
For example, Uniworld-V2 excels in the ``Utility Tasks'' dimension, demonstrating its effectiveness in handling low-level editing. In contrast, Qwen-Image-Edit-2509 achieves state-of-the-art performance on ``Object-Level'' and ``Image-Level Operations'' but falters on ``Utility Tasks'', suggesting its architecture or training data may be biased towards semantic editing tasks. As the goal of artificial general intelligence is to build models capable of handling diverse tasks, future development should aim to address these capability gaps and pursue the kind of robust generality, while still maintaining high performance on specialized tasks. Exploring modular designs that allow a model to dynamically invoke expert modules for specific tasks could be a fruitful direction for future research.

\section{Application} \label{secB6}

\subsection{GUI-Based  Application} 

\begin{figure*}[htbp]
	\centering
	\includegraphics[width=1.0\textwidth]{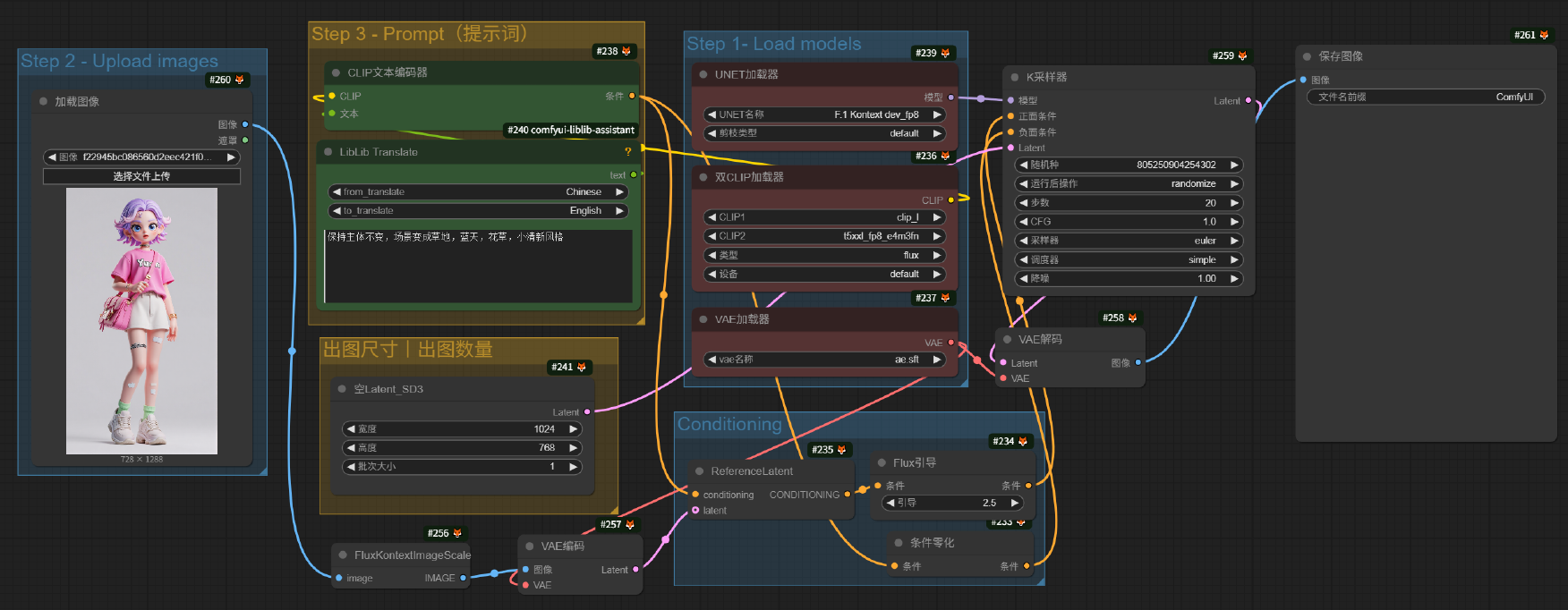}
	\caption{The interface of ComfyUI}
	\label{fig:comfyui}
\end{figure*}

Graphical user interface (GUI)-based applications interact with users through a visual interface. Users can select models, adjust parameters, enter prompts, or upload images within the interface to generate and edit images. These applications typically offer greater flexibility, support for complex workflow configurations, and are well-suited for users who demand high image quality and fine control over generation details.

\subsubsection{ComfyUI} 

ComfyUI is a node-based visual interface built upon the Stable Diffusion framework~\cite{wiki:ComfyUI}, designed to present the complex process of AI-driven image generation in a more intuitive and accessible manner. Unlike traditional image generation tools, ComfyUI employs a unique node-based workflow system that visually decomposes the operations of deep learning models. This approach enhances the transparency, flexibility, and debuggability of the entire AI image synthesis pipeline.

Each node within ComfyUI represents a specific image processing or generation task, such as noise manipulation, model inference, or image decoding. As shown in Fig.~\ref{fig:comfyui}, by simply dragging and connecting these nodes, users can construct fully customized image-generation workflows tailored to their specific needs. This design significantly lowers the technical barrier for users engaging in AI-based artistic creation.

Moreover, ComfyUI supports both the customization and reuse of workflows, enabling efficient batch processing of images. As a result, the tool not only facilitates individualized image synthesis but also streamlines large-scale image generation, making the process more convenient and productive.

\subsubsection{WebUI} 
Stable Diffusion WebUI is a graphical user interface (GUI) designed for advanced users. It provides rich functions and flexibility to meet the complex and advanced image generation requirements. It also simplifies the training and deployment process of the Stable Diffusion model, enabling users to develop and apply deep learning models more efficiently. Through the Stable Diffusion WebUI, users can easily manage and adjust the training and deployment of the model, and analyze and compare the model performance. This provides a powerful tool for practitioners and enthusiasts in the field of deep learning, helping them realize their creativity and ideas more quickly.

\subsection{Interactive Application}

Interactive applications use natural language as the primary mode of interaction. Users simply upload an image and enter a text prompt describing their needs, and the system automatically edits the image and provides the generated result. These applications are easy to operate, beginner-friendly, and require little to no professional background, making them especially suitable for general users.

\begin{figure}[htbp]
	\centering
	\includegraphics[width=\columnwidth]{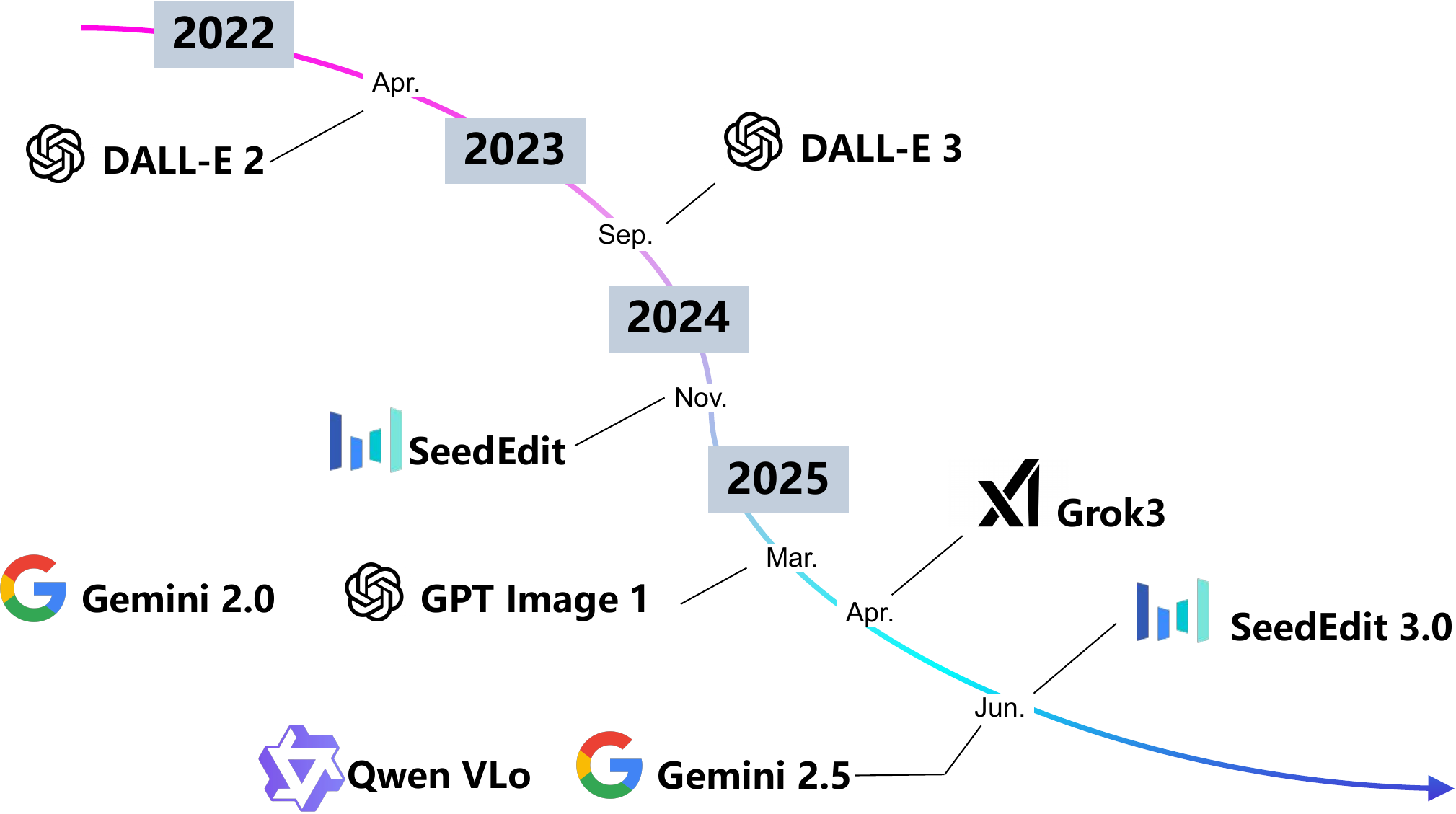}
	\caption{The timeline of mainstream close-sourced instruction-base edit models.}
	\label{fig:timeline}
\end{figure}

In recent years, major model developers have been intensifying their efforts to enrich the capabilities of their multimodal large models. Fig.~\ref{fig:timeline} shows the timeline of the development of mainstream instruction-base edit models. As a result, instruction-based editing, as a crucial component of multimodal generation, has witnessed rapid development. 

\begin{figure}[htbp]
	\centering
	\includegraphics[width=\columnwidth]{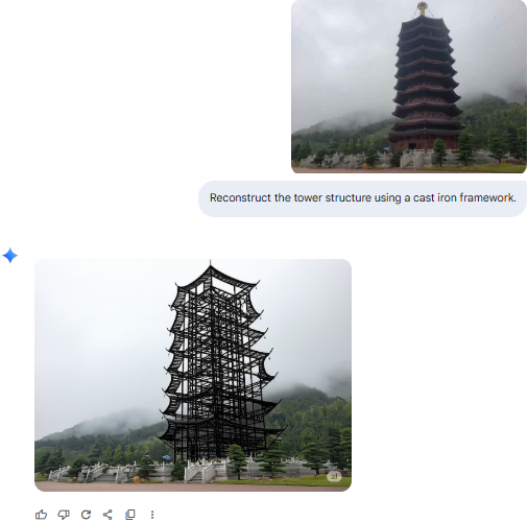}
	\caption{The interface of Gemini}
	\label{fig:gemini}
\end{figure}

\subsubsection{ChatGPT}
ChatGPT is an artificial intelligence application developed by OpenAI, known for its powerful natural language understanding and generation capabilities. In the past, ChatGPT image generation functionality primarily relied on DALL$\cdot$E~\cite{ramesh2022hierarchical, openai2023dalle3}, a standalone image generation model with limited conversational capabilities. 
Subsequently, OpenAI introduced GPT-4o, GPT-Image-1, and GPT-Image-1.5. The latest model can understand the full context of a conversation and generate images that better align with user intent based on the dialogue.

\subsubsection{Gemini}
Gemini is an artificial intelligence model developed by Google DeepMind. Google officially released the Gemini series, a fully multimodal image generator that supports native image generation capabilities, which demonstrates significant advancements in tasks such as web application development, agent-based code generation, and code translation. Especially, the recent updated Nano Banana Pro model exhibits remarkable performance in image generation and editing tasks (Fig. \ref{fig:gemini}).

\subsubsection{Grok}
Grok is an artificial intelligence application developed by xAI. xAI continue release Grok series, a multimodal model integrated with image editing capabilities. Grok models are powered by xAI's latest Aurora image generation model, allowing users to upload images and input natural language prompts to instantly generate edited images.

\subsubsection{Doubao}
Doubao is an artificial intelligence application developed by ByteDance, specifically optimized for Chinese-speaking users to enhance its understanding and execution of Chinese-language instructions~\cite{shi2024seededit}. 
This year, ByteDance released a series of Seedream models, which support many tasks in Jimeng and Toutiao products.

\subsubsection{Qwen-Max}
Qwen is an artificial intelligence application developed by the Alibaba Cloud Tongyi Qianwen team. This year, they released Qwen-Max, a unified multimodal understanding and generation model that supports image editing.

\subsection{Comments}
The IIE task, with its broad and practical applications, has become a fiercely contested area for major commercial applications. Recently, high-impact models and applications have emerged almost every month, underscoring the tremendous potential of this field.

\section{Future Work} \label{secB7} 
Instruction-based Image Editing (IIE) demonstrates substantial application potential across diverse domains, with numerous existing models capable of performing these tasks. Future research directions may focus on, but are not limited to, the following key aspects:

\textbf{Enhancing Controllability and Interpretability.}
Future image editing models should prioritize fine-grained controllability, enabling users to manipulate images at both global (e.g., scene layout) and local (e.g., object attributes) levels. For Stable Diffusion, this could involve improving conditional mechanisms, such as integrating spatial-aware attention layers or leveraging semantic segmentation maps for precise region-based editing. For autoregressive (AR) models, research could explore hierarchical latent representations to decouple content and style, allowing independent control over different image aspects. Additionally, integrating explainable AI (XAI) techniques could make model decisions more interpretable, helping users understand how edits are generated.

\textbf{Combining Strengths of Diffusion and Autoregressive Models.}
A promising direction is the fusion of diffusion and AR models to leverage their complementary strengths. Diffusion models excel at high-fidelity generation and global coherence, while AR models offer explicit likelihood modeling and sequential refinement capabilities. Future work could investigate:
\begin{itemize}
	\item Hybrid architectures, where diffusion models generate low-resolution layouts, and AR models refine details autoregressively.
	\item Latent-space AR diffusion, where an AR model operates in the compressed latent space of a diffusion model, balancing efficiency and quality.
	\item Dynamic sampling strategies, using AR models to guide diffusion sampling paths for faster convergence or improved detail preservation.
\end{itemize}

\textbf{Real-Time and Efficient Editing.}
Current limitations in inference speed (especially for AR models) and computational cost (for high-resolution diffusion models) must be addressed for practical applications. Potential solutions include:
\begin{itemize}
	\item Distillation techniques to compress large diffusion or AR models into lightweight variants without sacrificing quality.
	\item Sparse or patch-based autoregressive modeling, reducing sequential dependencies by processing image regions in parallel.
	\item Hardware-aware optimizations, such as leveraging neural acceleration or quantization for deployment on edge devices.
\end{itemize}

\textbf{Multimodal and Interactive Editing.}
Future image editors should support multimodal inputs (text, sketches, audio, or 3D cues) and real-time user interaction. Key advancements may involve:
\begin{itemize}
	\item Unified cross-modal encoders (e.g., CLIP-like models) to align diverse input modalities with latent representations.
	\item Interactive refinement loops, where users iteratively guide edits via natural language feedback or scribble-based corrections.
	\item Memory-augmented models that retain user preferences or editing histories for personalized adaptation.
\end{itemize}

\textbf{Ethical and Safe Image Editing.}
As image manipulation becomes more accessible, ensuring ethical use and detection of synthetic media is critical. Future work should focus on:
\begin{itemize}
	\item Watermarking or provenance tracking for AI-generated edits, possibly via cryptographic signatures or latent-space tagging.
	\item Bias mitigation in training data and models to prevent harmful stereotypes or misuse.
	\item Adversarial robustness to resist malicious edits (e.g., deepfake generation).
\end{itemize}

The next generation of image editing models will likely unify generation, understanding, and control through hybrid architectures, efficient inference, and multimodal interaction. By combining the strengths of diffusion and AR models, while addressing their limitations, researchers can enable powerful, real-time, and user-centric editing tools. Simultaneously, ethical considerations must remain central to ensure responsible deployment.

\section{Conclusion} \label{secB8} 
This paper presents a comprehensive investigation of Instruction-based Image Editing (IIE) through both theoretical analysis and empirical benchmarking. Our study systematically examines six critical dimensions of IIE research: (1) task definition and hierarchical taxonomy, (2) data construction methodologies for diverse editing tasks, (3) evolutionary trajectory of fundamental architectures (GANs, Diffusion Models, and AR Models), (4) progression of evaluation metrics and our proposed CDD-IIE benchmark, (5) comparative analysis between open-source models, and (6) emerging research directions. The work makes three primary contributions to the field: First, we establish a unified taxonomy that formally categorizes IIE tasks and their relationships. Second, we provide the most comprehensive analysis to date, encompassing data construction paradigms, architectural evolution, and evaluation methodologies. Third, we introduce a novel multi-dimensional evaluation framework comprising 5 principal categories and 21 fine-grained metrics, enabling standardized assessment across the IIE landscape.

\section*{Abbreviations} \label{secB9}

\begin{flushleft}
	\begin{tabular}{ll}
		IIE & Instruction-based Image Editing \\
		\multirow{2}{*}{CDD-IIE Bench}    & Comprehensive, in-Depth, and Diagnostic		 \\
		 & benchmark for IIE task \\
		LLMs    & Large Language Models  \\
		VLMs    & Vision-Language Models \\
		GANs    & Generative Adversarial Networks \\
		CLIP    & Contrastive Language-Image Pre-training \\
		DMs    & Diffusion Models \\
		AR    & AutoRegressive \\
		LDM    & Latent Diffusion Model \\
		VAE    & Variational AutoEncoder \\
		LPIPS    & Learned Perceptual Image Patch Similarity \\
		MLE    & Maximum Likelihood Estimation \\
		NLL    & Negative Log-Likelihood \\
		GMM    & Gaussian Mixture Model \\
		VGG    & Visual Geometry Group \\
		NAD    & Normalized Attention Difference \\
		OCR    & Optical Character Recognition \\
		PnP    & Plug-and-Play \\
		DiTs    & Diffusion Transformers \\
		AdaIN    & Adaptive Instance Normalization \\
		CFG    & Classifier-Free Guidance \\
		SD    & Stable Diffusion \\
		MLLMs    & Multimodal Large Language Models \\
		\multirow{2}{*}{DINO}    & DETR with Improved deNoising  \\
		& anchOr boxes \\
		ID    & IDentity \\
		IP    & Intellectual Property \\
		DDIM    & Denoising Diffusion Implicit Models \\
		DDPM    & Denoising Diffusion Probabilistic Models \\
		GT    & Ground Truth \\
		SSIM    & Structural Similarity Index Measure \\
		ViT    & Vision Transformer \\
		VQA    & Visual Question Answering \\
		PQ    & Perceptual Quality \\
		SC    & Semantic Consistency \\
		VIEScore & Visual Instruction-guided Explainable Score \\
	\end{tabular}
\end{flushleft}

\section*{Acknowledgments}
The authors express their gratitude to TeleAI for providing critical computational infrastructure and personnel support throughout this research initiative.

\section*{Authors' Contributions}
All authors made equal contributions to the conceptualization, design, and writing of this work. Each author took part in reviewing the literature, analyzing data, and drafting the manuscript. All authors have reviewed and approved the final version of the manuscript for publication.

\section*{Funding}
Not applicable.

\section*{Data Availability}
The datasets used in this study are publicly available at the referenced links. Additionally, the data accompanying the dataset introduced in this work will be publicly released in the near future.

\section*{Code Availability}
The relevant code for the dataset proposed in this paper will be made public in the near future.

\section*{Declarations}
\subsection*{Competing Interests}
The authors declare no known competing financial interests or personal relationships that could have influenced the work reported in this manuscript.

\bibliography{sn-bibliography}

@Article{li2024llava,
  author  = {Li, Bo and Zhang, Yuanhan and Guo, Dong and Zhang, Renrui and Li, Feng and Zhang, Hao and Zhang, Kaichen and Zhang, Peiyuan and Li, Yanwei and Liu, Ziwei and others},
  journal = {Trans. Mach. Learn. Res.},
  title   = {Llava-onevision: Easy visual task transfer},
  year    = {2025},
  pages   = {1--1},
}

@InProceedings{liu2023visual,
  author    = {Liu, Haotian and Li, Chunyuan and Wu, Qingyang and Lee, Yong Jae},
  booktitle = {Advances in Neural Information Processing Systems},
  title     = {Visual instruction tuning},
  year      = {2023},
  address   = {New Orleans, USA},
  pages     = {34892--34916},
  publisher = {Curran Associates},
  volume    = {36},
}

@InProceedings{Sushko2025REALEDITRE,
  author    = {Peter Sushko and Ayana Bharadwaj and Zhi Yang Lim and Vasily Ilin and Ben Caffee and Dongping Chen and Mohammadreza Salehi and Cheng-Yu Hsieh and Ranjay Krishna},
  booktitle = {Proceedings of the IEEE/CVF Conference on Computer Vision and Pattern Recognition},
  title     = {REALEDIT: Reddit Edits As a Large-scale Empirical Dataset for Image Transformations},
  year      = {2025},
  address   = {Nashville, USA},
  pages     = {13403-13413},
  publisher = {IEEE},
}

@InProceedings{Ku2023VIEScoreTE,
  author    = {Max W.F. Ku and Dongfu Jiang and Cong Wei and Xiang Yue and Wenhu Chen},
  booktitle = {Proceedings of the Annual Meeting of the Association for Computational Linguistics},
  title     = {VIEScore: Towards Explainable Metrics for Conditional Image Synthesis Evaluation},
  year      = {2023},
  address   = {Toronto, Canada},
  pages     = {12268-12290},
  publisher = {ACL},
}

@Article{Chang2025ByteMorphBI,
  author  = {Di Chang and Mingdeng Cao and Yichun Shi and Bo Liu and Shengqu Cai and Shijie Zhou and Weilin Huang and Gordon Wetzstein and Mohammad Soleymani and Peng Wang},
  journal = {arXiv preprint arXiv:2506.03107},
  title   = {ByteMorph: Benchmarking Instruction-Guided Image Editing with Non-Rigid Motions},
  year    = {2025},
}

@InProceedings{Wu2025KRISBenchBN,
  author    = {Yongliang Wu and Zonghui Li and Xinting Hu and Xinyu Ye and Xianfang Zeng and Gang Yu and Wenbo Zhu and Bernt Schiele and Mingzhuo Yang and Xu Yang},
  booktitle = {The Annual Conference on Neural Information Processing Systems Datasets and Benchmarks Track},
  title     = {KRIS-Bench: Benchmarking Next-Level Intelligent Image Editing Models},
  year      = {2025},
  pages = {1--20},
  address   = {San Diego, USA},
  publisher = {Curran Associates},
}

@InProceedings{Krojer2024LearningAA,
  author    = {Benno Krojer and Dheeraj Vattikonda and Luis Lara and Varun Jampani and Eva Portelance and Christopher Pal and Siva Reddy},
  booktitle = {Advances in Neural Information Processing Systems},
  title     = {Learning Action and Reasoning-Centric Image Editing from Videos and Simulations},
  year      = {2024},
  address   = {Vancouver, Canada},
  pages     = {38035-38078},
  publisher = {Curran Associates},
}

@InProceedings{Ma2024I2EBenchAC,
  author    = {Yiwei Ma and Jiayi Ji and Ke Ye and Weihuang Lin and Zhibin Wang and Yonghan Zheng and Qiang Zhou and Xiaoshuai Sun and Rongrong Ji},
  booktitle = {Advances in Neural Information Processing Systems},
  title     = {I2EBench: A Comprehensive Benchmark for Instruction-based Image Editing},
  year      = {2024},
  address   = {Vancouver, Canada},
  pages     = {41494-41516},
  publisher = {Curran Associates},
}

@Article{Yang2025ComplexEditCI,
  author  = {Siwei Yang and Mude Hui and Bingchen Zhao and Yuyin Zhou and Nataniel Ruiz and Cihang Xie},
  journal = {arXiv preprint arXiv:2504.13143},
  title   = {Complex-Edit: CoT-Like Instruction Generation for Complexity-Controllable Image Editing Benchmark},
  year    = {2025},
}

@InProceedings{Wang2025ComplexBenchEditBC,
  author    = {Chenglin Wang and Yucheng Zhou and Qianning Wang and Zhe Wang and Kai Zhang},
  booktitle = {Proceedings of the ACM International Conference on Multimedia},
  title     = {ComplexBench-Edit: Benchmarking Complex Instruction-Driven Image Editing via Compositional Dependencies},
  year      = {2025},
  address   = {Dublin, Ireland},
  pages     = {13391-13397},
  publisher = {ACM},
}

@InProceedings{Pathiraja2025RefEditAB,
  author    = {Bimsara Pathiraja and Maitreya Patel and Shivam Singh and Yezhou Yang and Chitta Baral},
  booktitle = {Proceedings of the IEEE/CVF International Conference on Computer Vision},
  title     = {RefEdit: A Benchmark and Method for Improving Instruction-based Image Editing Model on Referring Expressions},
  year      = {2025},
  address   = {Marrakech, Morocco},
  pages     = {15646-15656},
  publisher = {IEEE},
}

@InProceedings{Zhao2025EnvisioningBT,
  author    = {Xiangyu Zhao and Peiyuan Zhang and Kexian Tang and Hao Li and Zicheng Zhang and Guangtao Zhai and Junchi Yan and Hua Yang and Xue Yang and Haodong Duan},
  booktitle = {The Annual Conference on Neural Information Processing Systems Datasets and Benchmarks Track},
  title     = {Envisioning Beyond the Pixels: Benchmarking Reasoning-Informed Visual Editing},
  year      = {2025},
  pages = {1--20},
  address   = {San Diego, USA},
  publisher = {Curran Associates},
}

@InProceedings{Pan2025ICEBenchAU,
  author    = {Yulin Pan and Xiangteng He and Chaojie Mao and Zhen Han and Zeyinzi Jiang and Jingfeng Zhang and Yu Liu},
  booktitle = {Proceedings of the IEEE/CVF International Conference on Computer Vision},
  title     = {ICE-Bench: A Unified and Comprehensive Benchmark for Image Creating and Editing},
  year      = {2025},
  address   = {Marrakech, Morocco},
  pages     = {16586-16596},
  publisher = {IEEE},
}

@Article{Jia2025CompBenchBC,
  author  = {Bohan Jia and Wenxuan Huang and Yuntian Tang and Junbo Qiao and Jincheng Liao and Shaosheng Cao and Fei Zhao and Zhaopeng Feng and Zhouhong Gu and Zhenfei Yin and Lei Bai and Wanli Ouyang and Lin Chen and Zihan Wang and Yuan Xie and Shaohui Lin},
  journal = {arXiv preprint arXiv:2505.12200},
  title   = {CompBench: Benchmarking Complex Instruction-guided Image Editing},
  year    = {2025},
}

@Article{Basu2023EditValBD,
  author  = {Samyadeep Basu and Mehrdad Saberi and Shweta Bhardwaj and Atoosa Malemir Chegini and Daniela Massiceti and Maziar Sanjabi and Shell Xu Hu and Soheil Feizi},
  journal = {arXiv preprint arXiv:2310.02426},
  title   = {EditVal: Benchmarking Diffusion Based Text-Guided Image Editing Methods},
  year    = {2023},
}

@InProceedings{Wang2022ImagenEA,
  author    = {Su Wang and Chitwan Saharia and Ceslee Montgomery and Jordi Pont-Tuset and Shai Noy and Stefano Pellegrini and Yasumasa Onoe and Sarah Laszlo and David J. Fleet and Radu Soricut and Jason Baldridge and Mohammad Norouzi and Peter Anderson and William Chan},
  booktitle = {Proceedings of the IEEE/CVF Conference on Computer Vision and Pattern Recognition},
  title     = {Imagen Editor and EditBench: Advancing and Evaluating Text-Guided Image Inpainting},
  year      = {2022},
  address   = {New Orleans, USA},
  pages     = {18359-18369},
  publisher = {IEEE},
}

@InProceedings{wei2022hairclip,
  author    = {Wei, Tianyi and Chen, Dongdong and Zhou, Wenbo and Liao, Jing and Tan, Zhentao and Yuan, Lu and Zhang, Weiming and Yu, Nenghai},
  booktitle = {Proceedings of the IEEE/CVF Conference on Computer Vision and Pattern Recognition},
  title     = {Hairclip: Design your hair by text and reference image},
  year      = {2022},
  address   = {New Orleans, USA},
  pages     = {18072--18081},
  publisher = {IEEE},
  type      = {Conference Proceedings},
}

@InProceedings{patashnik2021styleclip,
  author    = {Patashnik, Or and Wu, Zongze and Shechtman, Eli and Cohen-Or, Daniel and Lischinski, Dani},
  booktitle = {Proceedings of the IEEE/CVF International Conference on Computer Vision},
  title     = {Styleclip: Text-driven manipulation of stylegan imagery},
  year      = {2021},
  address   = {Virtual},
  pages     = {2085--2094},
  publisher = {IEEE},
  type      = {Conference Proceedings},
}

@InProceedings{tero2019stylegan,
  author    = {Tero Karras and Samuli Laine and Timo Aila},
  booktitle = {Proceedings of the IEEE/CVF Conference on Computer Vision and Pattern Recognition},
  title     = {A Style-Based Generator Architecture for Generative Adversarial Networks},
  year      = {2019},
  address   = {Long Beach, USA},
  pages     = {4401--4410},
  publisher = {IEEE},
}

@InProceedings{tero2020stylegan2,
  author    = {Tero Karras and Samuli Laine and Miika Aittala and Janne Hellsten and Jaakko Lehtinen and Timo Aila},
  booktitle = {Proceedings of the IEEE/CVF Conference on Computer Vision and Pattern Recognition},
  title     = {Analyzing and Improving the Image Quality of StyleGAN},
  year      = {2020},
  address   = {Seattle, USA},
  pages     = {8107--8116},
  publisher = {IEEE},
}

@InProceedings{xia2021tedigan,
  author    = {Xia, Weihao and Yang, Yujiu and Xue, Jing-Hao and Wu, Baoyuan},
  booktitle = {Proceedings of the IEEE/CVF Conference on Computer Vision and Pattern Recognition},
  title     = {Tedigan: Text-guided diverse face image generation and manipulation},
  year      = {2021},
  address   = {Virtual},
  pages     = {2256--2265},
  publisher = {IEEE},
  type      = {Conference Proceedings},
}

@InProceedings{jiang2021talktoedit,
  author    = {Jiang, Yuming and Huang, Ziqi and Pan, Xingang and Loy, Chen Change and Liu, Ziwei},
  booktitle = {Proceedings of the IEEE/CVF International Conference on Computer Vision},
  title     = {Talk-to-edit: Fine-grained facial editing via dialog},
  year      = {2021},
  address   = {Virtual},
  pages     = {13799--13808},
  publisher = {IEEE},
  type      = {Conference Proceedings},
}

@InProceedings{gihyun2022clipstyler,
  author    = {Gihyun Kwon and Jong Chul Ye},
  booktitle = {Proceedings of the IEEE/CVF Conference on Computer Vision and Pattern Recognition},
  title     = {CLIPstyler: Image Style Transfer with a Single Text Condition},
  year      = {2022},
  address   = {New Orleans, USA},
  pages     = {18041--18050},
  publisher = {IEEE},
  type      = {Conference Proceedings},
}

@Article{baykal2023clipguided,
  author  = {Baykal, Ahmet Canberk and Anees, Abdul Basit and Ceylan, Duygu and Erdem, Erkut and Erdem, Aykut and Yuret, Deniz},
  journal = {ACM Trans. Graph.},
  title   = {CLIP-guided StyleGAN inversion for text-driven real image editing},
  year    = {2023},
  issn    = {0730-0301},
  number  = {5},
  pages   = {1--18},
  volume  = {42},
  type    = {Journal Article},
}

@Article{rinon2022stylegan-nada,
  author  = {Rinon Gal and Or Patashnik and Haggai Maron and Amit H. Bermano and Gal Chechik and Daniel Cohen{-}Or},
  journal = {ACM Trans. Graph.},
  title   = {StyleGAN-NADA: CLIP-guided domain adaptation of image generators},
  year    = {2022},
  number  = {4},
  pages   = {1--13},
  volume  = {41},
  type    = {Journal Article},
}

@InProceedings{lv2023controlnet,
  author    = {Lvmin Zhang and Anyi Rao and Maneesh Agrawala},
  booktitle = {Proceedings of the IEEE/CVF International Conference on Computer Vision},
  title     = {Adding Conditional Control to Text-to-Image Diffusion Models},
  year      = {2023},
  address   = {Paris, France},
  pages     = {3836--3847},
  publisher = {IEEE},
  type      = {Conference Proceedings},
}

@InProceedings{brooks2023instructpixpix,
  author    = {Brooks, Tim and Holynski, Aleksander and Efros, Alexei A},
  booktitle = {Proceedings of the IEEE/CVF Conference on Computer Vision and Pattern Recognition},
  title     = {Instructpix2pix: Learning to follow image editing instructions},
  year      = {2023},
  address   = {Vancouver, Canada},
  pages     = {18392--18402},
  publisher = {IEEE},
  type      = {Conference Proceedings},
}

@Article{ma2025x2edit,
  author  = {Ma, Jian and Zhu, Xujie and Pan, Zihao and Peng, Qirong and Guo, Xu and Chen, Chen and Lu, Haonan},
  journal = {arXiv preprint arXiv:2508.07607},
  title   = {X2edit: Revisiting arbitrary-instruction image editing through self-constructed data and task-aware representation learning},
  year    = {2025},
}

@InProceedings{zhang2023magicbrush,
  author    = {Zhang, Kai and Mo, Lingbo and Chen, Wenhu and Sun, Huan and Su, Yu},
  booktitle = {Advances in Neural Information Processing Systems},
  title     = {Magicbrush: A manually annotated dataset for instruction-guided image editing},
  year      = {2023},
  address   = {New Orleans, USA},
  pages     = {31428--31449},
  publisher = {Curran Associates},
}

@InProceedings{huang2024smartedit,
  author    = {Yuzhou Huang and Liangbin Xie and Xintao Wang and Ziyang Yuan and Xiaodong Cun and Yixiao Ge and Jiantao Zhou and Chao Dong and Rui Huang and Ruimao Zhang and Ying Shan},
  booktitle = {Proceedings of the IEEE/CVF Conference on Computer Vision and Pattern Recognition},
  title     = {SmartEdit: Exploring Complex Instruction-Based Image Editing with Multimodal Large Language Models},
  year      = {2024},
  address   = {Seattle, USA},
  pages     = {8362--8371},
  publisher = {IEEE},
  type      = {Conference Proceedings},
}

@InProceedings{yang2024diffusionmaster,
  author    = {Ling Yang and Zhaochen Yu and Chenlin Meng and Minkai Xu and Stefano Ermon and Bin Cui},
  booktitle = {Proceedings of the International Conference on Learning Representations},
  title     = {Mastering Text-to-Image Diffusion: Recaptioning, Planning, and Generating with Multimodal LLMs},
  year      = {2024},
  pages = {1--20},
  address   = {Vienna, Austria},
  publisher = {OpenReview.net},
}

@InProceedings{ju2024brushnet,
  author    = {Ju, Xuan and Liu, Xian and Wang, Xintao and Bian, Yuxuan and Shan, Ying and Xu, Qiang},
  booktitle = {European Conference on Computer Vision},
  title     = {Brushnet: A plug-and-play image inpainting model with decomposed dual-branch diffusion},
  year      = {2024},
  address   = {Milan, Italy},
  pages     = {150--168},
  publisher = {Springer},
}

@InProceedings{yu2024anyedit,
  author    = {Yu, Qifan and Chow, Wei and Yue, Zhongqi and Pan, Kaihang and Wu, Yang and Wan, Xiaoyang and Li, Juncheng and Tang, Siliang and Zhang, Hanwang and Zhuang, Yueting},
  booktitle = {Proceedings of the IEEE/CVF Conference on Computer Vision and Pattern Recognition},
  title     = {AnyEdit: Mastering Unified High-Quality Image Editing for Any Idea},
  year      = {2024},
  address   = {Seattle, USA},
  pages     = {26125-26135},
  publisher = {IEEE},
}

@InProceedings{zhao2024ultraedit,
  author    = {Haozhe Zhao and Xiaojian (Shawn) Ma and Liang Chen and Shuzheng Si and Rujie Wu and Kaikai An and Peiyu Yu and Minjia Zhang and Qing Li and Baobao Chang},
  title     = {UltraEdit: Instruction-based Fine-Grained Image Editing at Scale},
  year      = {2024},
  address   = {Vancouver, Canada},
  pages     = {3058-3093},
  publisher = {Curran Associates},
  booktitle   = {Advances in Neural Information Processing Systems},
}

@InProceedings{geng2024instructdiffusion,
  author    = {Geng, Zigang and Yang, Binxin and Hang, Tiankai and Li, Chen and Gu, Shuyang and Zhang, Ting and Bao, Jianmin and Zhang, Zheng and Li, Houqiang and Hu, Han},
  booktitle = {Proceedings of the IEEE/CVF Conference on Computer Vision and Pattern Recognition},
  title     = {Instructdiffusion: A generalist modeling interface for vision tasks},
  year      = {2024},
  address   = {Seattle, USA},
  pages     = {12709--12720},
  publisher = {IEEE},
  type      = {Conference Proceedings},
}

@Article{jin2024reasonpixpix,
  author  = {Jin, Ying and Ling, Pengyang and Dong, Xiaoyi and Zhang, Pan and Wang, Jiaqi and Lin, Dahua},
  journal = {arXiv preprint arXiv:2405.11190},
  title   = {Reasonpix2pix: instruction reasoning dataset for advanced image editing},
  year    = {2024},
  type    = {Journal Article},
}

@InProceedings{artur2024grounded-instructpix2pix,
  author    = {Artur Shagidanov and Hayk Poghosyan and Xinyu Gong and Zhangyang Wang and Shant Navasardyan and Humphrey Shi},
  booktitle = {IEEE International Conference on Acoustics, Speech and Signal Processing},
  title     = {Grounded-Instruct-Pix2Pix: Improving Instruction Based Image Editing with Automatic Target Grounding},
  year      = {2024},
  address   = {Seoul, Korea},
  pages     = {6585--6589},
  publisher = {IEEE},
  type      = {Conference Proceedings},
}

@InProceedings{kulikov2025flowedit,
  author    = {Kulikov, Vladimir and Kleiner, Matan and Huberman-Spiegelglas, Inbar and Michaeli, Tomer},
  booktitle = {Proceedings of the IEEE/CVF International Conference on Computer Vision},
  title     = {Flowedit: Inversion-free text-based editing using pre-trained flow models},
  year      = {2025},
  address   = {Marrakech, Morocco},
  pages     = {19721--19730},
  publisher = {IEEE},
}

@Article{wu2025qwenimagetechnicalreport,
  author        = {Chenfei Wu and Jiahao Li and Jingren Zhou and Junyang Lin and Kaiyuan Gao and Kun Yan and Sheng-ming Yin and Shuai Bai and Xiao Xu and Yilei Chen and Yuxiang Chen and Zecheng Tang and Zekai Zhang and Zhengyi Wang and An Yang and Bowen Yu and Chen Cheng and Dayiheng Liu and Deqing Li and Hang Zhang and Hao Meng and Hu Wei and Jingyuan Ni and Kai Chen and Kuan Cao and Liang Peng and Lin Qu and Minggang Wu and Peng Wang and Shuting Yu and Tingkun Wen and Wensen Feng and Xiaoxiao Xu and Yi Wang and Yichang Zhang and Yongqiang Zhu and Yujia Wu and Yuxuan Cai and Zenan Liu},
  journal       = {arXiv preprint arXiv:2508.02324},
  title         = {Qwen-Image Technical Report},
  year          = {2025},
}

@InProceedings{li2025superedit,
  author    = {Li, Ming and Gu, Xin and Chen, Fan and Xing, Xiaoying and Wen, Longyin and Chen, Chen and Zhu, Sijie},
  booktitle = {Proceedings of the IEEE/CVF International Conference on Computer Vision},
  title     = {Superedit: Rectifying and facilitating supervision for instruction-based image editing},
  year      = {2025},
  address   = {Marrakech, Morocco},
  pages     = {19206-19215},
  publisher = {IEEE},
}

@Article{liu2025stepxedit,
  author  = {Liu, Shiyu and Han, Yucheng and Xing, Peng and Yin, Fukun and Wang, Rui and Cheng, Wei and Liao, Jiaqi and Wang, Yingming and Fu, Honghao and Han, Chunrui},
  journal = {arXiv preprint arXiv:2504.17761},
  title   = {Step1x-edit: A practical framework for general image editing},
  year    = {2025},
  type    = {Journal Article},
}

@InProceedings{ye2025imgedit,
  author    = {Ye, Yang and He, Xianyi and Li, Zongjian and Lin, Bin and Yuan, Shenghai and Yan, Zhiyuan and Hou, Bohan and Yuan, Li},
  booktitle = {The Annual Conference on Neural Information Processing Systems Datasets and Benchmarks Track},
  title     = {Imgedit: A unified image editing dataset and benchmark},
  year      = {2025},
  pages = {1--20},
  address   = {San Diego, USA},
  publisher = {Curran Associates},
}

@Article{he2024freeedit,
  author  = {He, Runze and Ma, Kai and Huang, Linjiang and Huang, Shaofei and Gao, Jialin and Wei, Xiaoming and Dai, Jiao and Han, Jizhong and Liu, Si},
  journal = {IEEE Trans. Pattern Anal. Mach. Intell.},
  title   = {Freeedit: Mask-free reference-based image editing with multi-modal instruction},
  year    = {2025},
  number  = {3},
  pages   = {3319 - 3334},
  volume  = {48},
  type    = {Journal Article},
}

@InProceedings{zhang2025incontext,
  author    = {Zhang, Zechuan and Xie, Ji and Lu, Yu and Yang, Zongxin and Yang, Yi},
  booktitle = {Advances in Neural Information Processing Systems},
  title     = {In-context edit: Enabling instructional image editing with in-context generation in large scale diffusion transformer},
  year      = {2025},
  address   = {San Diego, USA},
  publisher = {Curran Associates},
  pages = {1--1},
}

@InProceedings{hu2025ieap,
  author    = {Hu, Yujia and Liu, Songhua and Tan, Zhenxiong and Yang, Xingyi and Wang, Xinchao},
  booktitle = {Advances in Neural Information Processing Systems},
  title     = {Image Editing As Programs with Diffusion Models},
  year      = {2025},
  address   = {San Diego, USA},
  publisher = {Curran Associates},
  pages = {1--1},
}

@InProceedings{kunyu2025dit4edit,
  author    = {Kunyu Feng and Yue Ma and Bingyuan Wang and Chenyang Qi and Haozhe Chen and Qifeng Chen and Zeyu Wang},
  booktitle = {Proceedings of the AAAI Conference on Artificial Intelligence},
  title     = {DiT4Edit: Diffusion Transformer for Image Editing},
  year      = {2025},
  address   = {Philadelphia, USA},
  pages     = {2969--2977},
  publisher = {AAAI},
  type      = {Conference Proceedings},
}

@InProceedings{zhou2025fireedit,
  author    = {Zhou, Jun and Li, Jiahao and Xu, Zunnan and Li, Hanhui and Cheng, Yiji and Hong, Fa-Ting and Lin, Qin and Lu, Qinglin and Liang, Xiaodan},
  booktitle = {Proceedings of the IEEE/CVF Conference on Computer Vision and Pattern Recognition},
  title     = {Fireedit: Fine-grained instruction-based image editing via region-aware vision language model},
  year      = {2025},
  address   = {Nashville, USA},
  pages     = {13093--13103},
  publisher = {IEEE},
  type      = {Conference Proceedings},
}

@InProceedings{wei2024omniedit,
  author    = {Wei, Cong and Xiong, Zheyang and Ren, Weiming and Du, Xeron and Zhang, Ge and Chen, Wenhu},
  booktitle = {Proceedings of the International Conference on Learning Representations},
  title     = {Omniedit: Building image editing generalist models through specialist supervision},
  year      = {2024},
  pages = {1--20},
  address   = {Vienna, Austria},
  publisher = {OpenReview.net},
  type      = {Conference Proceedings},
}

@InProceedings{wang2025unicombine,
  author    = {Wang, Haoxuan and Peng, Jinlong and He, Qingdong and Yang, Hao and Jin, Ying and Wu, Jiafu and Hu, Xiaobin and Pan, Yanjie and Gan, Zhenye and Chi, Mingmin},
  booktitle = {Proceedings of the IEEE/CVF International Conference on Computer Vision},
  title     = {Unicombine: Unified multi-conditional combination with diffusion transformer},
  year      = {2025},
  address   = {Marrakech, Morocco},
  pages     = {18325-18334},
  publisher = {IEEE},
}

@Article{cai2025hidreami,
  author  = {Qi Cai and Jingwen Chen and Yang Chen and Yehao Li and Fuchen Long and Yingwei Pan and Zhaofan Qiu and Yiheng Zhang and Fengbin Gao and Peihan Xu and Yimeng Wang and Kai Yu and Wenxuan Chen and Ziwei Feng and Zijian Gong and Jianzhuang Pan and Yi Peng and Rui Tian and Siyu Wang and Bo Zhao and Ting Yao and Tao Mei},
  journal = {arXiv preprint arXiv:2505.22705},
  title   = {HiDream-I1: {A} High-Efficient Image Generative Foundation Model with Sparse Diffusion Transformer},
  year    = {2025},
  type    = {Journal Article},
}

@InProceedings{fu2024guiding,
  author    = {Tsu{-}Jui Fu and Wenze Hu and Xianzhi Du and William Yang Wang and Yinfei Yang and Zhe Gan},
  booktitle = {Proceedings of the International Conference on Learning Representations},
  title     = {Guiding Instruction-based Image Editing via Multimodal Large Language Models},
  year      = {2024},
  pages = {1--20},
  address   = {Vienna, Austria},
  publisher = {OpenReview.net},
  type      = {Conference Proceedings},
}

@InProceedings{guo2024focus,
  author    = {Guo, Qin and Lin, Tianwei},
  booktitle = {Proceedings of the IEEE/CVF Conference on Computer Vision and Pattern Recognition},
  title     = {Focus on your instruction: Fine-grained and multi-instruction image editing by attention modulation},
  year      = {2024},
  address   = {Seattle, USA},
  pages     = {6986--6996},
  publisher = {IEEE},
  type      = {Conference Proceedings},
}

@InProceedings{mao2025ace,
  author    = {Mao, Chaojie and Zhang, Jingfeng and Pan, Yulin and Jiang, Zeyinzi and Han, Zhen and Liu, Yu and Zhou, Jingren},
  booktitle = {Proceedings of the IEEE/CVF International Conference on Computer Vision},
  title     = {Ace++: Instruction-based image creation and editing via context-aware content filling},
  year      = {2025},
  address   = {Marrakech, Morocco},
  pages     = {1958-1966},
  publisher = {IEEE},
}

@InProceedings{kawar2023imagic,
  author    = {Kawar, Bahjat and Zada, Shiran and Lang, Oran and Tov, Omer and Chang, Huiwen and Dekel, Tali and Mosseri, Inbar and Irani, Michal},
  booktitle = {Proceedings of the IEEE/CVF Conference on Computer Vision and Pattern Recognition},
  title     = {Imagic: Text-based real image editing with diffusion models},
  year      = {2023},
  address   = {Vancouver, Canada},
  pages     = {6007--6017},
  publisher = {IEEE},
  type      = {Conference Proceedings},
}

@InProceedings{mokady2023nulltext,
  author    = {Mokady, Ron and Hertz, Amir and Aberman, Kfir and Pritch, Yael and Cohen-Or, Daniel},
  booktitle = {Proceedings of the IEEE/CVF Conference on Computer Vision and Pattern Recognition},
  title     = {Null-text inversion for editing real images using guided diffusion models},
  year      = {2023},
  address   = {Vancouver, Canada},
  pages     = {6038--6047},
  publisher = {IEEE},
  type      = {Conference Proceedings},
}

@Article{valevski2023unitune,
  author  = {Valevski, Dani and Kalman, Matan and Molad, Eyal and Segalis, Eyal and Matias, Yossi and Leviathan, Yaniv},
  journal = {ACM Trans. Graph.},
  title   = {Unitune: Text-driven image editing by fine tuning a diffusion model on a single image},
  year    = {2023},
  issn    = {0730-0301},
  number  = {4},
  pages   = {1--10},
  volume  = {42},
  type    = {Journal Article},
}

@InProceedings{avrahami2023breakascene,
  author    = {Avrahami, Omri and Aberman, Kfir and Fried, Ohad and Cohen-Or, Daniel and Lischinski, Dani},
  booktitle = {Proceedings of the SIGGRAPH Asia Conference Papers},
  title     = {Break-a-scene: Extracting multiple concepts from a single image},
  year      = {2023},
  address   = {Sydney, Australia},
  pages     = {1--12},
  publisher = {ACM},
  type      = {Conference Proceedings},
}

@InProceedings{jiang2025energyguided,
  author    = {Jiang, Rui and Fu, Xinghe and Zheng, Guangcong and Li, Teng and Yao, Taiping and Li, Xi},
  booktitle = {Proceedings of the AAAI Conference on Artificial Intelligence},
  title     = {Energy-guided optimization for personalized image editing with pretrained text-to-image diffusion models},
  year      = {2025},
  address   = {Philadelphia, USA},
  pages     = {4048--4056},
  publisher = {AAAI},
  volume    = {39},
  isbn      = {2374-3468},
  type      = {Conference Proceedings},
}

@InProceedings{dong2023ptinversion,
  author    = {Wenkai Dong and Song Xue and Xiaoyue Duan and Shumin Han},
  booktitle = {Proceedings of the IEEE/CVF International Conference on Computer Vision},
  title     = {Prompt Tuning Inversion for Text-Driven Image Editing Using Diffusion Models},
  year      = {2023},
  address   = {Paris, France},
  pages     = {7396--7406},
  publisher = {IEEE},
}

@Article{li2023stylediffusion,
  author  = {Senmao Li and Joost van de Weijer and Taihang Hu and Fahad Shahbaz Khan and Qibin Hou and Yaxing Wang and Jian Yang},
  journal = {arXiv preprint arXiv:2303.15649},
  title   = {StyleDiffusion: Prompt-Embedding Inversion for Text-Based Editing},
  year    = {2023},
}

@Article{wang2023instructedit,
  author  = {Qian Wang and Biao Zhang and Michael Birsak and Peter Wonka},
  journal = {arXiv preprint arXiv:2305.18047},
  title   = {InstructEdit: Improving Automatic Masks for Diffusion-based Image Editing With User Instructions},
  year    = {2023},
}

@InProceedings{parmar2023pix2pixzero,
  author    = {Parmar, Gaurav and Kumar Singh, Krishna and Zhang, Richard and Li, Yijun and Lu, Jingwan and Zhu, Jun-Yan},
  booktitle = {ACM SIGGRAPH 2023 Conference Proceedings},
  title     = {Zero-shot image-to-image translation},
  year      = {2023},
  address   = {New York, USA},
  pages     = {1--11},
  publisher = {ACM},
  type      = {Conference Proceedings},
}

@InProceedings{zhang2021clwd,
  author    = {Liu, Yang and Zhu, Zhen and Bai, Xiang},
  booktitle = {Proceedings of the IEEE/CVF Winter Conference on Applications of Computer Vision},
  title     = {Wdnet: Watermark-decomposition network for visible watermark removal},
  year      = {2021},
  address   = {Virtual},
  pages     = {3685-3693},
  publisher = {IEEE},
  type      = {Conference Proceedings},
}

@InProceedings{qiao2024baret,
  author    = {Qiao, Yuming and Wang, Fanyi and Su, Jingwen and Zhang, Yanhao and Yu, Yunjie and Wu, Siyu and Qi, Guo-Jun},
  booktitle = {Proceedings of the AAAI Conference on Artificial Intelligence},
  title     = {Baret: Balanced attention based real image editing driven by target-text inversion},
  year      = {2024},
  address   = {Vancouver, Canada},
  pages     = {4560--4568},
  publisher = {AAAI},
  volume    = {38},
  isbn      = {2374-3468},
  type      = {Conference Proceedings},
}

@InProceedings{meng2022sdedit,
  author    = {Chenlin Meng and Yutong He and Yang Song and Jiaming Song and Jiajun Wu and Jun{-}Yan Zhu and Stefano Ermon},
  booktitle = {Proceedings of the International Conference on Learning Representations},
  title     = {SDEdit: Guided Image Synthesis and Editing with Stochastic Differential Equations},
  year      = {2022},
  pages = {1--20},
  address   = {Virtual},
  publisher = {OpenReview.net},
}

@Article{hertz2022prompttoprompt,
  author  = {Hertz, Amir and Mokady, Ron and Tenenbaum, Jay and Aberman, Kfir and Pritch, Yael and Cohen-Or, Daniel},
  journal = {arXiv preprint arXiv:2208.01626},
  title   = {Prompt-to-prompt image editing with cross attention control},
  year    = {2022},
  type    = {Journal Article},
}

@InProceedings{couairon2022diffedit,
  author    = {Couairon, Guillaume and Verbeek, Jakob and Schwenk, Holger and Cord, Matthieu},
  booktitle = {Proceedings of the International Conference on Learning Representations},
  title     = {Diffedit: Diffusion-based semantic image editing with mask guidance},
  year      = {2022},
  pages = {1--20},
  address   = {Virtual},
  publisher = {OpenReview.net},
}

@InProceedings{tumanyan2023plugandplay,
  author    = {Tumanyan, Narek and Geyer, Michal and Bagon, Shai and Dekel, Tali},
  booktitle = {Proceedings of the IEEE/CVF Conference on Computer Vision and Pattern Recognition},
  title     = {Plug-and-play diffusion features for text-driven image-to-image translation},
  year      = {2023},
  address   = {Vancouver, Canada},
  pages     = {1921--1930},
  publisher = {IEEE},
  type      = {Conference Proceedings},
}

@InProceedings{titov2024guideandrescale,
  author    = {Titov, Vadim and Khalmatova, Madina and Ivanova, Alexandra and Vetrov, Dmitry and Alanov, Aibek},
  booktitle = {European Conference on Computer Vision},
  title     = {Guide-and-rescale: Self-guidance mechanism for effective tuning-free real image editing},
  year      = {2024},
  address   = {Milan, Italy},
  pages     = {235--251},
  publisher = {Springer},
  type      = {Conference Proceedings},
}

@InProceedings{li2024zone,
  author    = {Li, Shanglin and Zeng, Bohan and Feng, Yutang and Gao, Sicheng and Liu, Xiuhui and Liu, Jiaming and Li, Lin and Tang, Xu and Hu, Yao and Liu, Jianzhuang},
  booktitle = {Proceedings of the IEEE/CVF Conference on Computer Vision and Pattern Recognition},
  title     = {Zone: Zero-shot instruction-guided local editing},
  year      = {2024},
  address   = {Seattle, USA},
  pages     = {6254--6263},
  publisher = {IEEE},
  type      = {Conference Proceedings},
}

@Article{yaowei2024brushedit,
  author  = {Yaowei Li and Yuxuan Bian and Xuan Ju and Zhaoyang Zhang and Ying Shan and Yuexian Zou and Qiang Xu},
  journal = {arXiv preprint arXiv:2412.10316},
  title   = {BrushEdit: All-In-One Image Inpainting and Editing},
  year    = {2024},
  type    = {Journal Article},
}

@InProceedings{simsar2024uip2p,
  author    = {Simsar, Enis and Tonioni, Alessio and Xian, Yongqin and Hofmann, Thomas and Tombari, Federico},
  booktitle = {Proceedings of the IEEE/CVF International Conference on Computer Vision},
  title     = {Uip2p: Unsupervised instruction-based image editing via cycle edit consistency},
  year      = {2025},
  address   = {Marrakech, Morocco},
  pages     = {18895-18905},
  publisher = {IEEE},
}

@InProceedings{brack2024ledits,
  author    = {Brack, Manuel and Friedrich, Felix and Kornmeier, Katharia and Tsaban, Linoy and Schramowski, Patrick and Kersting, Kristian and Passos, Apolinário},
  booktitle = {Proceedings of the IEEE/CVF Conference on Computer Vision and Pattern Recognition},
  title     = {Ledits++: Limitless image editing using text-to-image models},
  year      = {2024},
  address   = {Seattle, USA},
  pages     = {8861--8870},
  publisher = {IEEE},
  type      = {Conference Proceedings},
}

@InProceedings{cao2023masactrl,
  author    = {Cao, Mingdeng and Wang, Xintao and Qi, Zhongang and Shan, Ying and Qie, Xiaohu and Zheng, Yinqiang},
  booktitle = {Proceedings of the IEEE/CVF International Conference on Computer Vision},
  title     = {Masactrl: Tuning-free mutual self-attention control for consistent image synthesis and editing},
  year      = {2023},
  address   = {Paris, France},
  pages     = {22560--22570},
  publisher = {IEEE},
  type      = {Conference Proceedings},
}

@InProceedings{hertz2023delta,
  author    = {Hertz, Amir and Aberman, Kfir and Cohen-Or, Daniel},
  booktitle = {Proceedings of the IEEE/CVF International Conference on Computer Vision},
  title     = {Delta denoising score},
  year      = {2023},
  address   = {Paris, France},
  pages     = {2328--2337},
  publisher = {IEEE},
  type      = {Conference Proceedings},
}

@InProceedings{ju2024pnpinversion,
  author    = {Xuan Ju and Ailing Zeng and Yuxuan Bian and Shaoteng Liu and Qiang Xu},
  booktitle = {Proceedings of the International Conference on Learning Representations},
  title     = {PnP Inversion: Boosting Diffusion-based Editing with 3 Lines of Code},
  year      = {2024},
  pages = {1--20},
  address   = {Vienna, Austria},
  publisher = {OpenReview.net},
}

@InProceedings{arar2025negative,
  author    = {Arar, Moab and Voynov, Andrey and Fruchter, Sivan and Cohen-Or, Daniel},
  booktitle = {Proceedings of the IEEE/CVF Winter Conference on Applications of Computer Vision},
  title     = {Negative-prompt inversion: Fast image inversion for editing with text-guided diffusion models},
  year      = {2025},
  address   = {Tucson, USA},
  pages     = {2063-2072},
  publisher = {IEEE},
  type      = {Conference Proceedings},
}

@InProceedings{wallace2023edict,
  author    = {Wallace, Bram and Gokul, Akash and Naik, Nikhil},
  booktitle = {Proceedings of the IEEE/CVF Conference on Computer Vision and Pattern Recognition},
  title     = {{EDICT}: Exact Diffusion Inversion via Coupled Transformations},
  year      = {2023},
  address   = {Vancouver, Canada},
  pages     = {22532--22541},
  publisher = {IEEE},
  type      = {Conference Proceedings},
}

@InProceedings{han2024proxedit,
  author    = {Ligong Han and Song Wen and Qi Chen and Zhixing Zhang and Kunpeng Song and Mengwei Ren and Ruijiang Gao and Anastasis Stathopoulos and Xiaoxiao He and Yuxiao Chen and Di Liu and Qilong Zhangli and Jindong Jiang and Zhaoyang Xia and Akash Srivastava and Dimitris N. Metaxas},
  booktitle = {Proceedings of the IEEE/CVF Winter Conference on Applications of Computer Vision},
  title     = {ProxEdit: Improving Tuning-Free Real Image Editing with Proximal Guidance},
  year      = {2024},
  address   = {Waikoloa, USA},
  pages     = {4279--4289},
  publisher = {IEEE},
}

@InProceedings{huberman2024ddpminv,
  author    = {Inbar Huberman{-}Spiegelglas and Vladimir Kulikov and Tomer Michaeli},
  booktitle = {Proceedings of the IEEE/CVF Conference on Computer Vision and Pattern Recognition},
  title     = {An Edit Friendly {DDPM} Noise Space: Inversion and Manipulations},
  year      = {2024},
  address   = {Seattle, USA},
  pages     = {12469--12478},
  publisher = {IEEE},
}

@InProceedings{song2021ddim,
  author    = {Song, Jiaming and Meng, Chenlin and Ermon, Stefano},
  booktitle = {Proceedings of the International Conference on Learning Representations},
  title     = {Denoising Diffusion Implicit Models},
  year      = {2021},
  pages = {1--20},
  address   = {Vienna, Austria},
  publisher = {OpenReview.net},
  type      = {Conference Proceedings},
}

@InProceedings{lin2024scheduledit,
  author    = {Haonan Lin and Yan Chen and Jiahao Wang and Wenbin An and Mengmeng Wang and Feng Tian and Yong Liu and Guang Dai and Jingdong Wang and Qianying Wang},
  booktitle = {Advances in Neural Information Processing Systems},
  title     = {Schedule Your Edit: A Simple Yet Effective Diffusion Noise Schedule for Image Editing},
  year      = {2024},
  address   = {Vancouver, Canada},
  pages     = {115712-115756},
  publisher = {Curran Associates},
  type      = {Conference Proceedings},
}

@InProceedings{nie2024,
  author    = {Shen Nie and Hanzhong Allan Guo and Cheng Lu and Yuhao Zhou and Chenyu Zheng and Chongxuan Li},
  booktitle = {Proceedings of the International Conference on Learning Representations},
  title     = {The Blessing of Randomness: {SDE} Beats {ODE} in General Diffusion-based Image Editing},
  year      = {2024},
  pages = {1--20},
  address   = {Vienna, Austria},
  publisher = {OpenReview.net},
}

@InProceedings{hu2024instructimagen,
  author    = {Hu, Hexiang and Chan, Kelvin CK and Su, Yu-Chuan and Chen, Wenhu and Li, Yandong and Sohn, Kihyuk and Zhao, Yang and Ben, Xue and Gong, Boqing and Cohen, William},
  booktitle = {Proceedings of the IEEE/CVF Conference on Computer Vision and Pattern Recognition},
  title     = {Instruct-imagen: Image generation with multi-modal instruction},
  year      = {2024},
  address   = {Seattle, USA},
  pages     = {4754--4763},
  publisher = {IEEE},
  type      = {Conference Proceedings},
}

@InProceedings{phung2024grounded,
  author    = {Phung, Quynh and Ge, Songwei and Huang, Jia-Bin},
  booktitle = {Proceedings of the IEEE/CVF Conference on Computer Vision and Pattern Recognition},
  title     = {Grounded text-to-image synthesis with attention refocusing},
  year      = {2024},
  address   = {Seattle, USA},
  pages     = {7932--7942},
  publisher = {IEEE},
  type      = {Conference Proceedings},
}

@Article{yu2023cm3leon,
  author  = {Yu, Lili and Shi, Bowen and Pasunuru, Ramakanth and Muller, Benjamin and Golovneva, Olga and Wang, Tianlu and Babu, Arun and Tang, Binh and Karrer, Brian and Sheynin, Shelly},
  journal = {arXiv preprint arXiv:2309.02591},
  title   = {Scaling autoregressive multi-modal models: Pretraining and instruction tuning},
  year    = {2023},
  type    = {Journal Article},
}

@InProceedings{mu2025editar,
  author    = {Mu, Jiteng and Vasconcelos, Nuno and Wang, Xiaolong},
  booktitle = {Proceedings of the IEEE/CVF Conference on Computer Vision and Pattern Recognition},
  title     = {Editar: Unified conditional generation with autoregressive models},
  year      = {2025},
  address   = {Nashville, USA},
  pages     = {7899--7909},
  publisher = {IEEE},
  type      = {Conference Proceedings},
}

@Article{zhang2025nexusgen,
  author  = {Zhang, Hong and Duan, Zhongjie and Wang, Xingjun and Zhao, Yuze and Lu, Weiyi and Di, Zhipeng and Xu, Yixuan and Chen, Yingda and Zhang, Yu},
  journal = {arXiv preprint arXiv:2504.21356},
  title   = {Nexus-gen: A unified model for image understanding, generation, and editing},
  year    = {2025},
  type    = {Journal Article},
}

@InProceedings{lai2025instamanip,
  author    = {Lai, Bolin and Juefei-Xu, Felix and Liu, Miao and Dai, Xiaoliang and Mehta, Nikhil and Zhu, Chenguang and Huang, Zeyi and Rehg, James M and Lee, Sangmin and Zhang, Ning},
  booktitle = {Proceedings of the IEEE/CVF Conference on Computer Vision and Pattern Recognition},
  title     = {Unleashing in-context learning of autoregressive models for few-shot image manipulation},
  year      = {2025},
  address   = {Nashville, USA},
  pages     = {18346--18357},
  publisher = {IEEE},
  type      = {Conference Proceedings},
}

@Article{fan2025unified,
  author  = {Fan, Lijie and Tang, Luming and Qin, Siyang and Li, Tianhong and Yang, Xuan and Qiao, Siyuan and Steiner, Andreas and Sun, Chen and Li, Yuanzhen and Zhu, Tao},
  journal = {arXiv preprint arXiv:2503.13436},
  title   = {Unified autoregressive visual generation and understanding with continuous tokens},
  year    = {2025},
  type    = {Journal Article},
}

@Article{cheng2024dragsence,
  author  = {Chenghao Gu and Zhenzhe Li and Zhengqi Zhang and Yunpeng Bai and Shuzhao Xie and Zhi Wang},
  journal = {arXiv preprint arXiv: 2412.13552},
  title   = {DragScene: Interactive 3D Scene Editing with Single-view Drag Instructions},
  year    = {2024},
  type    = {Journal Article},
}

@InProceedings{wang2025trainingfree,
  author    = {Wang, Yufei and Guo, Lanqing and Li, Zhihao and Huang, Jiaxing and Wang, Pichao and Wen, Bihan and Wang, Jian},
  booktitle = {Proceedings of the IEEE/CVF International Conference on Computer Vision},
  title     = {Training-Free Text-Guided Image Editing with Visual Autoregressive Model},
  year      = {2025},
  address   = {Marrakech, Morocco},
  pages     = {17577-17586},
  publisher = {IEEE},
}

@InProceedings{tai2025islock,
  author    = {Taihang Hu and Linxuan Li and Kai Wang and Yaxing Wang and Jian Yang and Ming{-}Ming Cheng},
  booktitle = {Proceedings of the IEEE/CVF International Conference on Computer Vision},
  title     = {Anchor Token Matching: Implicit Structure Locking for Training-free {AR} Image Editing},
  year      = {2025},
  address   = {Marrakech, Morocco},
  pages     = {18166-18176},
  publisher = {IEEE},
}

@InProceedings{lu2024autoregressive,
  author    = {Lu, Zhuqiang and Hu, Kun and Wang, Chaoyue and Bai, Lei and Wang, Zhiyong},
  booktitle = {Proceedings of the AAAI Conference on Artificial Intelligence},
  title     = {Autoregressive omni-aware outpainting for open-vocabulary 360-degree image generation},
  year      = {2024},
  address   = {Vancouver, Canada},
  pages     = {14211--14219},
  publisher = {AAAI},
  volume    = {38},
  isbn      = {2374-3468},
  type      = {Conference Proceedings},
}

@InProceedings{bachmann20244m2i,
  author    = {Bachmann, Roman and Kar, Oğuzhan F and Mizrahi, David and Garjani, Ali and Gao, Mingfei and Griffiths, David and Hu, Jiaming and Dehghan, Afshin and Zamir, Amir},
  booktitle = {Advances in Neural Information Processing Systems},
  title     = {4m-21: An any-to-any vision model for tens of tasks and modalities},
  year      = {2024},
  address   = {Vancouver, Canada},
  pages     = {61872--61911},
  publisher = {Curran Associates},
}

@InProceedings{lu2024unifiedio,
  author    = {Lu, Jiasen and Clark, Christopher and Lee, Sangho and Zhang, Zichen and Khosla, Savya and Marten, Ryan and Hoiem, Derek and Kembhavi, Aniruddha},
  booktitle = {Proceedings of the IEEE/CVF Conference on Computer Vision and Pattern Recognition},
  title     = {Unified-io 2: Scaling autoregressive multimodal models with vision language audio and action},
  year      = {2024},
  address   = {Seattle, USA},
  pages     = {26439--26455},
  publisher = {IEEE},
  type      = {Conference Proceedings},
}

@Article{wang2025ovisu,
  author  = {Wang, Guo-Hua and Zhao, Shanshan and Zhang, Xinjie and Cao, Liangfu and Zhan, Pengxin and Duan, Lunhao and Lu, Shiyin and Fu, Minghao and Chen, Xiaohao and Zhao, Jianshan},
  journal = {arXiv preprint arXiv:2506.23044},
  title   = {Ovis-U1 Technical Report},
  year    = {2025},
  type    = {Journal Article},
}

@InProceedings{sheynin2024emuedit,
  author    = {Sheynin, Shelly and Polyak, Adam and Singer, Uriel and Kirstain, Yuval and Zohar, Amit and Ashual, Oron and Parikh, Devi and Taigman, Yaniv},
  booktitle = {Proceedings of the IEEE/CVF Conference on Computer Vision and Pattern Recognition},
  title     = {Emu edit: Precise image editing via recognition and generation tasks},
  year      = {2024},
  address   = {Seattle, USA},
  pages     = {8871--8879},
  publisher = {IEEE},
  type      = {Conference Proceedings},
}

@Article{labs2025flux,
  author  = {Labs, Black Forest and Batifol, Stephen and Blattmann, Andreas and Boesel, Frederic and Consul, Saksham and Diagne, Cyril and Dockhorn, Tim and English, Jack and English, Zion and Esser, Patrick},
  journal = {arXiv preprint arXiv:2506.15742},
  title   = {FLUX. 1 Kontext: Flow Matching for In-Context Image Generation and Editing in Latent Space},
  year    = {2025},
  type    = {Journal Article},
}

@Article{wu2025omnigen,
  author  = {Wu, Chenyuan and Zheng, Pengfei and Yan, Ruiran and Xiao, Shitao and Luo, Xin and Wang, Yueze and Li, Wanli and Jiang, Xiyan and Liu, Yexin and Zhou, Junjie},
  journal = {arXiv preprint arXiv:2506.18871},
  title   = {OmniGen2: Exploration to Advanced Multimodal Generation},
  year    = {2025},
  type    = {Journal Article},
}

@InProceedings{tu2025dreamo,
  author    = {Tu, Tao and Li, Ming-Feng and Lin, Chieh Hubert and Cheng, Yen-Chi and Sun, Min and Yang, Ming-Hsuan},
  booktitle = {Proceedings of the IEEE/CVF Winter Conference on Applications of Computer Vision},
  title     = {Dreamo: Articulated 3d reconstruction from a single casual video},
  year      = {2025},
  address   = {Tucson, USA},
  pages     = {2269--2279},
  publisher = {IEEE},
  isbn      = {9798331510831},
  type      = {Conference Proceedings},
}

@Article{xia2025dreamomni2,
  author  = {Xia, Bin and Peng, Bohao and Zhang, Yuechen and Huang, Junjia and Liu, Jiyang and Li, Jingyao and Tan, Haoru and Wu, Sitong and Wang, Chengyao and Wang, Yitong and others},
  journal = {arXiv preprint arXiv:2510.06679},
  title   = {DreamOmni2: Multimodal Instruction-based Editing and Generation},
  year    = {2025},
}

@InProceedings{xiao2025omnige,
  author    = {Shitao Xiao and Yueze Wang and Junjie Zhou and Huaying Yuan and Xingrun Xing and Ruiran Yan and Chaofan Li and Shuting Wang and Tiejun Huang and Zheng Liu},
  booktitle = {Proceedings of the IEEE/CVF Conference on Computer Vision and Pattern Recognition},
  title     = {OmniGen: Unified Image Generation},
  year      = {2025},
  address   = {Nashville, USA},
  pages     = {13294--13304},
  publisher = {IEEE},
  type      = {Conference Proceedings},
}

@Article{openai2024gpt4o,
  author  = {OpenAI},
  journal = {arXiv preprint arXiv:2410.21276},
  title   = {GPT-4o System Card},
  year    = {2024},
  school  = {OpenAI},
}

@Misc{openai2025gpt4-1,
  author = {{OpenAI}},
  note   = {Accessed 14 April 2025},
  title  = {Introducing GPT-4.1 in the API},
  year   = {2025},
  school = {OpenAI},
  url    = {https://openai.com/index/gpt-4-1},
}

@Article{google2025gemini2-5,
  author  = {{Google DeepMind Gemini Team}},
  journal = {arXiv preprint arXiv:2403.05530},
  title   = {Gemini 1.5: Unlocking Multimodal Understanding Across Millions of Tokens},
  year    = {2025},
}

@Misc{openai2023dalle3,
  author = {{OpenAI}},
  note   = {Accessed 03 October 2023},
  title  = {DALL·E 3 System Card},
  year   = {2023},
  school = {OpenAI},
  url    = {https://openai.com/index/dall-e-3-system-card},
}

@Misc{artificial2025grok3,
  author = {{Artificial Analysis}},
  note   = {Accessed 31 December 2025},
  title  = {Grok 3 vs Gemini 2.5 Pro: Model Comparison},
  year   = {2025},
  url    = {https://artificialanalysis.ai/models/comparisons/gemini-2-5-pro-vs-grok-3},
}

@Article{bytedance2025seededit,
  author  = {Peng Wang and Yichun Shi and Xiaochen Lian and Zhonghua Zhai and Xin Xia and Xuefeng Xiao and Weilin Huang and Jianchao Yang},
  journal = {arXiv preprint arXiv:2506.05083},
  title   = {SeedEdit 3.0: Fast and High-Quality Generative Image Editing},
  year    = {2025},
}

@Misc{google2025imagen3,
  author = {{Google Research}},
  note   = {Accessed 22 July 2025},
  title  = {{Imagen 3}: High-Fidelity Image Editing with {Gemini} Integration},
  year   = {2025},
  url    = {https://imagen.research.google/},
}

@Misc{qwen2025qwen25vlo,
  author = {{Qwen Team}},
  note   = {Accessed 03 August 2025},
  title  = {Qwen VLo: From "Understanding" the World to "Depicting" It},
  year   = {2025},
  url    = {https://qwenlm.github.io/blog/qwen-vlo/},
}

@Article{bai2025qwen2,
  author  = {{Qwen Team}},
  journal = {arXiv preprint arXiv:2502.13923},
  title   = {Qwen2.5-VL Technical Report},
  year    = {2025},
}

@Misc{adobe2024firefly,
  author = {{Adobe}},
  note   = {Accessed 22 July 2025},
  title  = {Adobe {Firefly Image 2 Model}},
  year   = {2024},
  url    = {https://www.adobe.com/products/firefly.html},
}

@Misc{google2025gemini2flash,
  author       = {{Google}},
  howpublished = {Google Developers Blog},
  note         = {Accessed 22 July 2025},
  title        = {Experiment with {Gemini 2.0 Flash} native image generation},
  year         = {2025},
  url          = {https://developers.googleblog.com/en/experiment- with-gemini-20-flash-native-image-generation/},
}

@Misc{midjourney2024v6,
  author = {{Midjourney}},
  note   = {Accessed 22 July 2025},
  title  = {Midjourney {V6} Release Notes},
  year   = {2024},
  url    = {https://docs.midjourney.com/hc/en-us},
}

@Misc{anthropic2024claude35,
  author = {{Anthropic}},
  note   = {Accessed 22 July 2025},
  title  = {Claude 3.5 Sonnet},
  year   = {2024},
  url    = {https://www.anthropic.com/news/claude-3-5-sonnet},
}

@Article{li2025uniworldv2,
  author  = {Li, Zongjian and Liu, Zheyuan and Zhang, Qihui and Lin, Bin and Yuan, Shenghai and Yan, Zhiyuan and Ye, Yang and Yu, Wangbo and Niu, Yuwei and Yuan, Li},
  journal = {arXiv preprint arXiv:2510.16888},
  title   = {Uniworld-V2: Reinforce Image Editing with Diffusion Negative-aware Finetuning and MLLM Implicit Feedback},
  year    = {2025},
}

@Misc{flux-2-2025,
  author = {Black Forest Labs},
  note   = {Accessed 25 November 2025},
  title  = {{FLUX.2: Frontier Visual Intelligence}},
  year   = {2025},
  url    = {https://bfl.ai/blog/flux-2},
}

@Article{shi2024seededit,
  author  = {Shi, Yichun and Wang, Peng and Huang, Weilin},
  journal = {arXiv preprint arXiv:2411.06686},
  title   = {Seededit: Align image re-generation to image editing},
  year    = {2024},
  type    = {Journal Article},
}

@InProceedings{jiang2025anyedit,
  author    = {Jiang, Houcheng and Fang, Junfeng and Zhang, Ningyu and Ma, Guojun and Wan, Mingyang and Wang, Xiang and He, Xiangnan and Chua, Tat-seng},
  booktitle = {Proceedings of the International Conference on Machine Learning},
  title     = {Anyedit: Edit any knowledge encoded in language models},
  year      = {2025},
  address   = {Vancouver, Canada},
  pages     = {27510-27533},
  publisher = {ACM},
}

@Article{huang2025diffusion,
  author  = {Huang, Yi and Huang, Jiancheng and Liu, Yifan and Yan, Mingfu and Lv, Jiaxi and Liu, Jianzhuang and Xiong, Wei and Zhang, He and Cao, Liangliang and Chen, Shifeng},
  journal = {IEEE Trans. Pattern Anal. Mach. Intell.},
  title   = {Diffusion model-based image editing: A survey},
  year    = {2025},
  number  = {6},
  pages   = {4409 - 4437},
  volume  = {47},
  type    = {Journal Article},
}

@InProceedings{wu2023a,
  author    = {Wu, Chen Henry and De la Torre, Fernando},
  booktitle = {Proceedings of the IEEE/CVF International Conference on Computer Vision},
  title     = {A latent space of stochastic diffusion models for zero-shot image editing and guidance},
  year      = {2023},
  address   = {Paris, France},
  pages     = {7378--7387},
  publisher = {IEEE},
  type      = {Conference Proceedings},
}

@InProceedings{goodfellow2014generative,
  author    = {Goodfellow, Ian J and Pouget-Abadie, Jean and Mirza, Mehdi and Xu, Bing and Warde-Farley, David and Ozair, Sherjil and Courville, Aaron and Bengio, Yoshua},
  title     = {Generative adversarial nets},
  year      = {2014},
  address   = {Montreal, Canada},
  publisher = {Curran Associates},
  pages = {2672--2680},
  booktitle   = {Advances in Neural Information Processing Systems},
}

@InProceedings{pathak2016context,
  author    = {Pathak, Deepak and Krahenbuhl, Philipp and Donahue, Jeff and Darrell, Trevor and Efros, Alexei A},
  booktitle = {Proceedings of the IEEE/CVF Conference on Computer Vision and Pattern Recognition},
  title     = {Context encoders: Feature learning by inpainting},
  year      = {2016},
  address   = {Las Vegas, USA},
  pages     = {2536--2544},
  publisher = {IEEE},
}

@InProceedings{ledig2017photo,
  author    = {Ledig, Christian and Theis, Lucas and Husz{\'a}r, Ferenc and Caballero, Jose and Cunningham, Andrew and Acosta, Alejandro and Aitken, Andrew and Tejani, Alykhan and Totz, Johannes and Wang, Zehan and others},
  booktitle = {Proceedings of the IEEE/CVF Conference on Computer Vision and Pattern Recognition},
  title     = {Photo-realistic single image super-resolution using a generative adversarial network},
  year      = {2017},
  address   = {Honolulu, USA},
  pages     = {4681--4690},
  publisher = {IEEE},
}

@InProceedings{qin2021watermarkremoval,
  author={Cheng, Danni and Li, Xiang and Li, Wei-Hong and Lu, Chan and Li, Fake and Zhao, Hua and Zheng, Wei-Shi},
  title={Large-Scale Visible Watermark Detection and Removal with Deep Convolutional Networks},
  booktitle={Proceedings of the Chinese Conference on Pattern Recognition and Computer Vision},
  year={2018},
  publisher={Springer},
  address={Guangzhou, China},
  pages={27--40},
}

@InProceedings{liu2024grounding,
  author    = {Liu, Shilong and Zeng, Zhaoyang and Ren, Tianhe and Li, Feng and Zhang, Hao and Yang, Jie and Jiang, Qing and Li, Chunyuan and Yang, Jianwei and Su, Hang and others},
  booktitle = {European Conference on Computer Vision},
  title     = {Grounding dino: Marrying dino with grounded pre-training for open-set object detection},
  year      = {2024},
  address   = {Milan, Italy},
  pages     = {38--55},
  publisher = {Springer},
}

@InProceedings{kirillov2023segment,
  author    = {Kirillov, Alexander and Mintun, Eric and Ravi, Nikhila and Mao, Hanzi and Rolland, Chloe and Gustafson, Laura and Xiao, Tete and Whitehead, Spencer and Berg, Alexander C and Lo, Wan-Yen and others},
  booktitle = {Proceedings of the IEEE/CVF International Conference on Computer Vision},
  title     = {Segment anything},
  year      = {2023},
  address   = {Paris, France},
  pages     = {4015--4026},
  publisher = {IEEE},
}

@InProceedings{ravi2024sam,
  author    = {Ravi, Nikhila and Gabeur, Valentin and Hu, Yuan-Ting and Hu, Ronghang and Ryali, Chaitanya and Ma, Tengyu and Khedr, Haitham and R{\"a}dle, Roman and Rolland, Chloe and Gustafson, Laura and others},
  booktitle = {Proceedings of the International Conference on Learning Representations},
  title     = {Sam 2: Segment anything in images and videos},
  year      = {2024},
  pages = {1--20},
  address   = {Vienna, Austria},
  publisher = {OpenReview.net},
}

@InProceedings{zhang2024recognize,
  author    = {Zhang, Youcai and Huang, Xinyu and Ma, Jinyu and Li, Zhaoyang and Luo, Zhaochuan and Xie, Yanchun and Qin, Yuzhuo and Luo, Tong and Li, Yaqian and Liu, Shilong and others},
  booktitle = {Proceedings of the IEEE/CVF Conference on Computer Vision and Pattern Recognition},
  title     = {Recognize anything: A strong image tagging model},
  year      = {2024},
  address   = {Seattle, USA},
  pages     = {1724--1732},
  publisher = {IEEE},
}

@Article{zheng2024bilateral,
  author  = {Zheng, Peng and Gao, Dehong and Fan, Deng-Ping and Liu, Li and Laaksonen, Jorma and Ouyang, Wanli and Sebe, Nicu},
  journal = {CAAI Artif. Intell. Res.},
  title   = {Bilateral reference for high-resolution dichotomous image segmentation},
  year    = {2024},
  pages   = {1-12},
  volume  = {3},
}

@InProceedings{teed2020raft,
  author    = {Teed, Zachary and Deng, Jia},
  booktitle = {European Conference on Computer Vision},
  title     = {Raft: Recurrent all-pairs field transforms for optical flow},
  year      = {2020},
  address   = {Glasgow, UK},
  pages     = {402--419},
  publisher = {Springer},
}

@InProceedings{hu2022lora,
  author    = {Hu, Edward J and Shen, Yelong and Wallis, Phillip and Allen-Zhu, Zeyuan and Li, Yuanzhi and Wang, Shean and Wang, Lu and Chen, Weizhu and others},
  booktitle = {Proceedings of the International Conference on Learning Representations},
  title     = {Lora: Low-rank adaptation of large language models.},
  year      = {2022},
  pages = {1--17},
  address   = {Virtual},
  publisher = {OpenReview.net},
}

@Article{du2009pp,
  author  = {Du, Yuning and Li, Chenxia and Guo, Ruoyu and Yin, Xiaoting and Liu, Weiwei and Zhou, Jun and Bai, Yifan and Yu, Zilin and Yang, Yehua and Dang, Qingqing and others},
  journal = {arXiv preprint arXiv:2009.09941},
  title   = {Pp-ocr: a practical ultra lightweight OCR system},
  year    = {2009},
}

@Misc{wiki:ComfyUI,
  author = {Comfy-Org},
  note   = {Accessed 31 July 2025},
  title  = {{ComfyUI}},
  year   = {2025},
  url    = {https://github.com/Comfy-Org/ComfyUI},
}

@Article{ramesh2022hierarchical,
  author  = {Ramesh, Aditya and Dhariwal, Prafulla and Nichol, Alex and Chu, Casey and Chen, Mark},
  journal = {arXiv preprint arXiv:2204.06125},
  title   = {Hierarchical text-conditional image generation with clip latents},
  year    = {2022},
}

@InProceedings{wang2025koala,
  author    = {Wang, Qiuheng and Shi, Yukai and Ou, Jiarong and Chen, Rui and Lin, Ke and Wang, Jiahao and Jiang, Boyuan and Yang, Haotian and Zheng, Mingwu and Tao, Xin and others},
  booktitle = {Proceedings of the IEEE/CVF Conference on Computer Vision and Pattern Recognition},
  title     = {Koala-36m: A large-scale video dataset improving consistency between fine-grained conditions and video content},
  year      = {2025},
  address   = {Nashville, USA},
  pages     = {8428--8437},
  publisher = {IEEE},
}

@InProceedings{avrahami2022blended,
  author    = {Avrahami, Omri and Lischinski, Dani and Fried, Ohad},
  booktitle = {Proceedings of the IEEE/CVF Conference on Computer Vision and Pattern Recognition},
  title     = {Blended diffusion for text-driven editing of natural images},
  year      = {2022},
  address   = {New Orleans, USA},
  pages     = {18208--18218},
  publisher = {IEEE},
}

@InProceedings{johnson2016perceptual,
  author    = {Johnson, Justin and Alahi, Alexandre and Fei-Fei, Li},
  booktitle = {European Conference on Computer Vision},
  title     = {Perceptual losses for real-time style transfer and super-resolution},
  year      = {2016},
  address   = {Amsterdam, The Netherlands},
  pages     = {694--711},
  publisher = {Springer},
}

@InProceedings{martin2001bsd500,
  author    = {Martin, David and Fowlkes, Charless and Tal, Doron and Malik, Jitendra},
  booktitle = {Proceedings of the IEEE/CVF International Conference on Computer Vision},
  title     = {A database of human segmented natural images and its application to evaluating segmentation algorithms and measuring ecological statistics},
  year      = {2001},
  address   = {Vancouver, Canada},
  pages     = {416-423},
  publisher = {IEEE},
}

@InProceedings{nah2017gopro,
  author    = {Nah, Seungjun and Kim, Tae Hyun and Lee, Kyoung Mu},
  booktitle = {Proceedings of the IEEE/CVF Conference on Computer Vision and Pattern Recognition},
  title     = {Deep multi-scale convolutional neural network for dynamic scene deblurring},
  year      = {2017},
  address   = {Honolulu, USA},
  pages     = {3883-3891},
  publisher = {IEEE},
}

@InProceedings{rim2020realblur,
  author    = {Rim, Joon-Young and Lee, Geonwoon and Won, Jongyoo and Cho, Sunghyun},
  booktitle = {European Conference on Computer Vision},
  title     = {Real-World Blur Dataset for Learning and Benchmarking Deblurring Algorithms},
  year      = {2020},
  address   = {Glasgow, UK},
  pages     = {184-201},
  publisher = {Springer},
}

@InProceedings{abdelhamed2018sidd,
  author    = {Abdelhamed, Abdelrahman and Lin, Stephen and Brown, Michael S},
  booktitle = {Proceedings of the IEEE/CVF Conference on Computer Vision and Pattern Recognition},
  title     = {A High-Quality Denoising Dataset for Smartphone Cameras},
  year      = {2018},
  address   = {Salt Lake City, USA},
  pages     = {1692-1700},
  publisher = {IEEE},
}

@InProceedings{lugmayr2022repaint,
  author    = {Lugmayr, Andreas and Danelljan, Martin and Romero, Andres and Yu, Fisher and Timofte, Radu and Van Gool, Luc},
  booktitle = {Proceedings of the IEEE/CVF Conference on Computer Vision and Pattern Recognition},
  title     = {Repaint: Inpainting using denoising diffusion probabilistic models},
  year      = {2022},
  address   = {New Orleans, USA},
  pages     = {11461--11471},
  publisher = {IEEE},
}

@Article{liew2023magicedit,
  author  = {Liew, Jun Hao and Yan, Hanshu and Zhang, Jianfeng and Xu, Zhongcong and Feng, Jiashi},
  journal = {arXiv preprint arXiv:2308.14749},
  title   = {Magicedit: High-fidelity and temporally coherent video editing},
  year    = {2023},
}

@InProceedings{bao2023sine,
  author    = {Bao, Chong and Zhang, Yinda and Yang, Bangbang and Fan, Tianxing and Yang, Zesong and Bao, Hujun and Zhang, Guofeng and Cui, Zhaopeng},
  booktitle = {Proceedings of the IEEE/CVF Conference on Computer Vision and Pattern Recognition},
  title     = {Sine: Semantic-driven image-based nerf editing with prior-guided editing field},
  year      = {2023},
  address   = {Vancouver, Canada},
  pages     = {20919--20929},
  publisher = {IEEE},
}

@InProceedings{rombach2021highresolution,
  author        = {Robin Rombach and Andreas Blattmann and Dominik Lorenz and Patrick Esser and Björn Ommer},
  booktitle     = {Proceedings of the IEEE/CVF Conference on Computer Vision and Pattern Recognition},
  title         = {High-Resolution Image Synthesis with Latent Diffusion Models},
  year          = {2022},
  address       = {New Orleans, USA},
  pages         = {10684-10695},
  publisher     = {IEEE},
  archiveprefix = {arXiv},
  eprint        = {2112.10752},
  primaryclass  = {cs.CV},
}

@InProceedings{zhang2018unreasonable,
  author    = {Zhang, Richard and Isola, Phillip and Efros, Alexei A and Shechtman, Eli and Wang, Oliver},
  booktitle = {Proceedings of the IEEE/CVF Conference on Computer Vision and Pattern Recognition},
  title     = {The unreasonable effectiveness of deep features as a perceptual metric},
  year      = {2018},
  address   = {Long Beach, USA},
  pages     = {586--595},
  publisher = {IEEE},
}

\end{document}